\documentclass{article}

\PassOptionsToPackage{numbers, compress}{natbib}

\usepackage[accepted]{style}

\usepackage[utf8]{inputenc} 
\usepackage[T1]{fontenc}    
\usepackage{hyperref}       
\usepackage{url}            
\usepackage{booktabs}       
\usepackage{amsfonts}       
\usepackage{nicefrac}       
\usepackage{microtype}      
\usepackage{xcolor}         
\usepackage{mathrsfs}       
\usepackage{subcaption}

\usepackage{xspace}

\usepackage{comment}
\usepackage{amsmath}
\usepackage{amsthm}
\usepackage{graphicx}
\usepackage{rotating}

\usepackage{amssymb}
\usepackage{pifont}
\usepackage{autobreak}
\usepackage{mathtools}
\usepackage{bbm}
\usepackage[ruled,vlined, linesnumbered]{algorithm2e}
\usepackage{algorithmic}
\usepackage[parfill]{parskip}

\newcommand{\FunOne}{\mathbbm{1}}
\newcommand{\Ploss}{\ell_{P}}

\newcommand{\G}{\mathcal{G}}

\newcommand{\PDE}{\textsc{\small{PDE}}\xspace}
\newcommand{\RNN}{\textsc{\small{RNN}}\xspace}
\newcommand{\KS}{\textsc{\small{KS}}\xspace}
\newcommand{\MNO}{\textsc{\small{MNO}}\xspace}
\newcommand{\RNO}{\textsc{\small{RNO}}\xspace}
\newcommand{\FNO}{\textsc{\small{FNO}}\xspace}
\newcommand{\GRU}{\textsc{\small{GRU}}\xspace}

\newcommand{\U}{\mathcal{U}}
\renewcommand{\P}{\mathcal{P}}

\newcommand{\CDF}{\textsc{{CDF}}\xspace}

\newcommand{\Q}{\mathcal{Q}}
\newcommand{\Real}{\mathbb{R}}

\newcommand{\F}{\mathcal F}

\newcommand{\A}{\mathcal A}

\newcommand{\W}{\mathcal W}

\newcommand{\D}{\mathcal D}

\renewcommand{\L}{\mathcal L}

\newcommand{\K}{\mathcal K}

\newcommand{\wh}{\widehat}

\newtheorem{theorem*}{Theorem}[section]
\newtheorem*{proposition*}{Proposition}

\newcommand{\cmark}{\ding{52}}%
\newcommand{\xmark}{\ding{56}}%

\title{Tipping Point Forecasting in Non-Stationary Dynamics on  \\ Function Spaces}

\author{\name{Miguel Liu-Schiaffini}\affilnum{1}, \name{Clare E. Singer}\affilnum{2}, \name{Nikola Kovachki}\affilnum{3}, \name{Sze Chai Leung}\affilnum{1}, \name{Hyunji Jane Bae}\affilnum{1}, \name{Kamyar Azizzadenesheli}\affilnum{3}, \name{Anima Anandkumar}\affilnum{1} \\
\affilnum{1}\addr{Caltech} \affilnum{2}\addr{CU Boulder} \affilnum{3}\addr{NVIDIA}}

\begin{document}

\maketitle

\begin{abstract}
Tipping points are abrupt, drastic, and often irreversible changes in the evolution of non-stationary and chaotic dynamical systems. For instance, increased greenhouse gas concentrations are predicted to lead to drastic decreases in low cloud cover, referred to as a climatological tipping point.  In this paper, we learn the evolution of such non-stationary dynamical systems using a novel recurrent neural operator (\RNO), which learns mappings between function spaces. After training \RNO on only the pre-tipping dynamics, we employ it to detect future tipping points using an uncertainty-based approach. In particular, we propose a conformal prediction framework to forecast tipping points by monitoring deviations from physics constraints (such as conserved quantities and partial differential equations), enabling forecasting of these abrupt changes along with a rigorous measure of uncertainty. We illustrate our proposed methodology on non-stationary ordinary and partial differential equations, such as the Lorenz-63 and Kuramoto-Sivashinsky equations. We also apply our methods to forecast a climate tipping point in stratocumulus cloud cover and airfoil wake and stall transitions using only limited knowledge of the governing equations. For the latter, we show that our proposed method zero-shot generalizes to forecasting multiple future tipping points under varying Reynolds numbers. In our experiments, we demonstrate that even partial or approximate physics constraints can be used to accurately forecast future tipping points.\footnote{The code is available at: \url{https://github.com/neuraloperator/tipping-point-forecast}.}
\end{abstract}

\section{Introduction}

\begin{figure}[t]
    \centering
        \begin{subfigure}{0.6\textwidth}
            \centering
            \includegraphics[width=\textwidth]{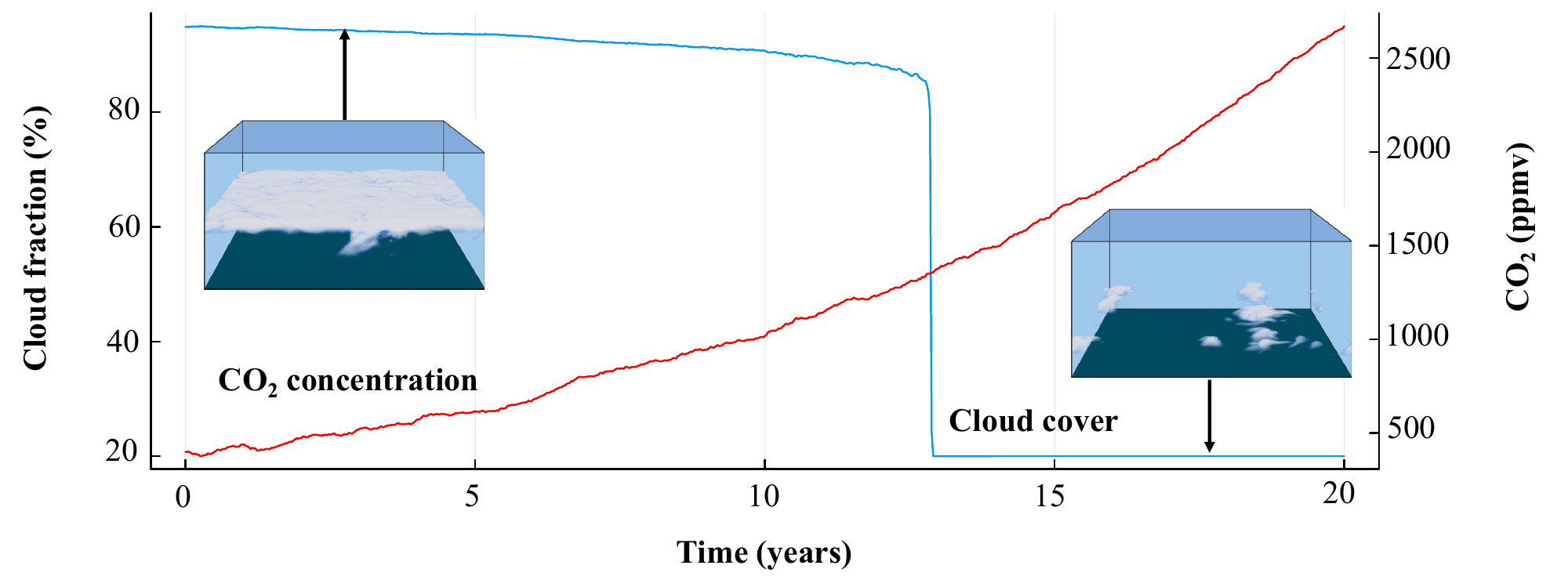}
            \caption{}
            \label{fig:cloud_diagram}
        \end{subfigure}%
        \begin{subfigure}{0.4\textwidth}
            \centering
            \includegraphics[width=\textwidth]{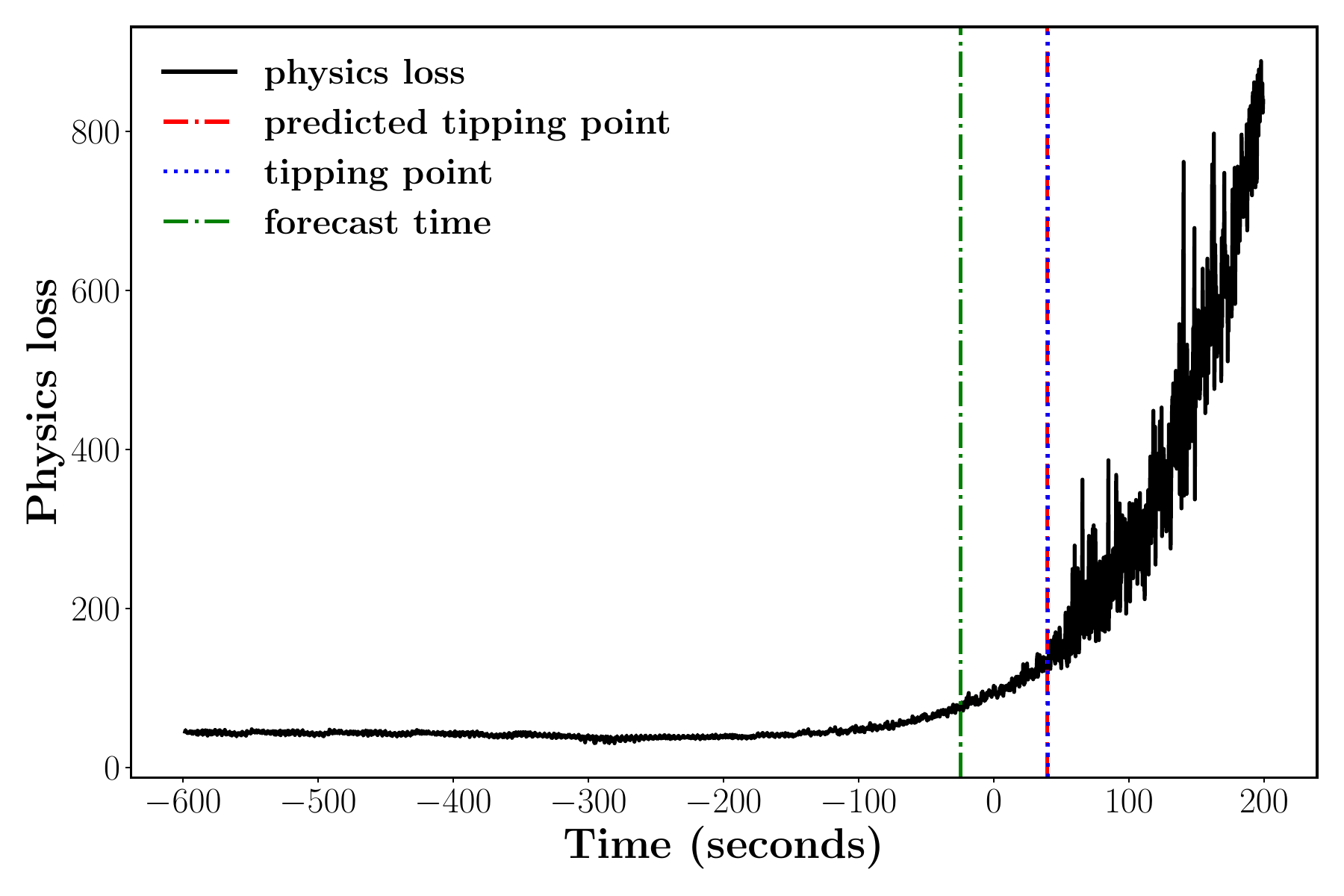}
            \caption{}
            \label{fig:tipping_pred}
        \end{subfigure}%
    \caption{\textbf{(a)} Tipping point in cloud fraction as a function of $\text{CO}_2$ concentration, from bulk model of the atmospheric boundary layer developed in \cite{Singer2023a, Singer2023b}. 3d cloud cover renderings reproduced from~\citep{schneider2019possible}. \textbf{(b)} \RNO accurately forecasts tipping point $64$ seconds ahead in non-stationary Lorenz-63 system. ``Predicted tipping point'' is the time at which our framework predicts a tipping point will occur, and ``forecast time'' is the time when our framework makes this prediction.}
    \label{fig:tipping_thresh_pred_fig}
\end{figure}

Non-stationary chaotic dynamics are a prominent part of the world around us. 
In chaotic systems, small perturbations in the initial conditions significantly affect the long-term trajectory of the dynamics. Non-stationary chaotic systems possess further complexity due to their time-varying nature. 
For instance, the atmosphere and ocean dynamics that govern Earth's climate are modeled by highly nonlinear partial differential equations (\PDE{}s). They exhibit non-stationary chaotic behavior due to changes in anthropogenic greenhouse gas emissions, insolation, and a myriad of complex internal feedbacks~\citep{lemoine2014watch,schneider2019possible} (Figure~\ref{fig:cloud_diagram}). Additionally, turbulent aerodynamics is governed by the Navier–Stokes equations, a set of nonlinear \PDE{}s that describe the motion of fluids. These equations give rise to chaotic and non-stationary behavior across a wide range of scales, especially in high-Reynolds-number flows commonly encountered in aerospace applications \cite{rogallo1984numerical}. One of the main areas in scientific computing is understanding such phenomena and providing computational methods to model their dynamics. Numerical methods, e.g., finite element and finite difference methods, have been widely used to solve \PDE{}s. 
However, numerical methods have enormous computational requirements to capture the fine scales in complex processes. Moreover, they do not provide a manageable way to learn from data to reduce scientific modeling errors.

\paragraph{Learning in non-stationary physical systems.} These complexities in modeling complex physical systems make learning the dynamics of their evolution in function spaces notoriously difficult. Prior works proposed various neural networks,  such as recurrent neural networks (\RNN{}s) \citep{rumelhart1985learning} and reservoir computing~\citep{patel2022using}, for learning such dynamical systems~\citep{rumelhart1985learning}. However, these methods learn maps between finite-dimensional spaces and  are thus not suitable for learning on function spaces, which are inherently infinite-dimensional objects. For example, when modeling spatiotemporal systems, the size of input and output layers in RNNs depend on the number of discretization points in physical space; the parameter count of RNNs depends on the input and output resolutions.

\begin{table*}
\begin{center}
\begin{tabular}{l|ccccccc}
\multicolumn{1}{c}{\bf Method}
&\multicolumn{1}{c}{\textbf{Generality}}
&\multicolumn{1}{c}{\textbf{Function space}}
&\multicolumn{1}{c}{\textbf{Partial physics}} 
&\multicolumn{1}{c}{\textbf{Speed}}
&\multicolumn{1}{c}{\textbf{Pre-tip data}}
\\
\hline 
\hline

\rule{0pt}{1em} Solver & \cmark & \cmark & \xmark & \xmark & N/A \\

\rule{0pt}{1em} EWS~\citep{scheffer2009early} & \xmark & \xmark & N/A & \cmark & \cmark \\

\rule{0pt}{1em} Bury et al.~\citep{bury2021deep} & \xmark & \xmark & N/A & \cmark & \xmark \\

\rule{0pt}{1em} Patel and Ott~\citep{patel2022using} & \cmark & \xmark & N/A & \cmark & \xmark \\

\hline
\hline 

\rule{0pt}{1em} \textbf{Ours} & \cmark & \cmark & \cmark & \cmark & \cmark \\

\hline
\hline 
\end{tabular}
\end{center}
\caption{Comparison of methods. ``Generality'' denotes applicability to arbitrary types of tipping points. ``Pre-tip data'' denotes whether the method can forecast tipping points using only pre-tipping data.}
\label{table:comparison}
\end{table*}

Recent research introduced neural operators \citep{kovachki2021neural, azizzadenesheli2024neural} for learning operators to remedy these shortcomings. Neural operators receive functions at any discretization as inputs, possess multiple layers of non-linear integral operators, and output functions that can be evaluated at any point. Neural operators are universal approximators of operators and can approximate continuous non-linear operators in \PDE{}s~\citep{kovachki2021neural}. Crucially, the number of parameters in neural operators stays constant with respect to the discretization of the inputs and outputs. These properties make neural operators discretization-invariant and suitable for learning in function spaces~\citep{berner2025principled}. Fourier neural operators (\FNO) are architectures that use Fourier representations to integrate, providing an efficient implementation for many real-world applications. These methods, in particular Markov neural operators (\MNO), have shown significant progress in learning Markov kernels of stationary dynamics systems~\citep{li2022learning}. However, non-stationary dynamical systems have potentially long memories, making earlier \MNO developments not directly applicable to the setting of this paper.

\begin{figure}
    \centering
    \includegraphics[width=\textwidth]{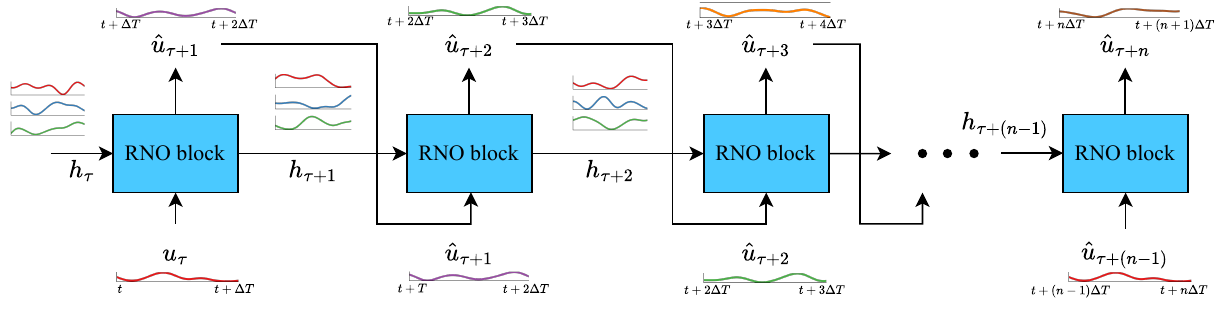}
    \caption{Illustration of \RNO auto-regressive forecasting.}
    \label{fig:rno_fig}
\end{figure}

{\bf In this work}, we introduce recurrent neural operators (\RNO{}s), a recurrent architecture for neural operators that enables learning memory-equipped maps on function spaces (Figure~\ref{fig:rno_fig}). \RNO receives a sequence of historical continuous (in time) function data and represents the memory in terms of a latent state function. This enables conditional predictions for future functions given the latent functional representations of the past.
In contrast to \RNN{}s and other fixed-discretization approaches, \emph{\RNO is discretization invariant in both space and time}. We show that \RNO{}s are capable of learning the dynamics of non-stationary chaotic systems and outperform \RNN{}s and the state-of-the-art \MNO by orders of magnitude on a model of stratocumulus cloud cover evolution~\citep{Singer2023a, Singer2023b}, as well the non-stationary Lorenz-63 system and the Kuramoto-Sivashinsky (\KS) equation.

\paragraph{Tipping point forecasting.} A commonly observed trait of non-stationarity is the existence of abrupt, drastic, and often irreversible changes in system dynamics, known as tipping points~\citep{ashwin2012tipping}. Tipping points are conceptual reference points in non-stationary systems where the evolution undergoes a sudden change in quantities of importance. 
For instance, in the climate system, several tipping point phenomena have been identified \citep{lenton2008tipping, ArmstrongMcKay2022, Wang2023}, such as the collapse of the oceanic thermohaline circulation \citep{rahmstorf2006thermohaline, bonan2022transient, rind2018multicentury}, ice sheet instability \citep{Mengel2014, favier2014retreat, seroussi2014sensitivity, joughin2021ocean}, permafrost loss \citep{koven2015simplified, miner2022permafrost}, and low cloud cover breakup \citep{schneider2019possible}. As shown in Figure~\ref{fig:cloud_diagram}, an increase in $\text{CO}_2$ in an idealized model is linked to a delayed drastic drop in cloud cover~\citep{schneider2019possible, Singer2023a, Singer2023b}. Another example of tipping points are the wake transition and static aerodynamic stall in a flow over an airfoil with an increasing angle of attack~\citep{gupta2023two, mccroskey1981phenomenon}.

\begin{figure}[t]
    \centering
    \includegraphics[width=0.8\textwidth]{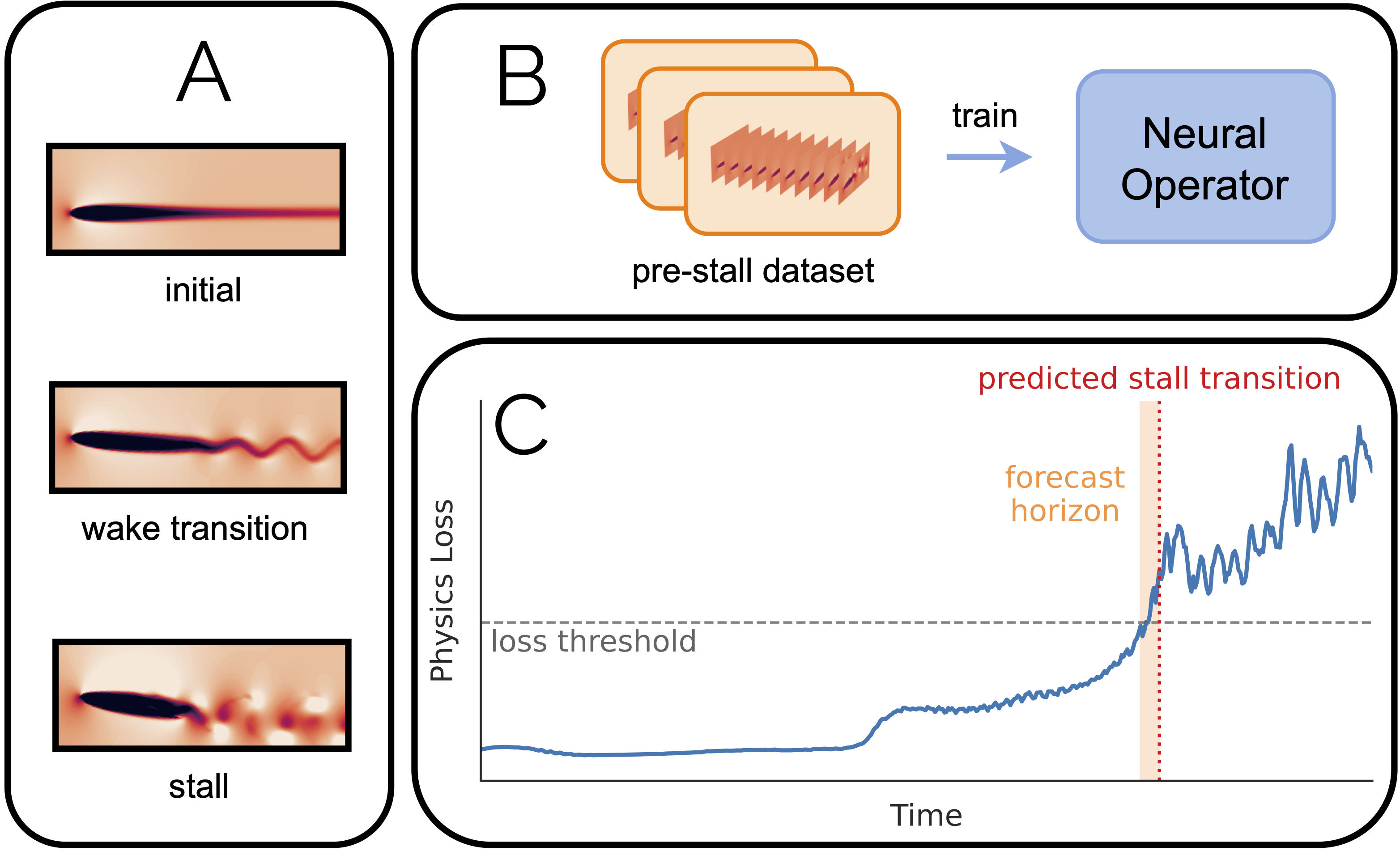}
    \caption{Conceptual diagram of our proposed method applied to a flow over an airfoil with a changing angle of attack. (A) The system exhibits two tipping points, the wake and stall transitions. In this example, we consider forecasting the stall transition. (B) We train a neural operator model to learn the pre-stall dynamics of the system. Note that our method does not require access to post-tipping data at training time. (C) At inference time, we autoregressively roll-out our model's dynamics predictions and compute the predictions' physics loss at each time-step. The first time the physics loss crosses a pre-defined loss threshold, we label this as the tipping point.}
    \label{fig:method_conceptual}
\end{figure}

Predicting tipping points in nonstationary chaotic systems poses significant challenges due to several reasons. First, chaotic systems are highly sensitive to initial conditions, making precise long-term forecasting inherently uncertain. Small variations in initial states can lead to divergent outcomes, complicating the identification of critical thresholds where abrupt transitions occur. Second, the dynamics near tipping points often exhibit complex nonlinear behaviors that defy straightforward analytical or numerical predictions. This complexity arises from the interaction of multiple variables and feedback mechanisms, further complicating the identification of impending transitions. In the context of climate change and airfoil stall, where small changes can lead to large-scale effects such as extreme weather events or loss of lift, accurately pinpointing tipping points is crucial yet remains a formidable scientific and computational challenge.

Identifying and analyzing tipping points in chaotic systems often involves running numerical solvers over extended time horizons, which results in significant computational expense \citep{schneider2019possible, bonan2022transient}. Therefore, solvers are often run at low resolution or over subsets of the full domain, which can result in loss of accuracy or generality~\citep{schneider2019possible}.

Recently, machine learning methods have been deployed for predicting the tipping points of non-stationary systems \citep{bury2021deep, patel2022using,patel2021using,kong2021machine,lim2020predicting, li2023tipping, deb2022machine, sleeman2023generative, huang2024deep, panahi2024machine, fabiani2024task, panahi2026unsupervised}. However, prior works either focus on detecting domain shift instead of forecasting tipping points far ahead in time, and those that do forecast tipping points lack a systematic approach to learn non-stationary systems at arbitrary resolutions and forecast tipping points in function space.

{\bf In this work}, we propose training an \RNO model to predict future states on the complex dynamics of non-stationary chaotic systems \emph{using only data of pre-tipping point events}. The set-up is motivated by climate modeling, where tipping points are not present in forty years of historical data~\citep{hersbach2020era5}. To forecast the time of a tipping point, we build our approach based on the common observation that, at a tipping point, the prediction accuracy of machine learning models degrades due to distribution shifts~\citep{patel2022using,patel2021using}. Since in our setting we do not have access to post-tipping data, we instead propose to forecast tipping points whenever the model prediction of the future exhibits extensive violations of domain-specific physical constraints. We use these physical constraints (e.g., conservation laws or governing \PDE{}s) to verify the correctness of our model's system forecasts at a given future time. Our method assumes that a given \RNO model is well-trained on pre-tipping dynamics. Crucially, since the model is trained only on pre-tipping dynamics, even a perfect model on the pre-tipping regime would degrade in prediction accuracy beyond the tipping point, as observed empirically. We show that this approach predicts tipping points very far in advance. See Figure~\ref{fig:method_conceptual} for a conceptual diagram.

To quantify the prediction accuracy, we propose a new conformal prediction method based on cumulative distribution functions (\CDF{}s)~\citep{shafer2008tutorial}. We estimate the \CDF of physics error, which quantifies the violation of physics constraints by the model's forecasts and plays the role of a non-conformity score. We employ the fact that using finitely-many samples, a \CDF of any distribution can be estimated accurately everywhere under the Kolmogorov–Smirnov distance, a property that is not true for other functions like quantiles~\citep{dvoretzky1956asymptotic,massart1990tight}. Using the empirical \CDF{} of the physics error, we quantify the distribution of the non-conformity score. This uncertainty quantification is directly connected to the false positive rate of our prediction approach. 

We consider a real-world system describing tipping points in stratocumulus cloud cover~\citep{Singer2023a, Singer2023b} and wake and stall tipping points in flow over an airfoil~\citep{mccroskey1981phenomenon}. We also consider the more classical non-stationary and chaotic Lorenz-63 and Kuramoto–Sivashinsky (\KS) systems. We show that \RNO consistently outperforms the state-of-the-art adapted baseline \MNO in forecasting tipping points in these systems, particularly for the \KS equation. We show that \RNO better fits the underlying physical equations of the systems. Furthermore, we demonstrate under our proposed approach, \RNO is able to accurately forecast tipping points when run for up to $200$ time intervals (Figure~\ref{fig:tipping_pred}). Our proposed method can also forecast real-world tipping points, such as the aforementioned cloud cover transition, using only partial knowledge of the underlying physics (Figure~\ref{fig:rno_clouds_case1}). It does so with an error of only 10 days while forecasting nearly 2.5 years in advance.

To demonstrate the scalability of our method to high-dimensional settings, we consider the problem of forecasting the wake and stall transitions in a flow over an airfoil. We find that we are able to successfully forecast the tipping points using limited knowledge of the underlying governing equations (specifically, using only the divergence-free condition) and for various airfoil velocities. Further, our method is capable of forecasting the stall tipping point while trained only on pre-wake data, as well as generalizing to forecasting tipping points at a Reynolds number of 5000 while only being trained on Reynolds number 1000 trajectories. We find that our method is also capable of avoiding false positive forecasts when the underlying nonstationary system does not exhibit a tipping point.

\textbf{In summary,} we propose the first tipping point forecasting framework that scales to \emph{arbitrary tipping points} and \emph{arbitrary spatiotemporal systems} on function spaces. In particular, we
\begin{enumerate}
    \item Introduce \RNO, an operator learning architecture to learn complex non-stationary dynamics, which far outperforms Markovian and \RNN baselines in our experiments. For instance, \RNO provides up to 70.9\% improvement in 20-step autoregressive prediction on the non-stationary Kuramoto-Sivashinsky equation.
    \item Construct a tipping point forecasting framework that relies only on (1) a data-driven model of \emph{pre-tipping dynamics} and (2) a physical constraint of the system.
    \item Propose a novel conformal prediction method to quantify deviations in model physics error in a statistically rigorous manner.
    \item Achieve accurate tipping point forecasting far in advance and demonstrate our method's scalability to spatiotemporal \PDE{}s such as wake and stall transitions in a flow over an airfoil.
    \item Show high accuracy in tipping point forecasting using \emph{approximate and partial} physical constraints, generalizing our method to settings when full physics laws are unknown.
    \item Demonstrate high accuracy in zero-shot tipping point forecasting in out-of-domain regimes unseen at training time.
\end{enumerate}

\section{Related Work}

Tipping points exhibit complex behavior, making their study a challenging task in the high-dimensional multi-modal setting~\citep{lemoine2014watch}. In recent years, tipping points and their root causes have been categorized for simple systems (such as low-dimensional ODEs), based on stochasticity, rate change, and bifurcations~\citep{ashwin2012tipping,ashwin2021physical,kaszas2019tipping}. Tipping points can be caused by changes in the condition and sourcing in differential equations, or sometimes, the root cause is the rate at which these conditions change~\citep{ashwin2012tipping,panahi2025global}. However, in the case of general scientific computing problems on natural phenomena, theoretical studies of tipping points are very challenging, and prior works have primarily focused on their empirical evaluation via numerical simulation. 

Tipping points are abundant in scientific computing of climate evolution and control modeling~\citep{lenton2008tipping, ArmstrongMcKay2022, Wang2023}. 
Previous work has highlighted the importance of predicting tipping points in the Earth system with the goal of giving early warning ahead of large changes in the climate \citep{Scheffer2009, Lenton2011}.
Approaches have varied depending on the type of tipping behavior considered \citep{Scheffer2012}.
For instance, it has been shown that changes in atmospheric CO$_2$ levels can cause rapid changes in low cloud cover on the Earth \citep{schneider2019possible, Singer2023b}. And furthermore, after bringing CO$_2$ levels below the tipping threshold, these changes may not be readily undone (the system exhibits hysteresis). 

Traditional early warning signals (EWS) of tipping points such as increased autocorrelation before tipping events have been studied extensively for many simple systems~\citep{lenton2011early, ditlevsen2010tipping} and for some specific real-world data~\citep{dakos2024tipping}. In the stationary regime, these empirical signals rest on a rigorous mathematical foundation: the statistical properties of systems near critical transitions are characterized through the spectrum (particularly by the spectral gap) of the generator of the transfer semigroup~\citep{lucarini2023theoretical, lucarini2024detecting, tantet2018resonances}. Finally, tipping points in the large spatiotemporal systems that motivate this work have not yet been sufficiently studied empirically and theoretically for these indicators to become evident, and for many spatiotemporal systems, thorough empirical evaluation is difficult due to the computational cost of simulations. Thus, data-driven approaches for learning system dynamics and forecasting tipping points is critical for most challenging real-world problems. In Appendix~\ref{appdx:ews_comparison}, we demonstrate that traditional EWS are not reliable at predicting tipping points in the systems we study. Recently, machine learning methods have been deployed for learning in non-stationary dynamical systems and to predict their tipping points~\citep{bury2021deep, patel2022using,patel2021using,kong2021machine,lim2020predicting, li2023tipping, deb2022machine, sleeman2023generative, huang2024deep, panahi2024machine, fabiani2024task, panahi2026unsupervised}. However, many works require post-tipping data~\citep{patel2022using, bury2021deep}, large datasets of labeled time series~\citep{huang2024deep}, full governing equations~\citep{lim2020predicting}, or are not generalizable to arbitrary types of tipping points~\citep{bury2021deep, deb2022machine, fabiani2024task} or large spatio-temporal systems~\citep{panahi2024machine}. In contrast, our method is scalable, does not need full governing equations, and is not trained on post-tipping data. See Table~\ref{table:comparison} and Appendix~\ref{appdx:comparison} for a detailed comparison with prior work.

\RNN{}s are often used to learn in finite-dimensional dynamical systems with memory or in time-series with discrete-time dynamics~\citep{rumelhart1985learning, gopakumar2023fourier}. They consist of blocks of neural networks that carry memory as a finite-dimensional latent state and enable conditional predictions on learned latent historical representations. But since neural networks are maps between finite-dimensional spaces, \RNN{}s are unfit for many scientific computing problems, which often involve \emph{functions} (i.e., infinite-dimensional objects) and their time-evolution. Neural ODE versions of \RNN{}s have similar issues, can only model specific time derivatives, and require expensive solvers for temporal simulation~\citep{rubanova2019latent,habiba2020neural, andrew2022learning}.

Neural operators are deep learning models that generalize neural networks to maps between function spaces~\citep{azizzadenesheli2024neural,li2020fourier,rahman2022u}, and they are universal approximators of general operators~\citep{kovachki2021neural}. The inputs to neural operators are functions, and the output function can be evaluated at any point in the function domain. These models are thus known to be \emph{discretization invariant} and can take the input function at any resolution. In this paper, we develop the recurrent neural operator (\RNO). \RNO receives a sequence of historical function data and represents the memory in terms of a latent state function, enabling conditional predictions given learned latent functional representations of the past.

After the initial version of this paper~\citep{liu2023tipping} was released, a few related methods were introduced in the literature. \citet{buitrago2024benefits} introduce MemNO, an architecture that combines state space models with \FNO by introducing memory, and \citet{guo2025history} propose a memory-free architecture where the model inputs are short time intervals. \citet{ye2025recurrent} train neural operators autoregressively for better long-time rollouts. \citet{koren2025merging} factorize the spatial and temporal dimensions and propose to use structured state space models to model each dimension.

\section{Preliminaries}
\label{sec:prelim}
A \PDE can be seen as a differential law on function spaces. For an input function $a$ from a function space $\A$ (e.g., a fluid velocity field), we denote $\D_\A$ as the domain and $\Real^{d_\A}$ as the function co-domain with dimension $d_\A$. For any point in the domain $x\in\D_\A$, the function $a$ maps this point to a $d_\A$-dimensional vector, i.e, $a(x)\in\Real^{d_\A}$ (e.g., the velocity of a fluid at a particular spatial location $x$). Correspondingly, let $\U$ denote the solution or output function space (e.g., the fluid velocity field at a later time), defined on domain $\D_\U$ with co-domain $\Real^{d_\U}$, i.e., for any function $u\in\U$ and any $x\in\D_\U$, we have $u(x)\in\Real^{d_\U}$. For a given $a$, we define a \PDE in generic form as,
\begin{align}
    \label{eq:general_pde}
    \begin{split}
        \L (u,a)(x) &= 0, \qquad \text{in } \D_\U, \\
        \L' (u,a)(x) &= 0, \qquad \text{in } \partial \D_\U,
    \end{split}
\end{align}

where $\L (u,a)$ is the governing law in the domain $\D_\U$, and $\L' (u,a)(x)$ is the constraint on the domain boundary $\partial \D_\U$. Following the velocity field example, this may be the Navier-Stokes equations with a given set of boundary conditions. A function $u$ is a solution to the above \PDE at input $a$ if the function $u$ satisfies both of the \PDE constraints (Eq.~\ref{eq:general_pde}), comprising an input-solution pair $(a,u)$.
We are concerned with learning maps from input function spaces $\A$ to output function spaces $\U$ in operator learning (e.g., learning the map of a fluid velocity field from time $t$ to $t + h$). For a given \PDE, let $\G^\star$ denote the operator that maps the input functions in  $\A$ to their corresponding solution functions in $\U$, which we seek to learn with a neural operator. 

A neural operator $\G$ is a deep learning architecture consisting of multiple layers of point-wise and integral operators~\citep{kovachki2021neural} to learn a map from input function $a$ to output function $u$. The first layer of the architecture is a pointwise lifting operator $\P_0$ such that for a given function $a$, the output of this layer $\nu_0=\P_0 a$ is computed so that for any $x\in\D_\A$, $\nu_0(x) = P_0(a(x))\in\Real^{d_0}$ where $P_0$ is a learnable neural network. This follows standard deep learning architectural practice, where the computation is performed in a higher-dimensional latent space. This step is followed by $L$ layers of nonlinear integral operators. For any layer $l$ we have,
\begin{align}
    \nu_l = \P_l(\sigma(\K_l \nu_{l-1} )+ \W_l\nu_{l-1}s_l ),
\end{align}
where $\nu_l:D_l\rightarrow \Real^d_l$, $\sigma$ is some pointwise nonlinearity, and $\K_l$ is an integral operator such that for any $x\in\D_l$, we have,
\begin{align}
\label{eq:kernel-integral}
    \K_l\nu_{l-1} (x) = \int_{\D_{l-1}} \kappa_l(x,y)\nu_{l-1}d\mu_{l}(y)+\W_l'\nu_{l-1}(s_l(x)).
\end{align}
Here $\kappa_l$ is a learnable kernel function, $\W_l'$ is a pointwise operator parameterized by a neural network $W_l'$, and $s_l$ is a deformation map from $\D_l$ to $\D_{l-1}$. Similarly, $\W_l$ is a pointwise operator parameterized by $W_l$ neural network, and $\P_l$ is a pointwise operator parameterized by a neural network $P_l$. $\mu_l$ represents the measure on each space and finally, we set $\D_L=\D_\U$. Intuitively, this kernel integral generalizes matrix multiplication in standard neural networks by integrating over the spatial domain $\D_\U$, where the specific discretization points enter as quadrature in the integral~\citep{berner2025principled}. Parameterizing the kernel $\kappa_l$ with a finite number of parameters makes the parameter count of a neural network fixed with respect to the spatial discretization. The last layer is a pointwise projection operator $\Q$ parameterized by a neural network $Q$. In particular, we use Fourier representations to compute the integration Eq.~\ref{eq:kernel-integral} in its convolution form,
\begin{align}
    \label{eq:fourier_layer}
    \K_l\nu_{l-1} (x) = \F^{-1} \left( F\kappa_l\cdot\F\nu_{l-1} \right) +\W_l'\nu_{l-1}(s_l(x)),
\end{align}
where $\F$ denotes the Fourier transform, and project its inputs to $k_l$ Fourier bases, for some hyperparameter $k_l$. This represents the integral operator with a finite number of parameters but still exhibits invariance to the input and output resolutions. Let $R_l$ denote the $k_l$ Fourier coefficients of $\kappa_l$, i.e., $R_l = \F\kappa_l$. Following \citep{li2020fourier}, we directly parameterize $R_l$ instead of $\kappa_l$ to learn the integral operator. We utilize these building blocks to construct the \RNO architecture in the next section. 

\section{Recurrent neural operator (\RNO)}\label{Sec:RNO}

We now describe and formulate \RNO, a generalization of \RNN{}s to function spaces. For a given \PDE describing the evolution of a function $u(x,t)$ of time and space, consider a partition of the time domain $t$ into equally-spaced intervals of length $\Delta T$ and indexed by $\tau \in \mathbb{N}$. A step $\tau$ is associated with the interval $[(\tau - 1) \cdot \Delta T,\tau \cdot \Delta T]$, where the input function restricted to this interval is defined for $t\in[t,t + \tau \cdot \Delta T]$ as $u_\tau(x,t):= u(x,t)$.

Our proposed architecture of the \RNO parallels that of the gated recurrent unit (\GRU) \citep{cho2014learning}, where the \GRU cell at step $\tau$ receives a hidden state vector $h_{\tau}$ and an input vector $x_\tau$, and predicts the next output vector $x_{\tau + 1}$ and next hidden state vector $h_{\tau + 1}$, which is also passed to the \GRU cell at step $\tau+1$.
The key difference between the \RNO and the \GRU is that at step $\tau$, an \RNO cell receives as input a \emph{hidden function} $h_{\tau}(x,t)$ (which itself is represented at arbitrary resolutions) and an input function $u_\tau(x,t)$. As such, we replace the linear maps in the \GRU architecture with Fourier layers, Eq.~\ref{eq:fourier_layer}). Given $u_\tau$ and $h_{\tau}$, we define the reset gate (which intuitively determines what to forget from the past) such that for any $t\in[t,t + \tau \cdot \Delta T]$,
\begin{equation}
    \label{eq:reset_gate}
    r_\tau(x,t) = \sigma \Big((\K_r u_\tau)(x,t) + (\K'_r h_{\tau})(x,t) + b_r(x,t) \Big),
\end{equation}
where $\K_r,\K'_r$ are Fourier layers, $b_r(x,t)$ is a learned bias function, and $\sigma$ is the sigmoid function applied pointwise. We define the update gate (which intuitively determines what to remember for the future) as,
\begin{equation}
    \label{eq:update_gate}
    z_\tau(x,t) = \sigma \Big((\K_z u_\tau)(x,t) + (\K'_z h_{\tau})(x,t) + b_z(x,t) \Big).
\end{equation}
We define the candidate hidden function as,
\begin{equation}
    \label{eq:candidate_hidden_state}
    \wh h_{\tau + 1}(x,t) = \phi \Big((\K_h u_\tau)(x,t) + (\K'_h (r_\tau \cdot h_{\tau}))(x,t) + b_h(x,t) \Big),
\end{equation}
where $\phi$ denotes a point-wise scaled exponential linear unit (SELU) activation and the function multiplication $(\cdot)$ is taken to be pointwise. Finally, we define the hidden state function $h_{\tau+1}(x,t)$ as
\begin{equation}
    \label{eq:output_function}
    h_{\tau + 1}(x,t) = \Big((1-z_\tau)\cdot h_{\tau}\Big)(x,t) + \Big(z_\tau \cdot \wh h_{\tau + 1} \Big)(x,t).
\end{equation}
We take $h_0(x,t)$ to be a learned prior function on the initial hidden state. Let $\text{\RNO}(v, h_{\tau})$ denote an \RNO cell. Given the input function $u_\tau$, an \emph{\RNO block} takes the following form:
\begin{align}
    \tilde u_{\tau}(x,t) &= \P_0(u_{\tau})(x,t) = P_0(u_{\tau}(x,t)), \\
    v^{(i+1)}_{\tau}(x,t) &= \text{\RNO}^{(i+1)}\left(v^{(i)}_{\tau}, h^{(i+1)}_{\tau} \right)(x,t), \\
    \wh u_{\tau}(x,t) &= \Q(v^{(L)}_{\tau})(x,t) = Q(v^{(L)}_{\tau}(x,t)),
\end{align}
where $\P_0$ and $Q$ are pointwise lifting and projection operators, respectively parameterized by neural networks $P_0$ and $Q$. $v^{(0)}_{\tau}(x,t) := \tilde u_\tau(x,t)$, the result of the lifting operation, and $L$ is the number of \RNO layers. An \emph{\RNO} is then a \RNO block evolving in time with $h^{(i)}_{\tau + 1} := v^{(i)}_{\tau}$ for each \RNO cell $i \geq 1$, see Figure~\ref{fig:rno_fig}.

We divide the inference stage of \RNO into two phases:
\begin{enumerate}
    \item \textbf{Warm-up phase:} In the warm-up phase, we have access to the ground-truth trajectory between steps $\tau=0$ and $\tau = N_\tau - 1$, corresponding to the ground-truth solution $u(t)$ between times $t \in [0, N_\tau \Delta T]$. During this phase, the \RNO blocks receive as input $u_\tau$, and the predicted $\wh u_{\tau+1}$ is discarded.
    \item \textbf{Prediction phase:} In the prediction phase, we assume we no longer have access to the ground-truth trajectory beyond time $\tau = N_\tau -1$, so the model predictions $\wh u_{\tau+j}$ are taken as input to the \RNO block at time $\tau + j + 1$, and the model is thus composed with itself (with access to its hidden state $h_\tau$), with its outputs fed back on itself as inputs.
\end{enumerate}

The prediction phase is similar to Markov neural operator (\MNO)~\citep{li2022learning}, with the exception that \RNO now has access to a continually-updating hidden representation $h_\tau(x,t)$, which encodes the history of the non-stationary system, making \RNO a generalization of \MNO to systems with memory.

\section{Tipping-point Forecasting}
\label{sec:dkw}

Forecasting tipping points requires rigorous uncertainty quantification, and any prediction must be accompanied by the event's potential probabilities. Conformal prediction is a field of study that accompanies predictions with certainty levels in terms of a conformity score~\citep {shafer2008tutorial}. Most prior work on traditional conformal prediction quantifies the probability distribution of the conformity score at certain levels. Another approach is to utilize quantile regression to quantify the distribution of conformity scores ~\citep{takeuchi2006nonparametric}. Both of these approaches have shortcomings in tipping point prediction. The former approach utilizes exchangeability and requires new and fresh draws of samples to quantify the distribution of conformity scores, imposing a union bound over a long period of time. Quantile regression suffers from two downsides: (1) quantiles are not generally estimable, and (2) a union bound over an infinite set is required to quantify the conformity score distribution~\citep{huang2021off}. These limitations are addressed below by introducing \CDF estimation methods. 

Our tipping point forecasting utilizes training on the pre-tipping dynamics of the system. We train \RNO on data sets of time-evolving functions that \emph{do not} contain any tipping points. We make this choice to resemble real-world settings such as climate where recorded historical data does not contain tipping points. At any time $t$, we use \RNO to predict the long evolution of the dynamics for an interval of $[t,t+ T]$, i.e., $\wh u(x,[t,t+ T])$ where $T$ is a multiple of $\Delta T$. At the forecast time $t$, we do \emph{not} have access to the future $u(x,[t,t+ T])$. If we did, we could use the deviation of $\wh u(x,[t,t+ T])$ from $ u(x,[t,t+ T])$ as a hindsight indicator for the tipping point as observed in prior works~\citep{patel2022using,patel2021using}. 

Instead, we propose to use the physical constraints $\L(u)=0$ of the underlying system (such as the \PDE that governs the dynamics) to evaluate the deviation of $\wh u(x,[t,t+ T])$ from physics laws, and we use this signal as an indicator of tipping points. 
For a predicted function $\wh u(x,[t,t+ T])$, we define the conformity score as $\| \L(\wh u)-0 \|$, and we refer to it by the physics loss $\Ploss(\wh u)$. Note that if $\wh u(x,[t,t+  T]) = u(x,[t,t+  T])$ (the true future function), then $\L(\wh u) = 0$, and thus our proposed indicator aligns with the hindsight indicator mentioned above. At any time $t$, we compute $\Ploss(\wh u)$, and if it takes values above a certain threshold, we forecast that a tipping point is anticipated. This method enables tipping point forecasting. However, since the trained model comes with training generalization error, this approach makes mistakes, confusing large errors on pre-tipping regimes and actual tipping incidence. This is of particular importance when the forecasting interval of interest $T$ is large. We propose the following novel conformal prediction to address uncertainty quantification in tipping point forecasting.
\begin{figure}
    \centering
    \includegraphics[width=0.8\textwidth]{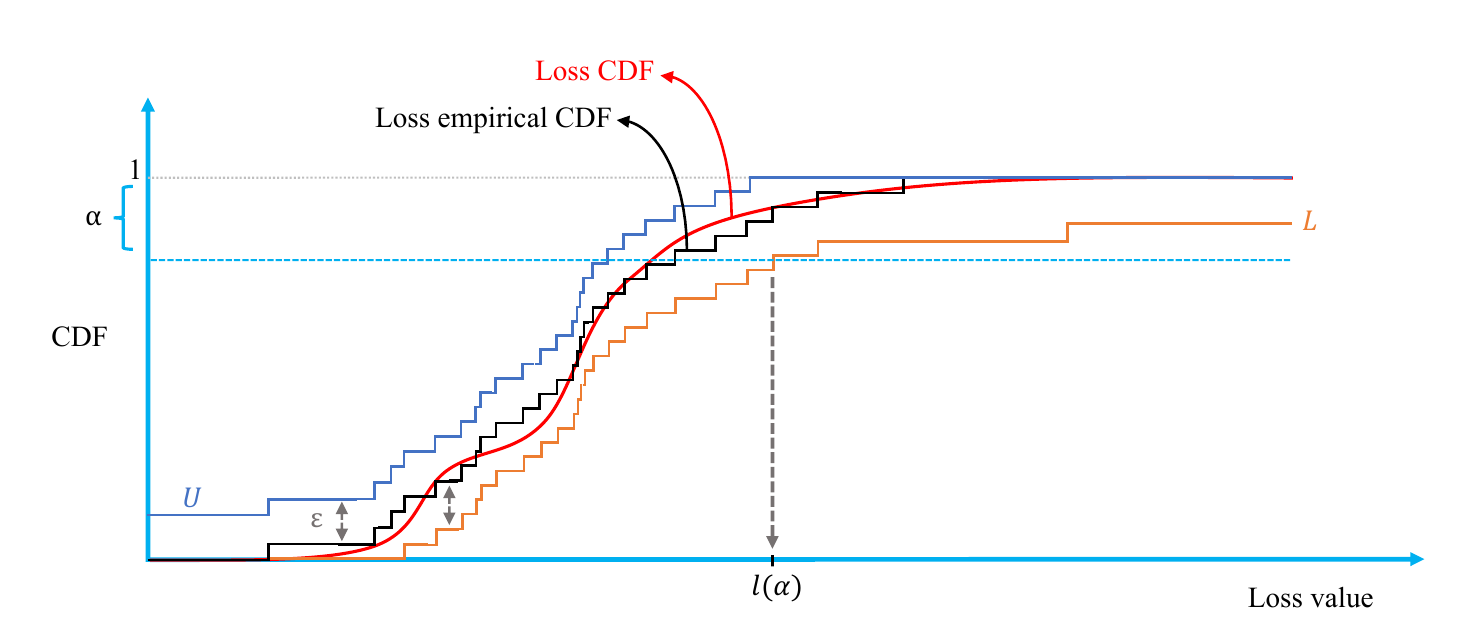}
    \caption{Diagram of the setup for the tipping point prediction method. $U$ and $L$ correspond to the upper and lower bounds given by the \CDF concentration inequality, differing from the empirical \CDF by $\varepsilon$. $\alpha$ is the significance level, and $l(\alpha)$ is the corresponding loss of $L$ at $\alpha$.}
    \label{fig:CDF}
\end{figure}
For a trained model and given a forecasting interval $T$, let $F$ denote the true \CDF of the conformity score. The \CDF captures the probabilities at which the model makes errors of a given magnitude on pre-tipping data. We use the \CDF to develop our conformal prediction method, which allows for uncertainty quantification of tipping point forecasts. In particular, for any loss threshold at which we call a tipping point, we can denote $\alpha$ to be the probability of falsely forecasting a tipping point when the data was simply from the pre-tipping regime. We first compute the empirical \CDF $\wh F$ of the conformity score $\Ploss(u)$ using $n$ calibration samples. Formally, given $n$ samples, we construct $\lbrace {\Ploss}^{(j)}\rbrace_{j=1}^{n}$ and compute $\wh F$ as follows. For any $p\in\Real$, 
\begin{align*}
    \wh F(p) :=\frac{1}{n}\sum_j^n \FunOne(p\geq  {\Ploss}^{(j)}).
\end{align*}
We then use the concentration inequality of \CDF{}s to compute its upper and lower bounds~\citep{dvoretzky1956asymptotic,massart1990tight}. Using the concentration inequality of \CDF{}s, with probability at least $1-\delta$ we have,
\begin{align*}
    \sup_{p\in\Real} \left| \wh F(p) -F(p) \right| \leq \varepsilon:=\sqrt{\frac{\log(2/\delta)}{2n}},
\end{align*}
where the left-hand side is the Kolmogorov–Smirnov distance between the two \CDF{}s. We define the \CDF  lower bound $L(p):=\max\lbrace 0,\wh F(p) - \varepsilon \rbrace$ and the \CDF  upper bound $U(p):=\min\lbrace 1,\wh F(p) + \varepsilon \rbrace$. Figure~\ref{fig:CDF} depicts this procedure. Intuitively, $\alpha$ denotes the maximum probability of a mistake that one is willing to tolerate. In our approach, as shown in Figure~\ref{fig:CDF}, we first find $l(\alpha)$, the conformity level at which the \CDF lower bound $L$ has probability $1-\alpha$. Using the level $l(\alpha)$, we call the presence of a tipping point whenever the conformity score of our prediction $\Ploss (\wh u)$ is above $l(\alpha)$.

\begin{proposition*}
    For any given $\alpha\in[0,1]$, the decision rule of calling for a tipping point at level $l(\alpha)$ has a false positive rate of at most $\alpha$, and this statement holds with probability at least $1-\delta$.
\end{proposition*}

The proof is based on the fact that for any $\alpha$, $F^{-1}(1-\alpha)\leq l(\alpha)$ and $F(l(\alpha))\geq 1-\alpha$. Therefore, when $\Ploss\geq l(\alpha) $, signaling a tipping point, we also have $\Ploss \geq F^{-1}(1-\alpha)$. Thus, the probability of the event $\Ploss\geq l(\alpha) $ is lower bounded by the probability of $\Ploss \geq F^{-1}(1-\alpha)$ which at most $\alpha$. This proposition simultaneously holds for all the time steps $\tau$ and all $\alpha$. This property enables assessing tipping predicting at various levels, a crucial feature for risk assessment and policy making.

\section{Experimental Results}
\label{sec:exp}

We present experimental results on learning dynamics and forecasting tipping points in non-stationary systems. We showcase the performance of \RNO and the tipping point forecasting method on a simplified model of cloud cover equations and on predicting the turbulent wake and stall tipping points in a 2-d flow over an airfoil. In Appendix~\ref{appdx:lorenz}, we also present detailed results for the finite-dimensional Lorenz-63 system, and for the (infinite-dimensional) \KS \PDE in Appendix~\ref{appdx:results_ks}. For Lorenz-63, we show that our method is capable of forecasting tipping points far in advance \emph{using only approximate knowledge of the underlying system}. In Appendix~\ref{appdx:rate_induced_no_tip}, we show that our method is robust to different rates of parameter change and can avoid predicting a tipping point when the underlying system lacks one. For the non-stationary \KS equation, we show that \RNO far outperforms \MNO and \RNN in state prediction for rollouts of various lengths. Discussion on the technical details of \RNO (e.g., memory usage, inference times, hyperparameter selection, etc.) can be found in Appendix~\ref{appdx:rno_technical_analysis}.

\subsection{Learning non-stationary dynamics}
\label{subsec:learning_dynamics}

\begin{table*}
\begin{center}
\begin{tabular}{l|ccccccc}
\multicolumn{1}{c}{\bf Model}
&\multicolumn{1}{c}{\textbf{1-step}}
&\multicolumn{1}{c}{\textbf{2-step}}
&\multicolumn{1}{c}{\textbf{4-step}} 
&\multicolumn{1}{c}{\textbf{8-step}}
&\multicolumn{1}{c}{\textbf{12-step}}
&\multicolumn{1}{c}{\textbf{16-step}} 
&\multicolumn{1}{c}{\textbf{20-step}}
\\
\hline 
\hline 
\rule{0pt}{1em} \RNO & $\mathbf{0.0030}$ & $\mathbf{0.0027}$ & $\mathbf{0.0031}$ & $\mathbf{0.0419}$ & $\mathbf{0.1410}$ & $\mathbf{0.2919}$ & $\mathbf{0.4008}$ \\

\rule{0pt}{1em} \MNO & $0.0050$ & $0.0073$ & $0.0150$ & $0.2901$ & $1.0751$ & $1.2626$ & $1.3772$\\

\rule{0pt}{1em} \RNN & $0.0525$ & $0.0481$ & $0.1443$ & $0.5488$ & $0.9031$ & $1.0031$ & $1.1561$ \\
\hline
\hline 
\end{tabular}
\end{center}
\caption{Relative $L^2$ errors on the non-stationary \KS equation for different $n$-step prediction settings. \RNO outperforms both \MNO and the baseline \RNN on every forecasting intervals.} 
\label{table:ks_results}
\end{table*}  

We compare \RNO against Markov neural operators (\MNO) \citep{li2022learning} and \RNN{}s in forecasting non-stationary dynamics. There are two key properties that any learned model must exhibit: (1) low step-wise error and (2) error stability in long-time predictions (i.e., when the model is composed with itself many times). Low step-wise error is a necessary prerequisite to adequately learning the dynamics of any system. However, for many downstream tasks of significant scientific importance, it is also crucial that our model's error remain low when forecasting far into the future. This ensures the accuracy of tipping point prediction is credible in complex real-world settings. Hereafter, we define ``$n$-step prediction'' to be a model's forecasting prediction looking ahead $\tau = n$ steps.

In our experiments, we employ a multi-step training regimen for all models, where the total loss is a combination of the loss on the model's $n$-step prediction for $n \in \{1, 2, \dots, M\}$. That is, our data loss function is given by
\begin{equation}
    \label{eq:loss}
    \ell_D (u, \wh u) = \sum_{n=1}^M \lambda_n \| u_n - \wh u_n \|_2,
\end{equation}
where $\| \cdot \|_2$ is the $L^2$ norm and $\lambda_n$ is some scaling hyperparameter.

Table~\ref{table:ks_results} compares the relative $L^2$ errors of \RNO, \MNO, and \RNN on forecasting the non-stationary KS equation (setup and details in Appendix~\ref{appdx:results_ks}). We observe that \RNO outperforms \MNO and \RNN on every forecasting interval. Further, we can observe the disadvantages of \RNN and other finite-dimensional, fixed-resolution models on learning in \PDE (i.e., infinite-dimensional) settings. The \RNN's error is orders of magnitude larger than that of \RNO or \MNO for small forecasting windows. At large forecasting windows (i.e., 12-step and above), the lack of explicit history in \MNO becomes a severe disadvantage. In contrast, \RNO maintains much lower error at large forecasting intervals than the other two models. Both the \MNO and \RNO shown in Table~\ref{table:ks_results} were trained with $M = 5$ steps of fine-tuning (Eq.~\ref{eq:loss}). Observe that \MNO and \RNN have difficulty generalizing to forecasting intervals larger than $M$, whereas \RNO is capable of maintaining stable error even at longer forecasting intervals.

\subsection{Cloud cover equations: tipping points from partial physics}
\label{sec:cloud_cover_exp}
\begin{figure}[t]
    \centering
        \begin{subfigure}{0.5\textwidth}
            \centering
            \includegraphics[width=\textwidth]{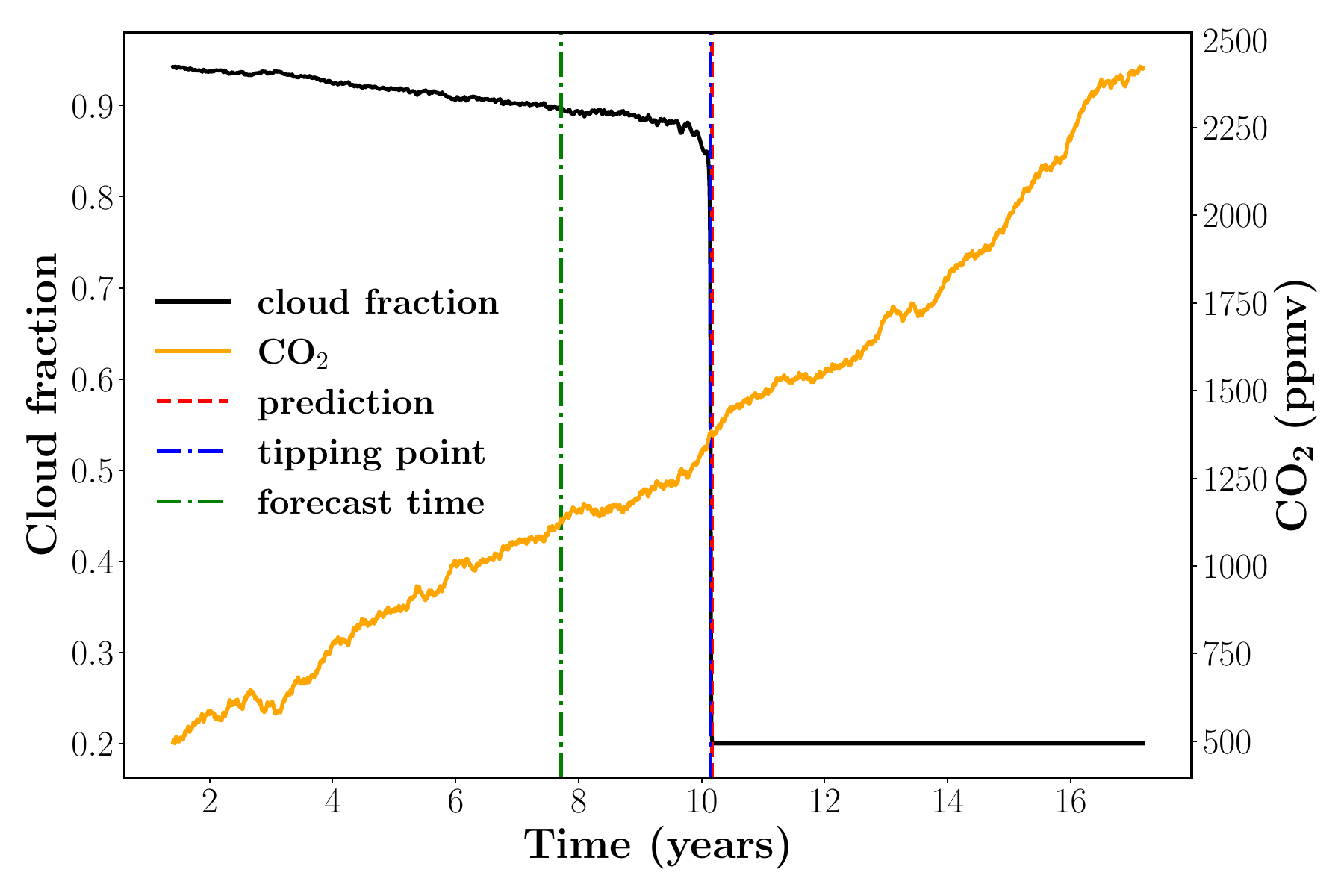}
            \caption{Tipping point forecast w.r.t. cloud fraction}
            \label{fig:rno_clouds_case1}
            \end{subfigure}%
            \begin{subfigure}{0.5\textwidth}
            \centering
            \includegraphics[width=\textwidth]{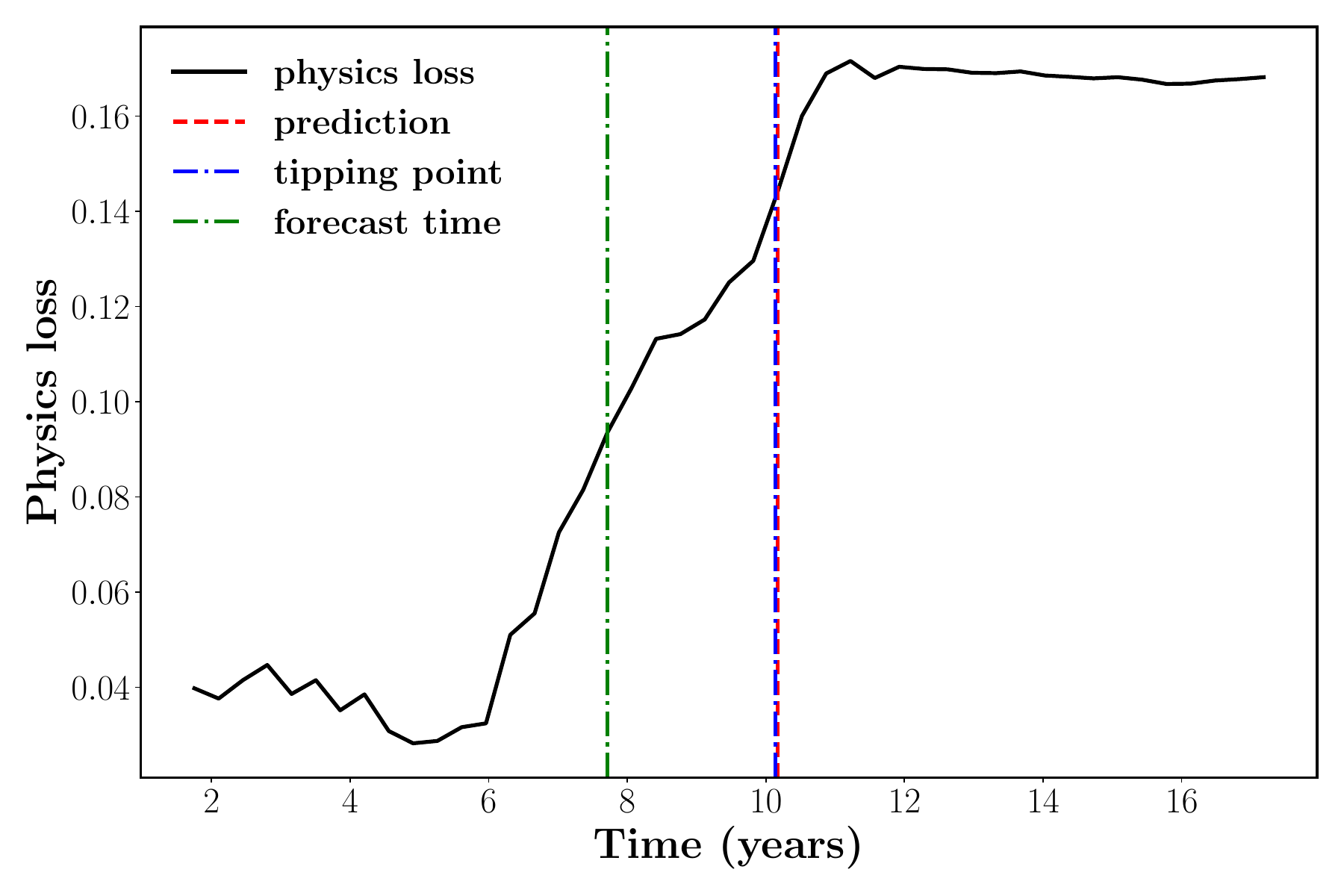}
            \caption{Tipping point forecast w.r.t. partial physics loss}
            \label{fig:rno_clouds_case2}
        \end{subfigure}%
    \caption{For the cloud cover model, using only a mass conservation constraint (as opposed to the full system), \RNO is still capable of identifying the true tipping point of the system with an error of $0.03$ years, predicting $T = 2.45$ years ahead, at $\alpha = 0.07$.}
    \label{fig:tipping_clouds}
\end{figure}

We now consider a case study of tipping point forecasting in the Earth's climate. We also demonstrate that our method is still accurate under \emph{partial physics}, e.g., conservation laws (not the full governing equations). We use the ODE model developed in \citet{Singer2023a, Singer2023b} as an idealization of the physical climate system; this model exhibits a tipping point. The model represents a stratocumulus-topped atmospheric boundary layer, which experiences rapid loss of cloud cover under high CO$_2$ concentrations and shows hysteresis behavior, where the original state is not immediately recovered when CO$_2$ concentrations are lowered. This model was developed, in part, to help explain the results found from the high-resolution simulations in \citet{schneider2019possible, Schneider2020}. More details on the model can be found in Appendix~\ref{sec:cloud_cover_model}, and details on numerical experiments can be found in Appendix~\ref{appdx:cloud_cover_experiment_details}.

We train an \RNO to forecast the non-stationary evolution of the system with a data loss given by Eq.~\ref{eq:loss}, which incorporates the $L^2$ loss across longer time horizons. We compute the empirical \CDF of the physics loss for the cloud cover system (analogous to Eq.~\ref{eq:ks_phys_loss} in the KS setting) as a function of model predictions and CO$_2$. However, in this case, the physics loss is normalized component-wise by the state of the system at a given time, since the magnitudes of each component in this model differ by several orders.

For tipping point forecasting, we base the physics loss \emph{solely on mass conservation} (Eq.~\ref{eq:zi}), instead of the full ODE. Specifically, we let $\ell_P$ be the $L^2$ norm of the difference between the left and right-hand-sides of Eq.~\ref{eq:zi}, which are both functions of time within a time interval. In our experiments, we set the tipping point forecasting horizon to be $T = 7 \cdot \Delta T = 2.45$ years (with forecasting intervals of $\Delta T = 0.36$, or 128 days). We found that the choice of $\Delta T$ (on time-scales on the order of $O(100)$ days) does not significantly impact the \RNO's ability to learn the dynamics. $T$ is chosen to be an order of magnitude larger than $\Delta T$, to demonstrate that our framework is capable of forecasting tipping points at time-scales much longer than $\Delta T$. The results on a given test trajectory can be seen in Figure~\ref{fig:tipping_clouds}. Despite the abruptness of the tipping point and the highly-nonlinear dynamics of the system, our forecasting framework is successfully able to identify the true tipping point in this system using a physics loss derived from a partial picture of the full physics.

\subsection{Forecasting airfoil wake and stall transitions}
\label{subsec:airfoil_results}

To demonstrate the scalability of our method to complex higher-dimensional settings, we consider the problem of forecasting two tipping points in a 2-d flow over an airfoil with angle of attack that increases at a constant rate over time. We simulated the flow past an airfoil of chord length $C$ with velocity $U_\infty$ by solving the 2-d incompressible Navier-Stokes equation using a large-eddy simulation (LES) solver (more details can be found in Appendix~\ref{appdx:airfoil_dataset} and in \cite{leung2025smart}). Two distinct tipping points in the lift coefficients can be observed as a function of the angle of attack (alternatively, as a function of time, since the angle of attack increases at a constant rate over time). The first tipping point observable in the lift coefficient is known as the critical bifurcation from singular-vortex shedding to vortex-pair shedding, which we refer to simply as the wake transition \cite{gupta2023two}. The second tipping point corresponds to the static aerodynamic stall, which is characterized by a rapid drop in the lift coefficient for a small increase in the angle of attack \cite{mccroskey1981phenomenon}. At low Reynolds numbers (Re), however, this drop tends to occur more gradually at angles of attack around 15 degrees \cite{kurtulus2015unsteady}.

In our experiments, we consider three settings: (1) a Re of 1000, where only the static stall occurs, (2) the wake transition in a flow with Re 5000, and (3) static stall in a flow with Re 5000. Creating these three setups allows us to create a direct progression in the difficulty of each subtask. For instance, we would expect a priori that forecasting (1) is easier than forecasting (2) or (3) because the Re 1000 dynamics are less turbulent than the Re 5000 dynamics. Similarly, we would a priori expect that (3) is more challenging than (2) because it requires forecasting one tipping point after another and thus would require a model that has learned dynamics under multiple regimes.

We first train three separate \RNO models to forecast the non-stationary pre-tipping evolution of systems (1), (2), and (3) using the data loss Eq.~\ref{eq:loss}. The tipping point forecasting methodology is identical to Section~\ref{sec:cloud_cover_exp}, except the physics loss is now defined by the divergence-free condition of incompressible Navier-Stokes $\ell_P(u) = \| \nabla \cdot u \|_2$, where $u$ is the velocity field. We note that this physical constraint assumes only minimal physical knowledge of the system.

In our experiments, we also compare the tipping point forecasting performance using \RNO, \MNO, and \RNN. The full results for these settings can be found in Appendix~\ref{appdx:additional_results_airfoil}, and experiments exploring robustness to different rates of change of the airfoil angle of attack, as well as non-tipping settings, can be found in Appendix~\ref{appdx:rate_induced_no_tip}. We find that \RNO produces the most accurate forecasts for each setting, over various choices of $\alpha$. However, \MNO and \RNN are often still able to forecast the transitions, suggesting that our method is  robust to the quality of the temporal evolution model.

\begin{figure}[t]
    \centering
        \begin{subfigure}{0.32\textwidth}
            \centering
            \includegraphics[width=\textwidth]{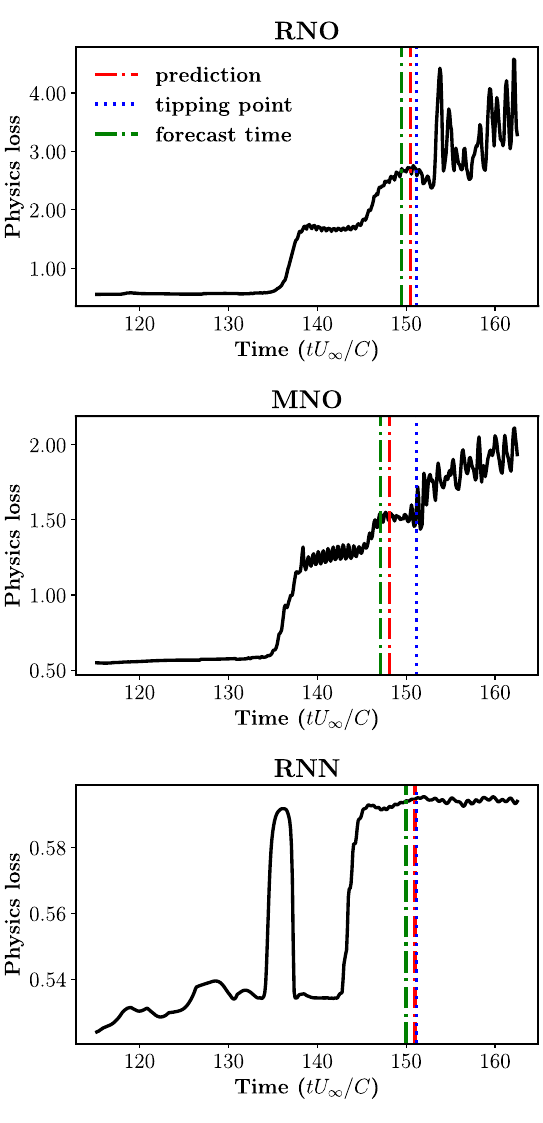}
            \caption{\begin{minipage}{0.7\linewidth}\centering Forecast Re 5000 stall, \\ trained pre-wake Re 5000 \end{minipage}}
            \label{fig:summary_re5000_stall_trained_on_pre_wake}
        \end{subfigure}%
        \begin{subfigure}{0.32\textwidth}
            \centering
            \includegraphics[width=\textwidth]{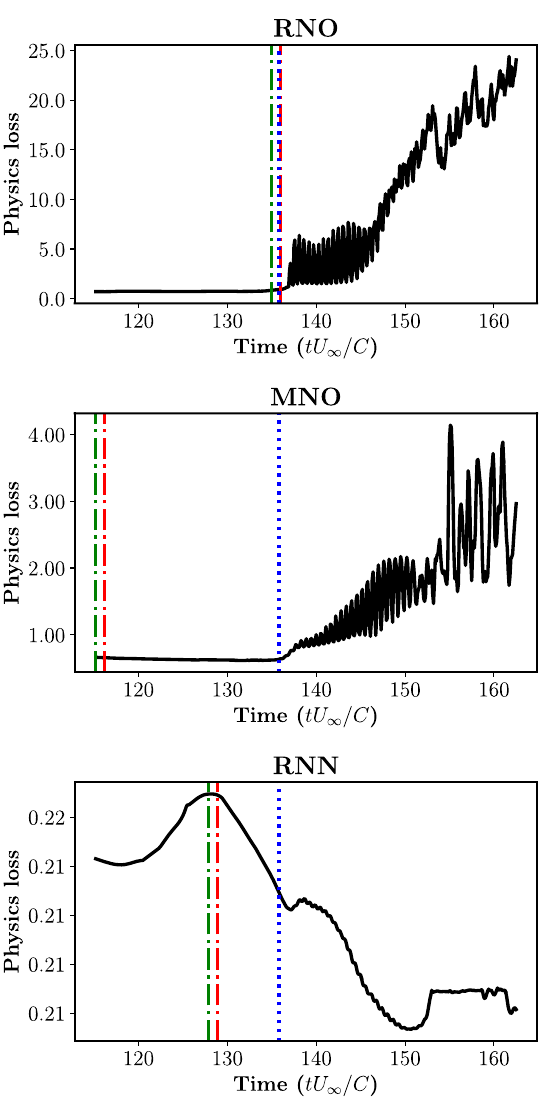}
            \caption{\begin{minipage}{0.7\linewidth}\centering Forecast Re 5000 wake, \\ trained pre-stall Re 1000 \end{minipage}}
            \label{fig:summary_re5000_wake_re1000_transfer}
        \end{subfigure}%
        \begin{subfigure}{0.32\textwidth}
            \centering
            \includegraphics[width=\textwidth]{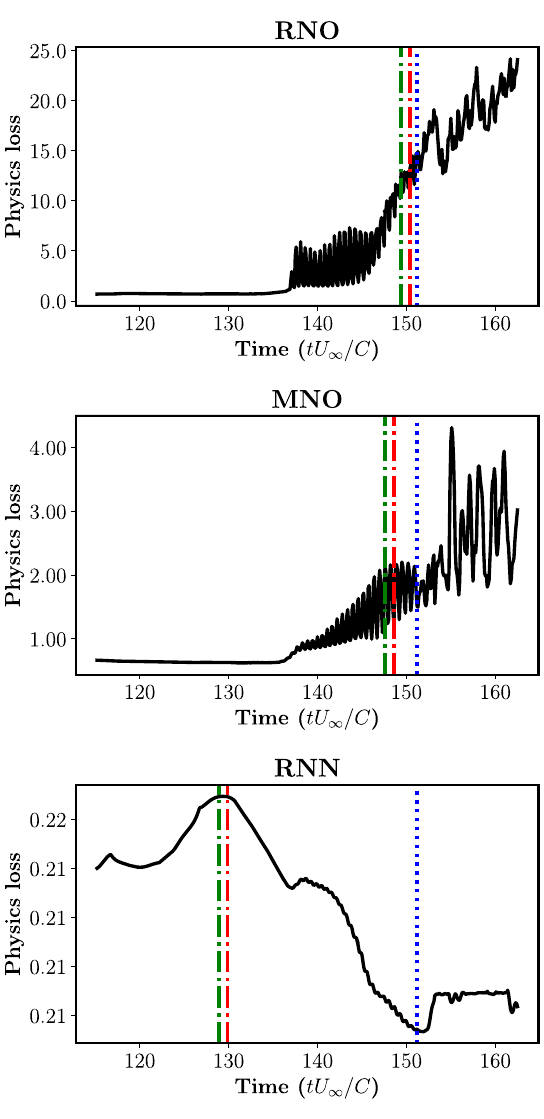}
            \caption{\begin{minipage}{0.7\linewidth}\centering Forecast Re 5000 stall, \\ trained pre-stall Re 1000 \end{minipage}}
            \label{fig:summary_re5000_stall_re1000_transfer}
        \end{subfigure}%
    \caption{Tipping point forecast comparison between \RNO, \MNO, and \RNN in three generalization settings for the flow over the airfoil setting. (a) considers forecasting the stall transition in a Re 5000 flow with the dynamics model trained on pre-wake Re 5000 data. (b) considers forecasting the wake transition in a Re 5000 flow with the dynamics model trained on pre-stall Re 1000 data. (c) considers forecasting the stall transition in a Re 5000 flow with the dynamics model trained on pre-stall Re 1000 data. (a) and (c) use $\alpha = 0.11$, and (b) uses $\alpha = 0.14$.}
    \label{fig:summary_generalization}
\end{figure}

These results motivate the evaluation of the generalizability of our methodology to future tipping points. Figure~\ref{fig:summary_re5000_stall_trained_on_pre_wake} shows the results of the \RNO, \MNO, and \RNN trained on pre-wake data for Re 5000 but used to forecast the stall tipping point for Re 5000 at inference time. Once again, we observe that the stall point is mostly correctly forecasted by all three models.

Lastly, we evaluate the performance of our methods on generalizing to forecasting tipping points at different Reynolds numbers. Specifically, we train each of the three models on Re 1000 pre-stall data and use these models to forecast the wake and stall tipping points in the Re 5000 setting. Figure~\ref{fig:summary_re5000_wake_re1000_transfer} shows the results of \RNO, \MNO, and \RNN trained on Re 1000 pre-stall dynamics and used directly to forecast the Re 5000 wake tipping point. Figure~\ref{fig:summary_re5000_stall_re1000_transfer} shows the analogous results for the Re 5000 stall tipping point. The results show that \RNO far outperforms \MNO and \RNN in forecasting the tipping points. This suggests that the generalization ability of our tipping point forecasting method depends on the accuracy of the underlying dynamics model.

\subsection{Discussion}
From the experiments above (and those in the appendix), we can extract several key takeaways about our proposed tipping point forecasting method:
\begin{enumerate}
    \item Tipping point forecasting remains robust under approximate physical constraints (Appendix~\ref{sec:tipping_lorenz_approx_physics}).
    \item Tipping point forecasting remains robust using a physics loss based on partial physics laws (i.e., not the full governing equations) and conservation laws (Sections~\ref{sec:cloud_cover_exp} and \ref{subsec:airfoil_results}).
    \item Our proposed method is easily applicable to large-scale systems with complex tipping points, even when knowledge of the full governing equations is not known (Section~\ref{subsec:airfoil_results}).
    \item Our proposed method generalizes to forecasting tipping points occurring later in time and across different regimes than seen during training (e.g., different Reynolds numbers for the airfoil experiment), although the robustness depends on the generalizability of the models used to learn the non-stationary dynamics (Section~\ref{subsec:airfoil_results}).
\end{enumerate}

\section{Conclusion}
In this paper, we introduce recurrent neural operator (\RNO), an extension of recurrent neural networks to function spaces, and an extension of neural operators~\citep{kovachki2021neural} to systems with memory. We also propose a model-agnostic conformal prediction method with considerable statistical guarantees for forecasting tipping points in non-stationary systems by measuring deviations in a model's violations of underlying physical constraints or equations. In our experiments, we demonstrate \RNO{}'s ability to learn non-stationary dynamical systems, particularly its stability in error even when auto-regressively applied for long time scales. We also demonstrate the effectiveness of our tipping point forecasting method on infinite-dimensional \PDE systems, and we demonstrate that our proposed methodology maintains good performance even when using partial or approximate physics laws.

\section*{Acknowledgements}
We thank Tapio Schneider for the helpful discussion. M. Liu-Schiaffini was supported in part by the Mellon Mays Undergradaute Fellowship. 
C.E. Singer was supported by the generosity of Eric and Wendy Schmidt by recommendation of the Schmidt Futures program and by Charles Trimble.
A. Anandkumar is supported in part by Bren endowed chair.
We gratefully acknowledge the funding provided for the airfoil research by the SciAI Center, supported by the Office of Naval Research (ONR) under grant number N00014-23-1-2729, the Carver Mead New Adventures Fund, and the Center for Autonomous Systems and Technologies at the California Institute of Technology. Computational time for running the airfoil stall simulation was provided by the Discover project at Pittsburgh Supercomputing Center through allocation PHY240020 from the Advanced Cyberinfrastructure Coordination Ecosystem: Services \& Support (ACCESS) program, which is supported by NSF grants No.~2138259, No.~2138286, No.~2138307, No.~2137603, and No.~2138296. We gratefully acknowledge the funding provided for this research by the Keck Scholar-Fellow Bridge Initiative from the W. M. Keck Foundation, the SciAI Center, supported by the Office of Naval Research (ONR) under grant number N00014-23-1-2729, the Carver Mead New Adventures Fund, and the Center for Autonomous Systems and Technologies at the California Institute of Technology. Computational time was provided by the Discover project at Pittsburgh Supercomputing Center through allocation PHY240020 from the Advanced Cyberinfrastructure Coordination Ecosystem: Services \& Support (ACCESS) program, which is supported by NSF grants No.~2138259, No.~2138286, No.~2138307, No.~2137603, and No.~2138296.

\newpage
\bibliographystyle{unsrtnat}

\bibliography{main}

\newpage

\appendix

\section{Lorenz-63 experimental results}
\label{appdx:lorenz}

We consider tipping point forecasting in the non-stationary chaotic Lorenz-63 system~\citep{lorenz1963deterministic}, a simplified model for atmospheric dynamics. The non-stationary Lorenz-63 system is given by
\begin{equation}
    \label{eq:L63}
    \dot{u}_x = \sigma(u_y-u_x), \qquad
    \dot{u}_y = u_x(\rho(t) -u_z)-u_y,\qquad
    \dot{u}_z = u_x u_y - \beta u_z,
\end{equation}
where the state space is $u = (u_x, u_y, u_z)$, and the parameters of the system are $\sigma, \rho(t), \beta$, where $\rho$ depends on time as in \citep{patel2022using}.

We find that \RNO substantially outperforms \MNO and \RNN in learning the evolution of the non-stationary system. Furthermore, we use the Lorenz-63 system as a case study to justify our tipping point forecasting framework. Figure~\ref{fig:tipping_pred} shows that \RNO is capable of accurately forecasting the tipping point $64$ seconds ($T = 200 \cdot \Delta T$) ahead, even though the size of each input/output interval is $\Delta T = 0.32$ seconds.

We also demonstrate that our method is capable of forecasting tipping points far in advance \emph{using only approximate knowledge of the underlying system}. In particular, we perturb the parameters $\sigma, \rho, \beta$ by some fixed quantity $\eta$ and show that the error in the tipping point forecast for the Lorenz-63 system is still less than $1$ second, even when predicting $64$ seconds in advance.

In this paper, we follow the setting of \citet{patel2022using} where $\sigma = 10$, $\beta = 8/3$, and $\rho$ depends on time via the parameterization $\rho(t) = \rho_0 + \rho_1 \exp \left(t/{\gamma} \right)$, where $\rho_0 = 154$, $\rho_1 = 8$, and $\gamma = 100$. 

A tipping point for this system occurs at approximately $\rho^* = 166$ \citep{manneville1979intermittency}, which corresponds approximately to time $t^* = 40$ under the parameterization above. This tipping point is induced by an intermittency route to chaos \citep{manneville1979intermittency} as the dynamics transition from periodic $t < t^*$ to chaotic $t > t^*$. The tipping point can be classified as an instance of bifurcation-induced tipping \citep{ashwin2012tipping} at a saddle-node bifurcation \citep{patel2022using}. Details on data generation and experimental setup can be found in Appendix~\ref{appdx:numerical_details}.

\subsection{Learning non-stationary dynamics}
\label{sec:exp_nonstat_dynamics}

Numerical results from our experiments are shown in Table~\ref{table:lorenz}. We observe that \RNO outperforms \MNO in relative $L^2$ error by \emph{at least an order of magnitude} for every $n$-step prediction setting up to and including 32-step prediction. We observe that even when composed with itself 32 times, \RNO is capable of maintaining relative $L^2$ error under $0.05$. Further, we find that despite training \MNO with a multi-step procedure up to $M = 12$ steps, this still does not prevent \MNO error from steadily increasing as the number of steps to compose increases. While \RNN is somewhat stable in its error when composed multiple times, \RNO vastly outperforms it in $L^2$ error. We attribute the performance of \RNO to its status as an architectural generalization of \MNO to systems with memory. Further, \RNO{}'s discretization invariance and adaptability to function spaces makes it an improvement over fixed-resolution \RNN{}s.

\subsection{Tipping point forecasting: fully-known physics}
\label{sec:tipping_lorenz}
As described in Section~\ref{sec:dkw}, our conformal prediction approach for tipping point prediction relies on a model trained in pre-tipping dynamics. We then monitor variations in the physics constraint loss of model forecasts to predict tipping points in the future. Recall that the physics constraint loss is only a function of the model predictions $\wh u$ and the time $t$. We define the physics constraint loss to be
\begin{equation}
    \label{eq:lorenz_phys_loss}
    \ell_{P} (\wh u, t) = \left\| \dot{\wh u}(t) - \wh{\dot u}(t) \right\|_2,
\end{equation}
where $\dot{\wh u}$ is the time-derivative\footnote{Note that $\dot{\wh u}(t)$ can in principle be computed to arbitrary precision due to the discretization invariance of \RNO in time. In practice, we find that approximating $\dot{\wh u}(t)$ using finite difference methods is sufficient.} of the model's forecasted trajectory at time $t$ and $\wh{\dot u(t)}$ is the time derivative defined by the Lorenz-63 system (eq.~\ref{eq:L63}) using the model's predicted state $\wh u(t)$ at time $t$. That is, 
\begin{equation}
    \wh{\dot u_x} = \sigma(\wh u_y - \wh u_x), \qquad
    \wh{\dot u_y} = \wh u_x(\rho(t) - \wh u_z) - \wh u_y,\qquad
    \wh{\dot u_z} = \wh u_x  \wh u_y - \beta \wh u_z.
\end{equation}
Thus, $\ell_P$ is minimized when the time derivative of the model's predictions is equal to the expected derivative that all solutions to the Lorenz-63 system must satisfy (Eq.~\ref{eq:L63}). In practice, $\ell_P$ is computed over an entire calibration set, which is unseen during training, and these samples are used to construct an empirical \CDF of the physics loss.

In our experiments, we set $\Delta T = 0.32$ seconds. Here we study the difficult task of forecasting the tipping point $64$ seconds ahead, corresponding to two hundred times the time scale of the model. For a fixed $\delta = 10^{-3}$, Figure~\ref{fig:lorenz_loss_thresh} shows the effect of varying the critical loss threshold $l(\alpha)$ for a given $\alpha$ on \RNO and \MNO tipping point predictions. 

\begin{figure}[t]
    \centering
        \begin{subfigure}{0.5\textwidth}
            \centering
            \includegraphics[width=\textwidth]{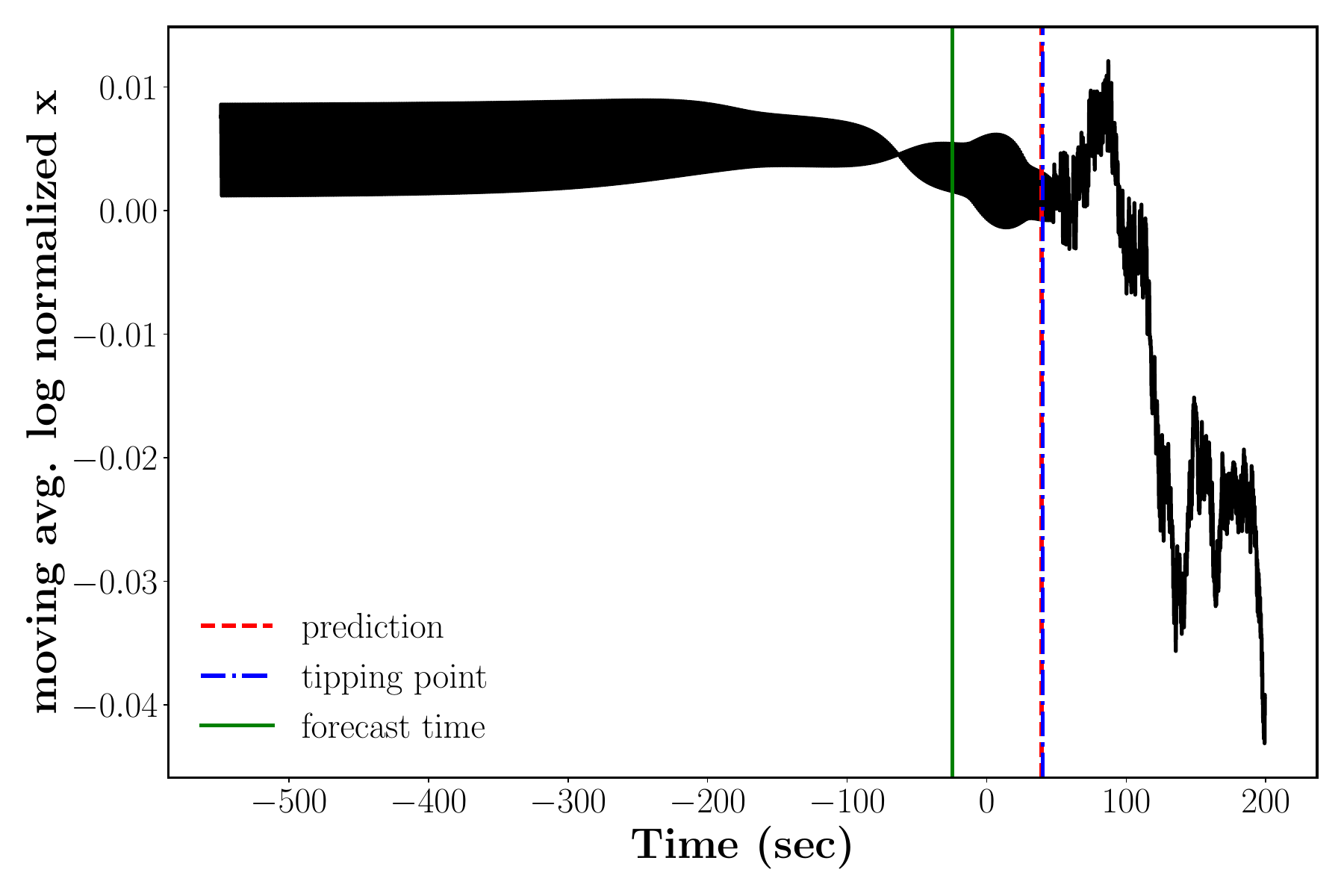}
            \caption{}
            \label{fig:lorenz_x_tipping}
        \end{subfigure}
        \begin{subfigure}{0.5\textwidth}
            \centering
            \includegraphics[width=\textwidth]{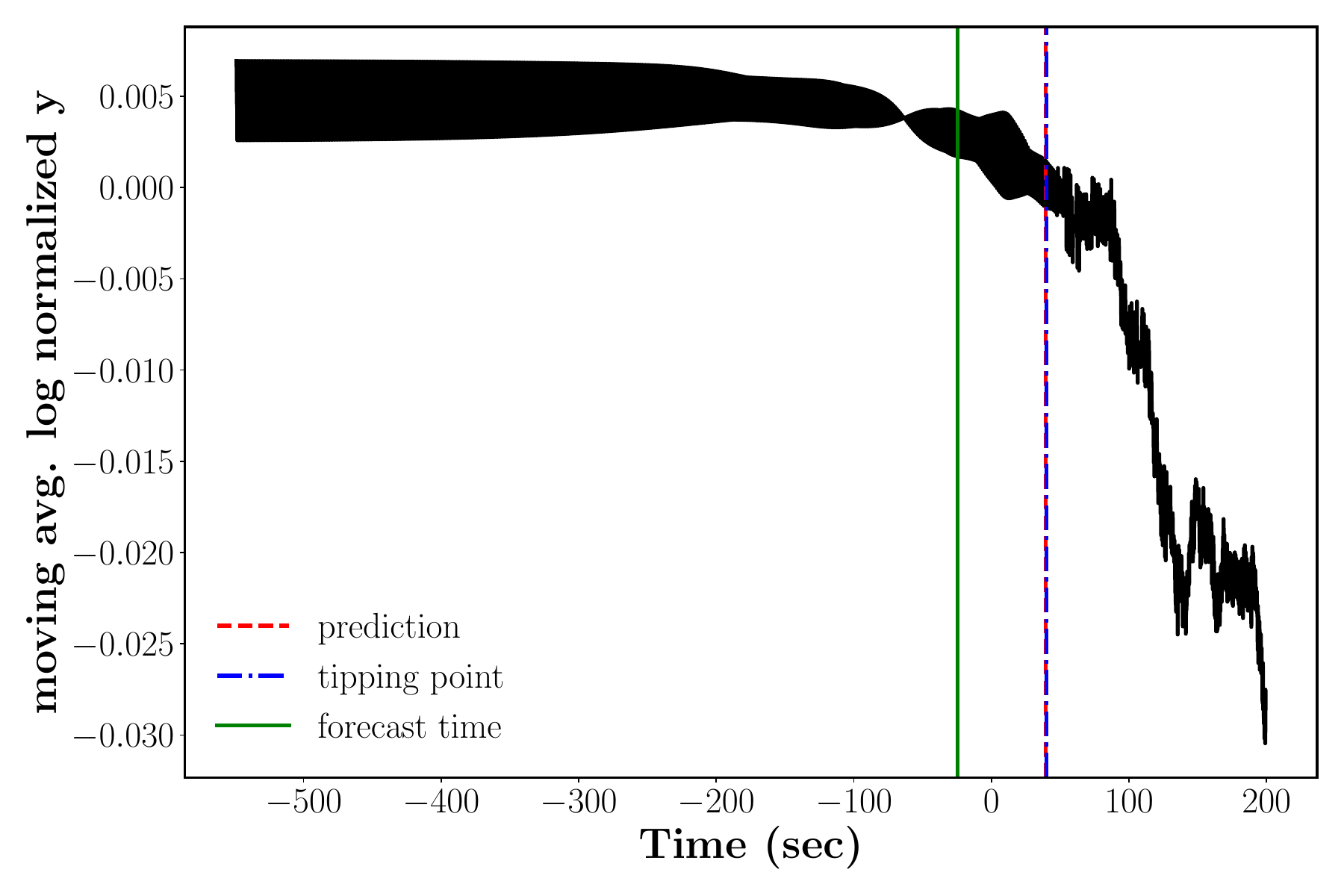}
            \caption{}
            \label{fig:lorenz_y_tipping}
            \end{subfigure}%
            \begin{subfigure}{0.5\textwidth}
            \centering
            \includegraphics[width=\textwidth]{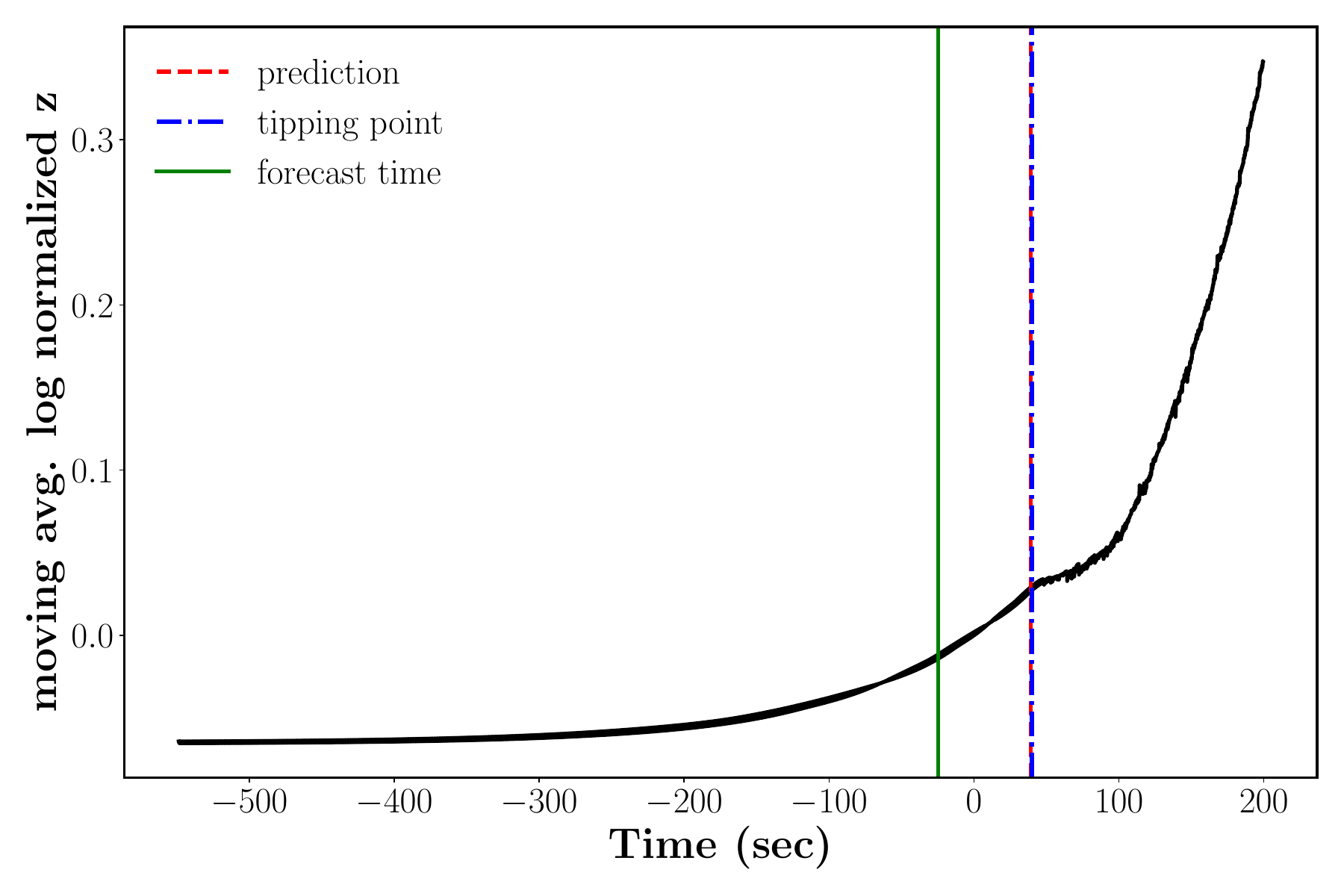}
            \caption{}
            \label{fig:lorenz_z_tipping}
        \end{subfigure}%
    \caption{Log normalized components of the non-stationary Lorenz-63 system plotted with a 50-second moving average. The predicted tipping point aligns nearly exactly with the true periodic-chaotic tipping point of the system, which can be seen as a qualitative change in dynamics of the system components. A false-positive rate of $\alpha = \varepsilon \approx 0.03$ is used, where $\varepsilon$ is discussed in Section~\ref{sec:dkw}.}
    \label{fig:lorenz_tipping_components}
\end{figure}

\begin{figure}[t]
    \centering
        \begin{subfigure}{0.75\textwidth}
            \centering
            \includegraphics[width=\textwidth]{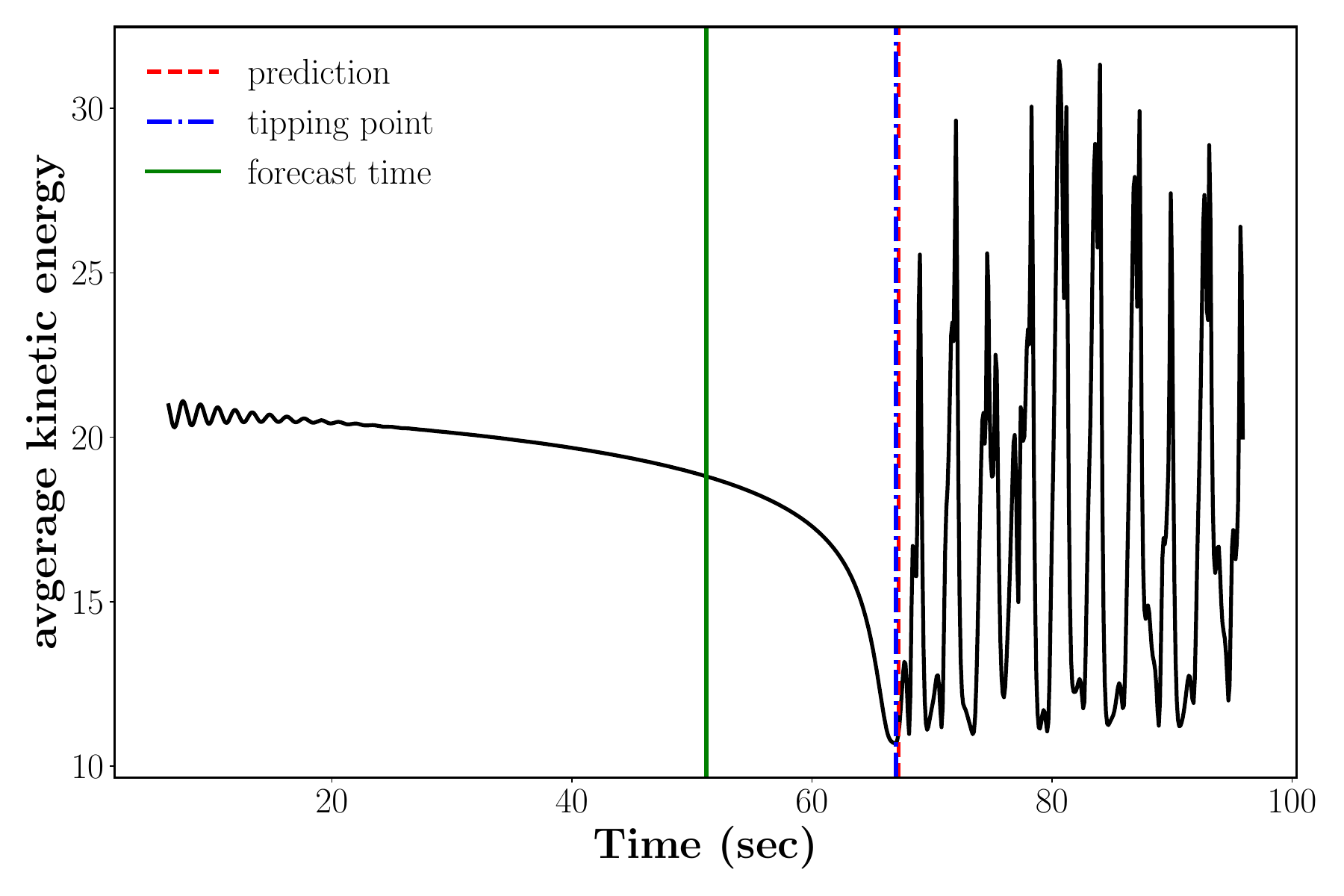}
        \end{subfigure}
        \caption{Kinetic energy of the non-stationary KS system. The predicted tipping point aligns nearly exactly with the true tipping point of the system, which can be seen as a qualitative change in the kinetic energy of the system. The false-positive rate is $\alpha = 0.07$.}
    \label{fig:ks_energy}
\end{figure}

\begin{table*}
\begin{center}
\begin{tabular}{l|cccccccc}
\multicolumn{1}{c}{\bf Model}
&\multicolumn{1}{c}{\textbf{1-step}}
&\multicolumn{1}{c}{\textbf{2-step}}
&\multicolumn{1}{c}{\textbf{4-step}} 
&\multicolumn{1}{c}{\textbf{8-step}}
&\multicolumn{1}{c}{\textbf{16-step}}
&\multicolumn{1}{c}{\textbf{32-step}} 
&\multicolumn{1}{c}{\textbf{64-step}}
&\multicolumn{1}{c}{\textbf{128-step}} 
\\
\hline 
\hline 
\rule{0pt}{1em} \RNO & $\mathbf{0.0053}$ & $\mathbf{0.0033}$ & $\mathbf{0.0027}$ & $\mathbf{0.0028}$ & $\mathbf{0.0068}$ & $\mathbf{0.0434}$ & $\mathbf{0.1549}$ & $\mathbf{0.2585}$ \\
\rule{0pt}{1em} \MNO & $0.0156$ & $0.0242$ & $0.0311$ & $0.0579$ & $0.1035$ & $0.1638$ & $0.2274$ & $0.3114$ \\
\rule{0pt}{1em} \RNN & $0.0141$ & $0.0139$ & $0.0173$ & $0.0150$ & $0.0332$ & $0.1096$ & $0.2366$ & $0.3434$ \\
\hline
\hline 
\end{tabular}
\end{center}
\caption{Relative $L^2$ errors on non-stationary Lorenz-63 for different $n$-step prediction settings. \RNO outperforms \MNO on every forecasting interval and maintains low error even for long intervals.} 
\label{table:lorenz}
\end{table*}  

From Figure~\ref{fig:lorenz_loss_thresh} we make a key observation: the distribution of \RNO{}'s physics loss has both smaller mean and variability than that of \MNO, as well as a shorter right tail. This is deduced from the observation that the first sudden spike in Figure~\ref{fig:lorenz_loss_thresh} corresponds to the mode of the histogram of each model's physics losses over the calibration set. Furthermore, observe that the loss range between this spike and the time when the model achieves its delay in the forecasted tipping point of the smallest magnitude is much smaller for \RNO than for \MNO, implying the difference in the right-tail lengths of the two models. In particular, this analysis of the distribution of the physics loss for each model reinforces the results in Section~\ref{sec:exp_nonstat_dynamics} that \RNO is more successful than \MNO in learning the underlying non-stationary dynamical system. Figures \ref{fig:tipping_pred} and \ref{fig:lorenz_tipping_components} show \RNO achieving near-zero error in forecasting the tipping point $64$ seconds ahead.

\subsection{Tipping point forecasting: approximate physics}
\label{sec:tipping_lorenz_approx_physics}

\begin{figure}[t]
    \centering
        \begin{subfigure}{0.5\textwidth}
            \centering
            \includegraphics[width=\textwidth]{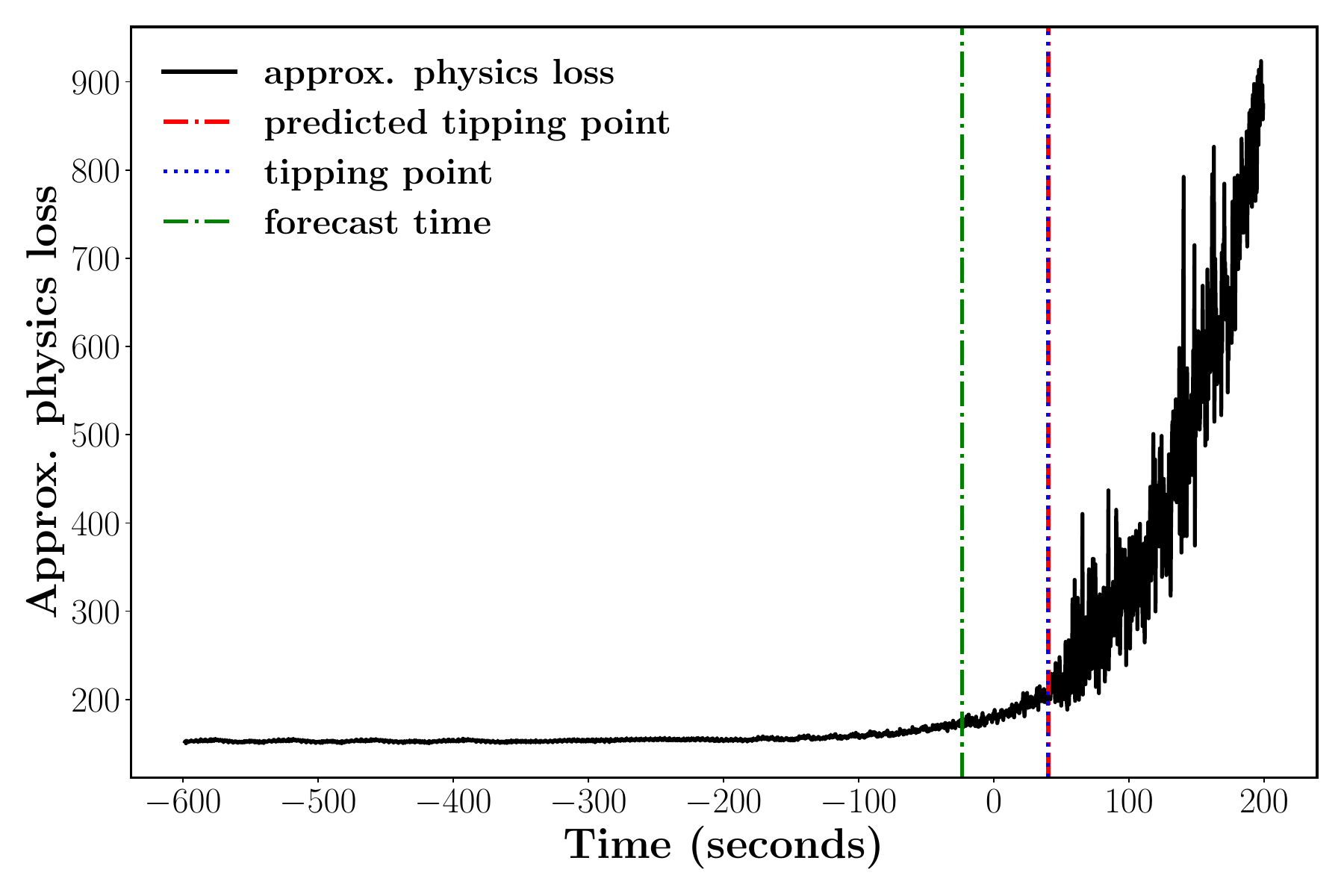}
            \caption{$\eta = 1$}
            \label{fig:lorenz_approx_eta_1}
        \end{subfigure}%
        \begin{subfigure}{0.5\textwidth}
            \centering
            \includegraphics[width=\textwidth]{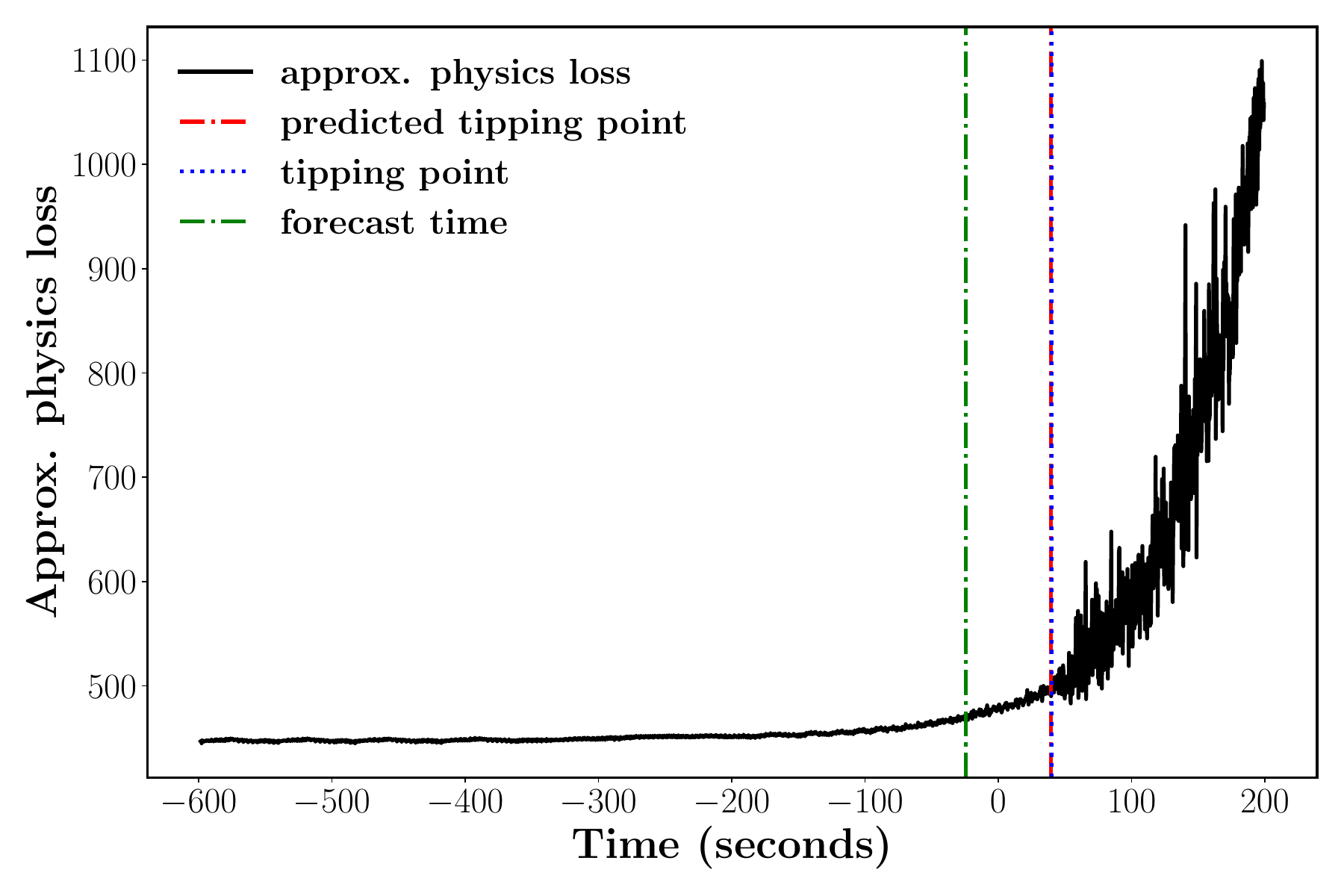}
            \caption{$\eta = 3$}
            \label{fig:lorenz_approx_eta_3}
            \end{subfigure}
            \begin{subfigure}{0.5\textwidth}
            \centering
            \includegraphics[width=\textwidth]{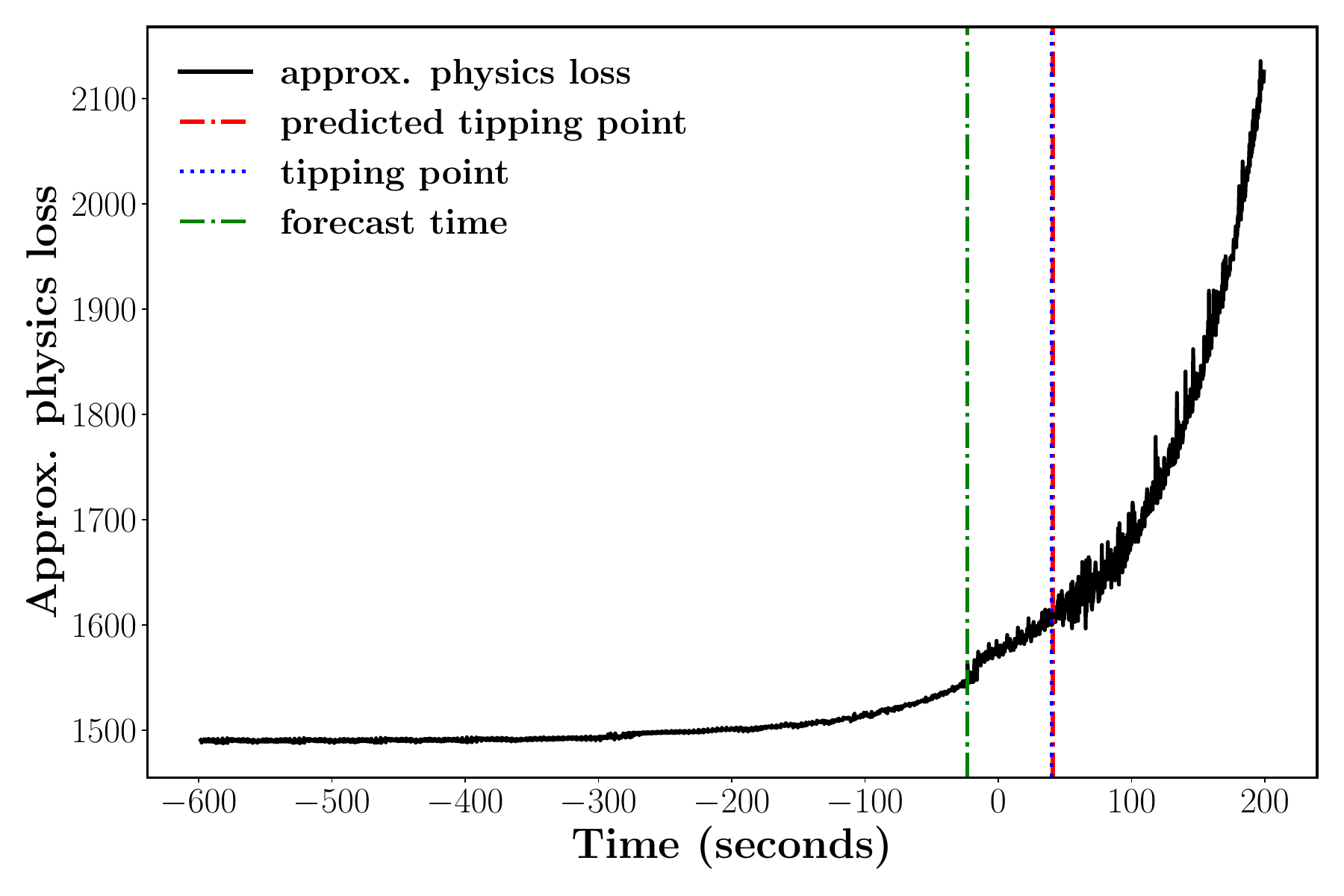}
            \caption{$\eta = 10$}
            \label{fig:lorenz_approx_eta_10}
        \end{subfigure}%
        \begin{subfigure}{0.5\textwidth}
            \centering
            \includegraphics[width=\textwidth]{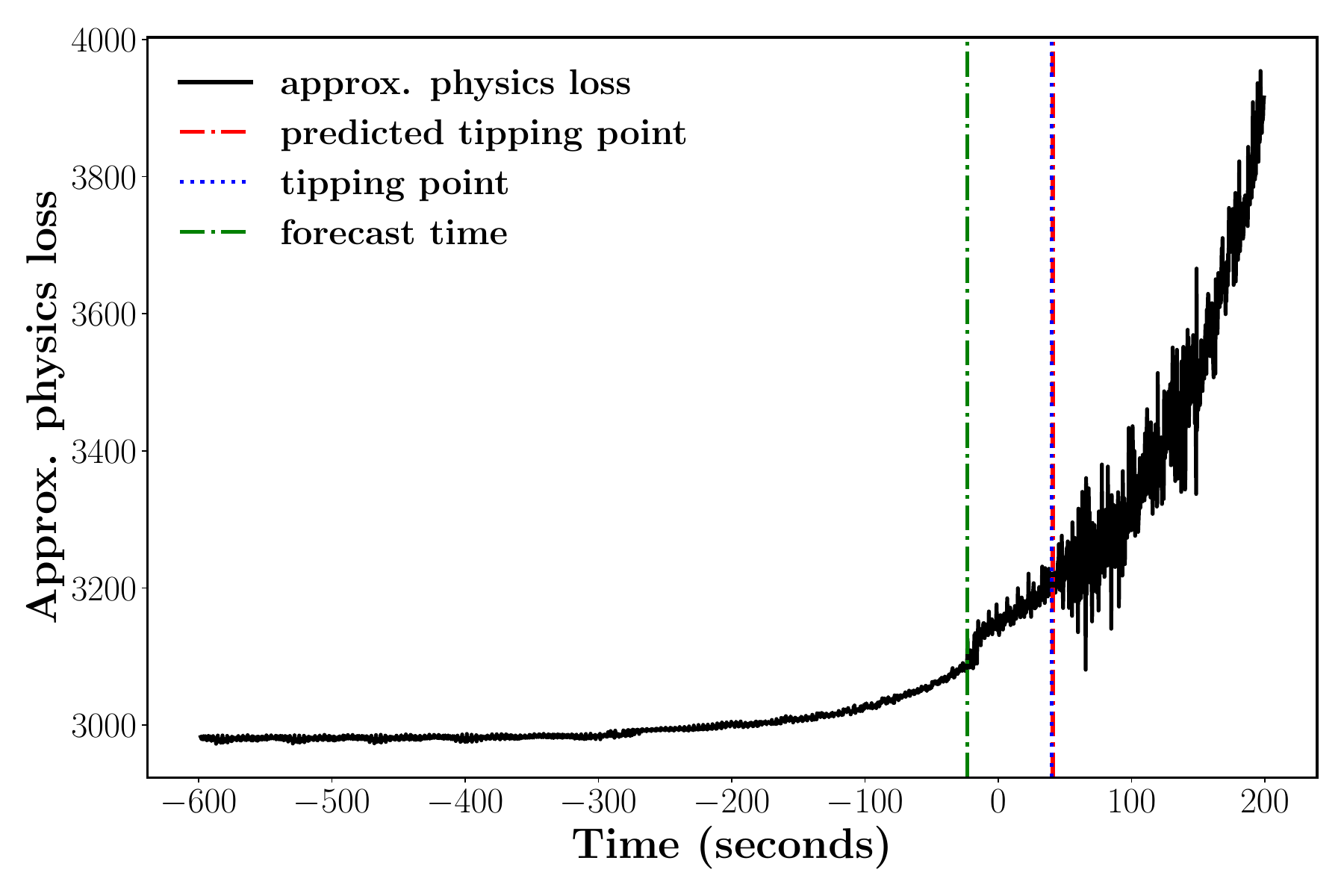}
            \caption{$\eta = 20$}
            \label{fig:lorenz_approx_eta_20}
        \end{subfigure}%
    \caption{Lorenz-63 tipping point forecasting performance $64$ seconds ahead using approximate physical constraints, where $\eta$ is the perturbation on the ODE parameters for the physical law. Our framework is capable of forecasting the tipping points with very low error even with large perturbations $\eta$. A false positive rate of $\alpha = \varepsilon \approx 0.03$ is used, where $\varepsilon$ is discussed in Section~\ref{sec:dkw}.}
    \label{fig:lorenz_tipping_approx_physics}
\end{figure}

In this section, we investigate our method's performance in tipping point forecasting using \emph{approximate physics constraints.} For instance, such a setting may occur when certain coefficients in a \PDE or ODE are known only approximately. In this section, we consider this example of approximate physics for the nonstationary Lorenz-63 system, where we evaluate our proposed framework's capability of forecasting tipping points under different perturbations of the ODE parameters. Specifically, we again define the physics loss as in Eq.~\ref{eq:lorenz_phys_loss}, but in this case, we redefine $\wh{\dot u}(t)$ by replacing $\sigma, \rho(t), \beta$ with $\tilde \sigma = \sigma + \eta, \tilde \rho(t) = \rho(t) + \eta, \tilde \beta = \beta + \eta$, respectively, where $\eta$ is some fixed perturbation.

We use the same experimental setup and as in Section~\ref{sec:tipping_lorenz}. As shown in Fig.~\ref{fig:lorenz_tipping_approx_physics}, our proposed framework is robust to perturbations in the physical constraints and is able to forecast tipping points with very low error far in advance. This is particularly useful for real-world scenarios in which only approximate physical laws are known for a given system.

In particular, we observe that with even approximate physical knowledge of the underlying system, the notion of using the physics loss to verify the correctness of the model's predictions does not break down. This can be observed in Figures~\ref{fig:lorenz_approx_eta_10} and \ref{fig:lorenz_approx_eta_20}, where a ``bump'' in the approximate physics loss can be seen at the forecast time, signaling the shift in distribution of the underlying dynamics.

\section{Non-stationary Kuramoto–Sivashinsky experimental results}
\label{appdx:results_ks}

We demonstrate the effectiveness of \RNO and our tipping point forecasting method in \PDE systems by considering the 1-d non-stationary KS equation, which takes the form
\begin{align}
    \label{eq:KS}
    \frac{\partial u}{\partial t} + u \frac{\partial u}{\partial x} + \frac{\partial^2 u}{\partial x^2} + \kappa(t) \frac{\partial^4 u}{\partial x^4} = 0
\end{align}
where $\kappa(t)$ is some time-dependent parameter of the system, $u(x,t)$ is a function of space $x \in [0,2\pi]$ and time $t \in (0, \infty)$, and the initial condition is $u(\cdot, 0) = u_0$. Following \citet{patel2022using}, we parameterize $\kappa(t) = \kappa_0 + \kappa_1 \exp(t / \gamma)$, where $\kappa_0 =  0.0753$, $\kappa_1 = 0.0034$, and $\gamma = 75.3$. This system exhibits a tipping point near $\kappa^* = 0.08$ from periodic dynamics to chaotic dynamics. The tipping point can be classified as an instance of bifurcation-induced tipping \citep{ashwin2012tipping} at a saddle-node bifurcation \citep{patel2022using}.

Analogously to the Lorenz-63 case (Eq.~\ref{eq:lorenz_phys_loss}), we define the KS physics loss to be
\begin{equation}
    \label{eq:ks_phys_loss}
    \ell_P(\wh u, t) = \left\| \frac{\partial \wh u}{\partial t} + \wh u \frac{\partial \wh u}{\partial x} + \frac{\partial^2 \wh u}{\partial x^2} + \kappa(t) \frac{\partial^4 \wh u}{\partial x^4} \right\|_2.
\end{equation}
In our experiments, we consider the setting of forecasting tipping points far ahead in time, at several orders of magnitude longer time scales than the model is trained on. We compare the performance of \RNO in tipping point forecasting with the performance of \MNO, for varying values of $\alpha$, the false-positive rate of our method. Note that as shown in Table~\ref{table:ks_results}, the relative $L^2$ error of \RNN on forecasting the evolution of the KS equation is an order of magnitude larger than that of \RNO and \MNO.

            %

\begin{figure}[t]
    \centering
        \begin{subfigure}{0.5\textwidth}
            \centering
            \includegraphics[width=\textwidth]{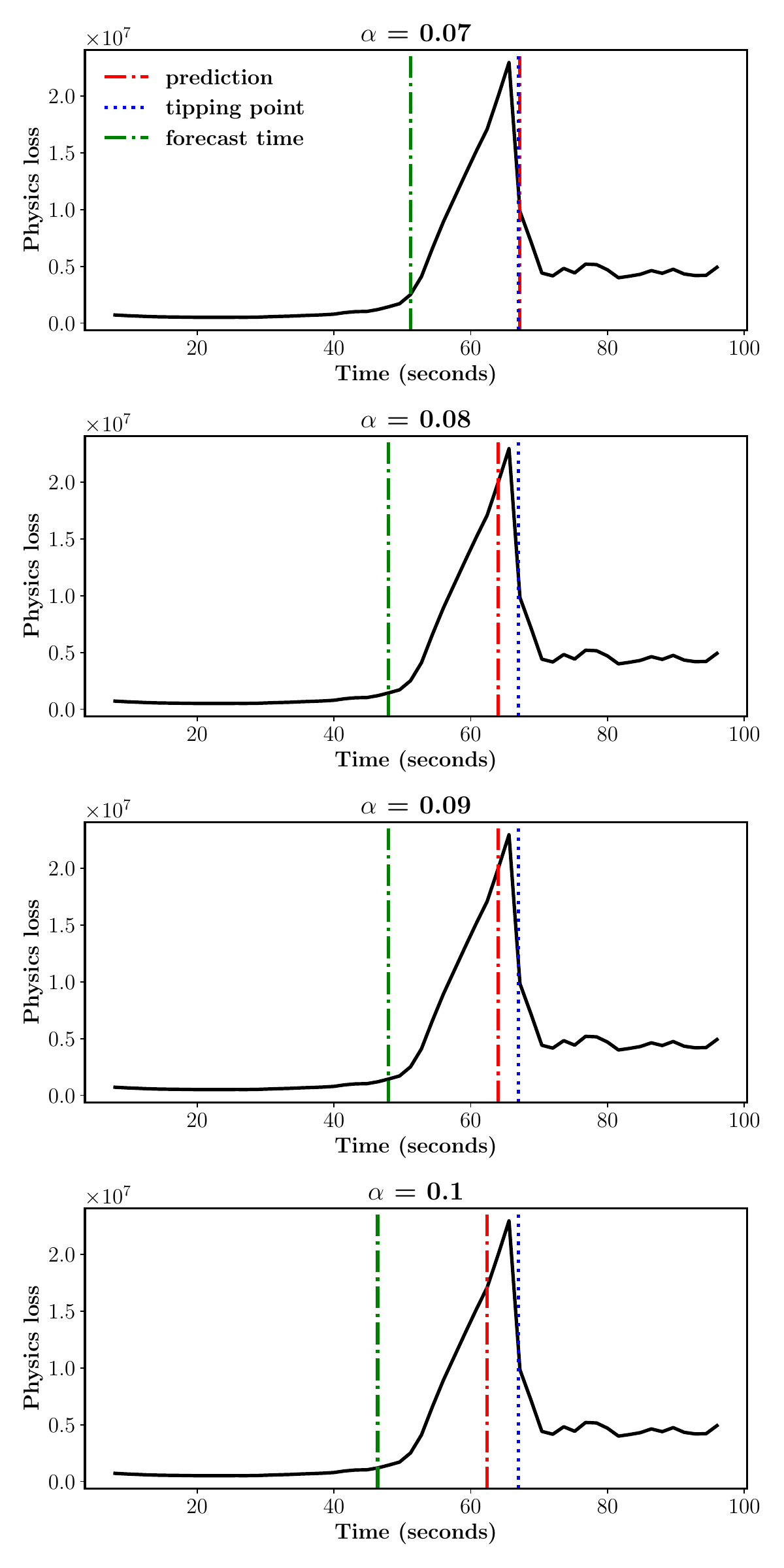}
            \caption{\RNO}
            \label{fig:rno_ks_vary_alpha}
            \end{subfigure}%
            \begin{subfigure}{0.5\textwidth}
            \centering
            \includegraphics[width=\textwidth]{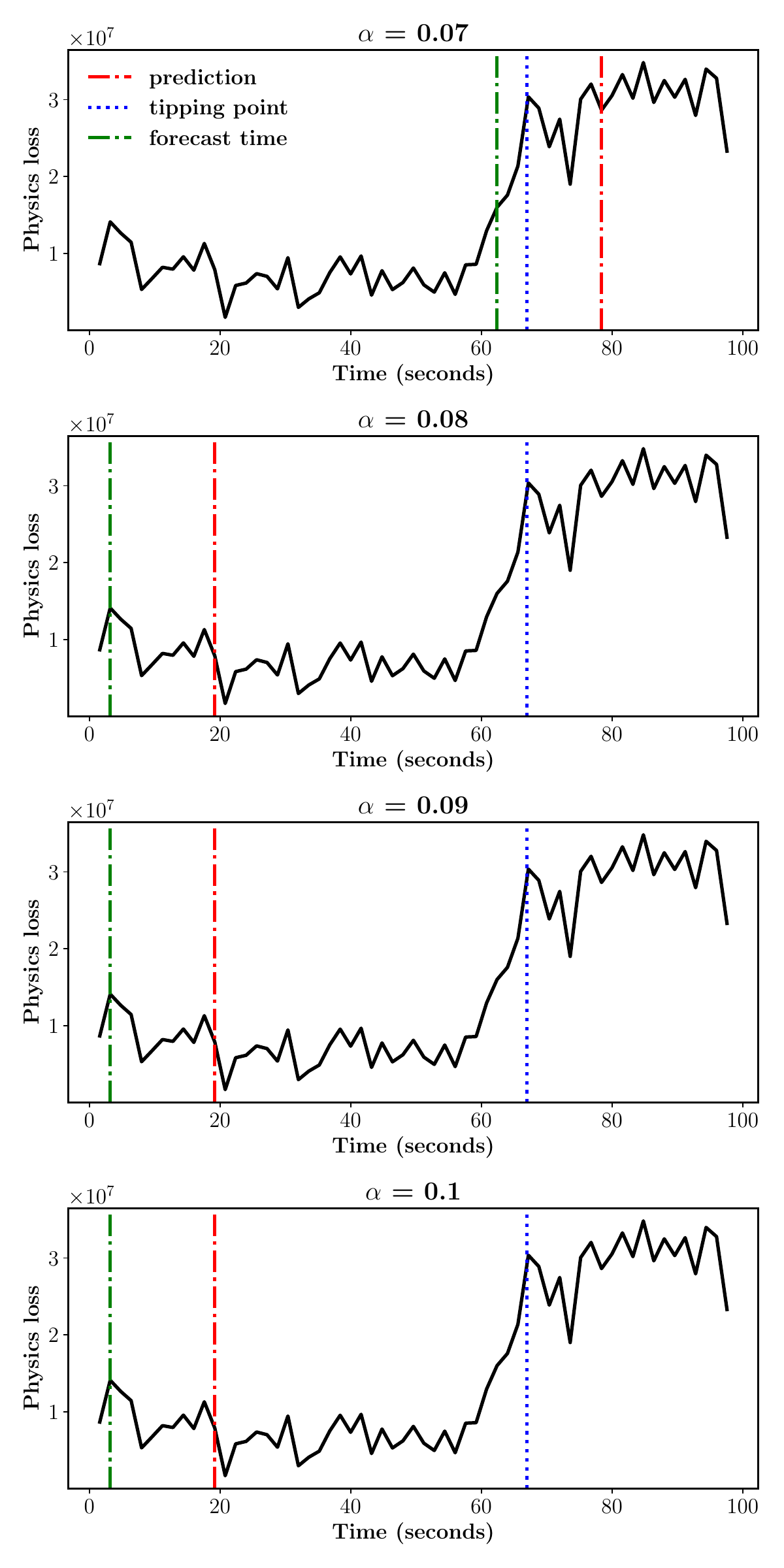}
            \caption{\MNO}
            \label{fig:mno_ks_vary_alpha}
        \end{subfigure}%
    \caption{Comparison of tipping point forecasting performance between \RNO and \MNO for various false-positive rates $\alpha$ for the KS equation, forecasting 16 seconds ahead. For all presented $\alpha$, \RNO outperforms \MNO in tipping point forecasting error.}
    \label{fig:ks_varying_alpha}
\end{figure}

For the KS equation, we set $\Delta T = 1.6$ seconds and seek to forecast the tipping point $T = 16$ seconds ahead. As shown in Figure~\ref{fig:ks_varying_alpha}, we observe that for all values of $\alpha$ shown, \RNO vastly outperforms \MNO in the proximity of tipping point prediction to the true tipping point of the system. At short forecasting intervals (e.g., $T = \Delta T$ or $T = 2 \cdot \Delta T$), \RNO and \MNO both accurately capture the tipping point, but at long time-scales only \RNO retains good performance. This can likely be attributed to the lower performance of \MNO in learning the nonstationary system dynamics (see Table~\ref{table:ks_results}), which causes \MNO to produce fluctuating physics losses even in the pre-tipping regime, whereas \RNO has a low, stable physics loss during pre-tipping.

\begin{figure}[t]
    \centering
        \begin{subfigure}{0.5\textwidth}
            \centering
            \includegraphics[width=\textwidth]{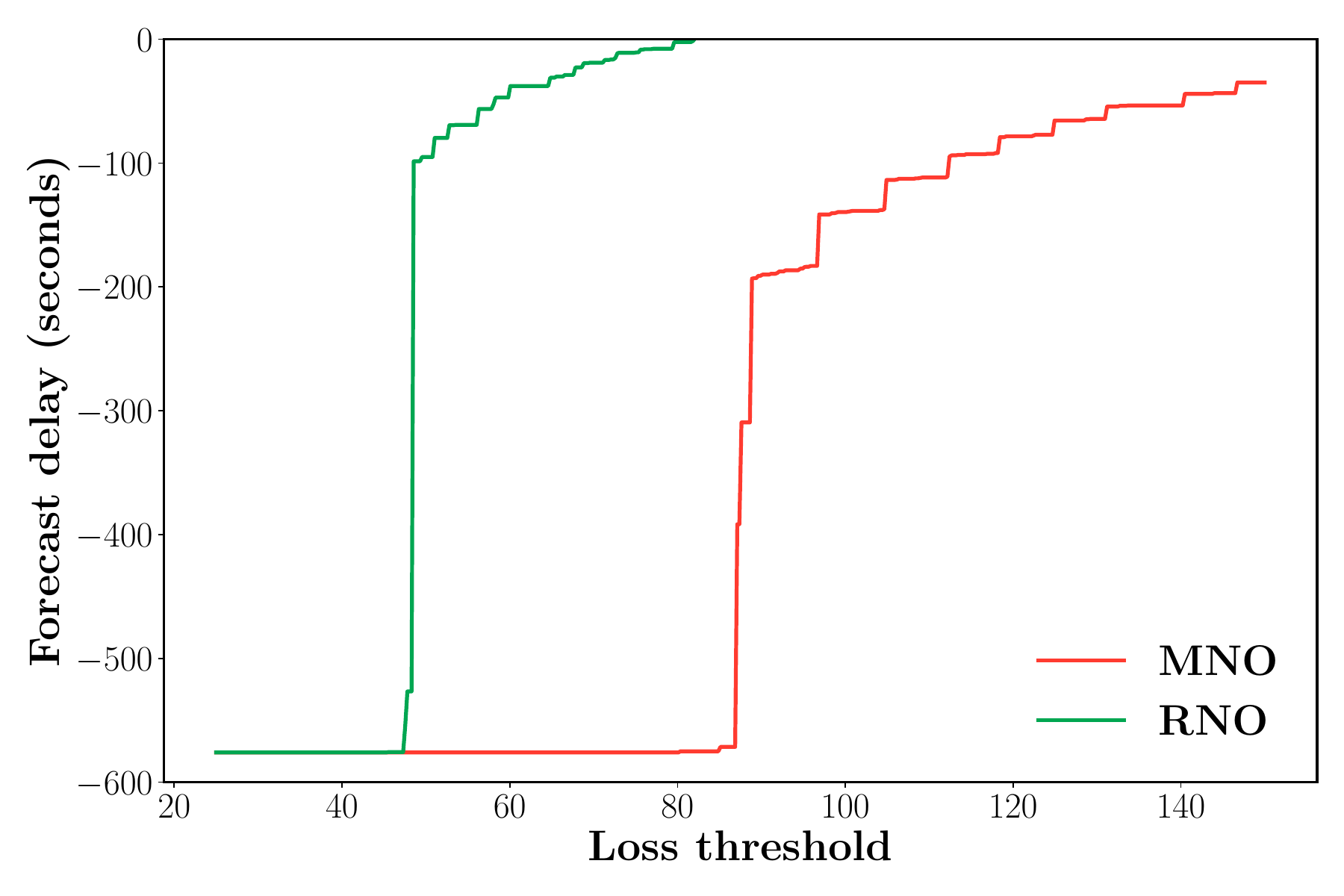}
            \caption{}
            \label{fig:lorenz_loss_thresh}
        \end{subfigure}%
        \begin{subfigure}{0.5\textwidth}
            \centering
            \includegraphics[width=\textwidth]{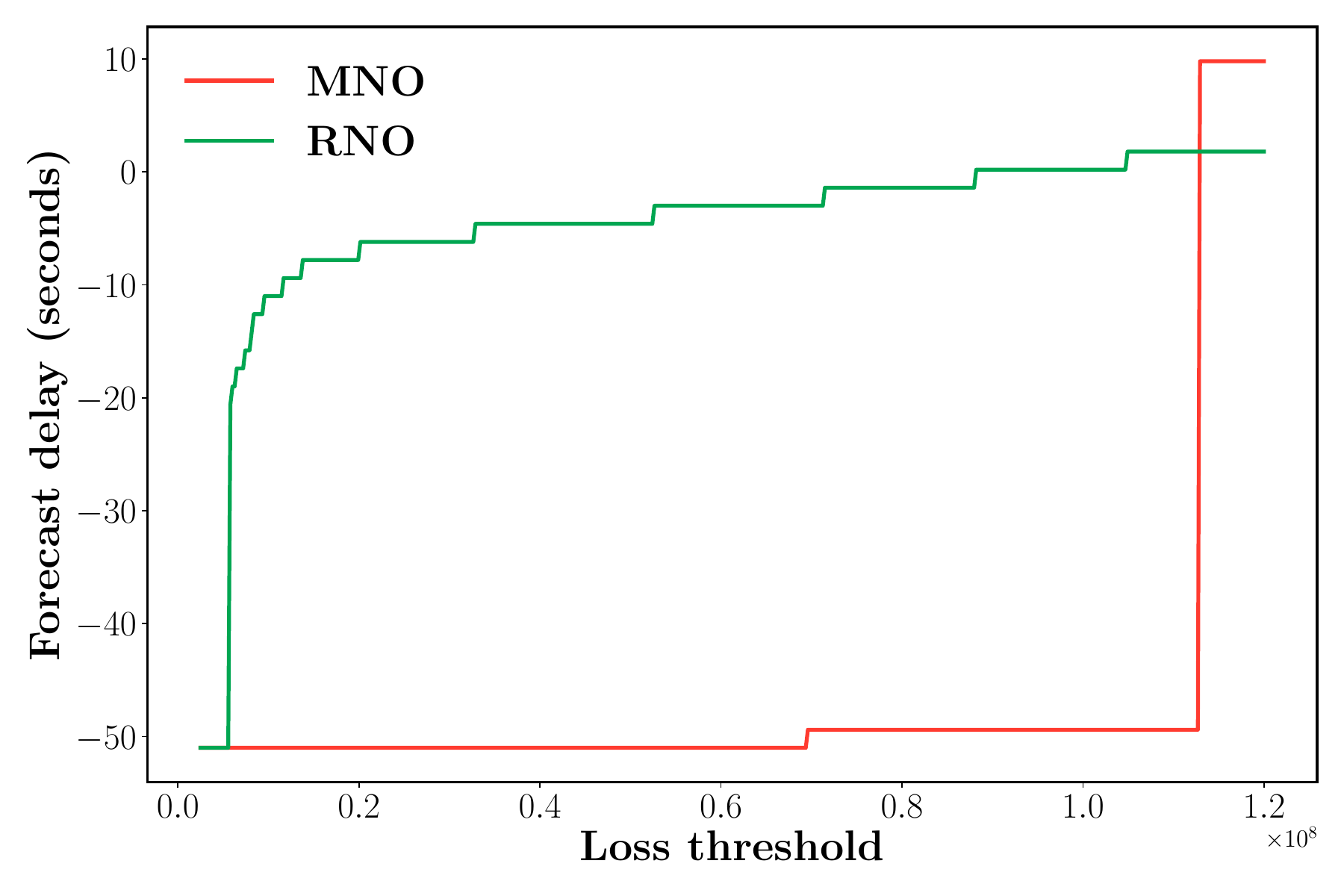}
            \caption{}
            \label{fig:ks_loss_thresh}
        \end{subfigure}
    \caption{ \textbf{(a)} Physics loss threshold comparison between \RNO and \MNO for non-stationary Lorenz-63 system predicting 64 seconds ahead. \textbf{(b)} Physics loss threshold comparison between proposed \RNO and the state-of-the-art baseline \MNO in non-stationary Kuramoto–Sivashinsky system for forecasting the tipping point $16$ seconds ahead. Negative numbers correspond to tipping points predicted early. \RNO consistently outperforms \MNO in tipping point forecasting for a given physics loss threshold, demonstrating that \RNO{} learns the physics of the underlying system better that \MNO.}
    \label{fig:tipping_thresh}
\end{figure}

Figure~\ref{fig:ks_loss_thresh} compares the critical physics loss threshold $l(\alpha)$ for \RNO and \MNO when forecasting the KS tipping point $16$ seconds ahead. For lower physics loss thresholds, \RNO forecasts the tipping point with much better performance than \MNO. As the critical physics loss threshold is increased, the performance of \RNO improves slowly, suggesting that the distribution of \RNO{}'s physics loss has little weight on the right tail. In contrast, \MNO exhibits the opposite behavior, suggesting that it is less reliable than \RNO in learning the non-stationary dynamics. Crucially, we note that comparisons on the accuracy of tipping point predictions between two methods must be made at a fixed false-positive rate $\alpha$ (e.g., see Figure~\ref{fig:ks_varying_alpha}). In Figures~\ref{fig:ks_loss_thresh} and \ref{fig:lorenz_loss_thresh}, we only seek to compare the distributions of physics loss and each model's respective ability to learn the underlying dynamics.

\section{Background on simplified cloud cover model}
\label{sec:cloud_cover_model}
The model developed in \citet{Singer2023a, Singer2023b} is an extension of a traditional bulk boundary layer model \citep{Stevens2006}, which represents the state of the atmospheric boundary layer and cloud by five coupled ordinary differential equations:
\begin{subequations}
\begin{align}
    \frac{dz_i}{dt} &= w_e(\Delta R, s) - D z_i + w_{\mathrm{vent}}(\mathrm{CF}) \label{eq:zi} \\
    \frac{ds}{dt} &= \frac{1}{z_i} \left[ V \cdot (s_0(\mathrm{SST}) - s) + w_e(\Delta R, s) \cdot (s_+ - s) - \Delta R (\mathrm{CO}_2, q_{t,+}) \right] + s_{\mathrm{exp}} \label{eq:sM} \\
    \frac{dq_{t}}{dt} &= \frac{1}{z_i} \left[ V \cdot (q_{t,0}(\mathrm{SST}) - q_{t}) + w_e(\Delta R, s) \cdot (q_{t,+} - q_{t}) \right] + q_{t,\mathrm{exp}} (\mathrm{SST}) \label{eq:qtM} \\
    \frac{d \mathrm{CF}}{dt} &= \frac{f_{\mathrm{CF}}(\mathscr{D}) - \mathrm{CF}}{\tau_{\mathrm{CF}}}. \label{eq:CF} \\
    C \frac{d \mathrm{SST}}{dt} &= \mathrm{SW}_{\mathrm{net}} (\mathrm{CF}) - \mathrm{LW}_{\mathrm{net}} - \mathrm{LHF}(\mathrm{SST}, s, q_t) - \mathrm{SHF}(\mathrm{SST}, s) - \mathrm{OHU} \label{eq:SST}.
\end{align}
\end{subequations}
The model is formulated such that CO$_2$ is the only external parameter and all other processes are represented by physically motivated, empirical formulations, with their parameters based on data from satellite observations or high-resolution simulations.

The first three equations physically represent conservation of mass \eqref{eq:zi}, energy \eqref{eq:sM}, and water \eqref{eq:qtM}, and the final two are equations for the cloud fraction and the sea surface temperature.
In \eqref{eq:zi}, $z_i$ is the depth of the boundary layer [m], $w_e$ is the entrainment rate which is itself parameterized as a function of the radiative cooling and the inversion strength, $D$ is the subsidence rate [s$^{-1}$], $w_{\mathrm{vent}}$ is an additional additive entrainment term used to parameterize ventilation and mixing from overshooting cumulus convective thermals.
In \eqref{eq:sM} and \eqref{eq:qtM}, $s$ is the liquid water static energy [J~kg$^{-1}$] and $q_t$ is the total water specific humidity [kg~kg$^{-1}$].
$V$ is the surface wind speed [m~s$^{-1}$], $\Delta R$ is the cloud-top radiative cooling per unit density [W~m~kg$^{-1}$], which is a function of CO$_2$ and H$_2$O, and $s_{\mathrm{exp}}$ and $q_{t,\mathrm{exp}}$ are export terms representing the effect of large-scale dynamics (synoptic eddies and Hadley circulation) transporting energy and moisture laterally out of the model domain into other regions.

The cloud fraction is modeled as a linear relaxation on timescale $\tau_{\mathrm{CF}}$ to a state $f_{\mathrm{CF}}(\mathscr{D})$ which depends on the degree of decoupling in the boundary layer:
\begin{subequations}
\begin{align}
    f_{\mathrm{CF}}(\mathscr{D}) &= \mathrm{CF}_{\mathrm{max}} - \frac{\mathrm{CF}_{\mathrm{max}} - \mathrm{CF}_{\mathrm{min}}}{1 + \frac{1}{9} \exp(-m(\mathscr{D}-\mathscr{D}_c))} \label{eq:f_CF} \\
    \mathscr{D} &= \left(\frac{\mathrm{LHF}}{\rho \Delta R}\right) \left( \frac{z_i - z_b}{z_i} \right) \label{eq:De}.
\end{align}
\end{subequations}
This parameterization is inspired by theoretical and observational work from \citet{Bretherton1997, Chung2012} and parameters $\mathrm{CF_{max}}$, $\mathrm{CF_{min}}$, and $m$ are fit to data from \citet{Cesana2019-casccad, Zheng2021, schneider2019possible}.
    
Equation \eqref{eq:SST} is the standard surface energy budget equation for SST. 
On the left-hand side, $C = \rho_w c_w H_w$ is a heat capacity per unit area, where $\rho_w$ and $c_w$ are the density and specific heat capacity of water and $H_w$ is the depth of the slab ocean.
The value of $H_w$ is arbitrary (here 1~m): it affects the equilibration time, but not the equilibrium results, which is appropriate given that the forcing is much slower than the equilibration timescale (approx. 50~days).
On the right-hand side are the source terms from shortwave and longwave radiation, latent and sensible heat fluxes, and ocean heat uptake.
Ocean heat uptake is solved for implicitly such that $\mathrm{SST} = 290$~K for $\mathrm{CO}_2 = 400$~ppmv and assumed constant in time.

To generate the data for our experiments, we set the CO$_2$ forcing to follow
\begin{equation}
    \label{eq:noisy_co2}
    \text{CO}_2(t) = c_0 (1 + r)^t + \sigma W(t),
\end{equation}
where $t$ is the time (in years), $W(t)$ is a Wiener process, $c_0$ is the initial CO$_2$ concentration, $r$ is the annual rate of CO$_2$ increase, and $\sigma$ is the scaling parameter for $W(t)$. We use $c_0 = 400$ ppm, $r = 0.1$, and $\sigma = 2.5$.

\section{Details on dataset for airfoil experiments}
\label{appdx:airfoil_dataset}
\begin{figure}[t]
    \centering
    \includegraphics[width=\textwidth]{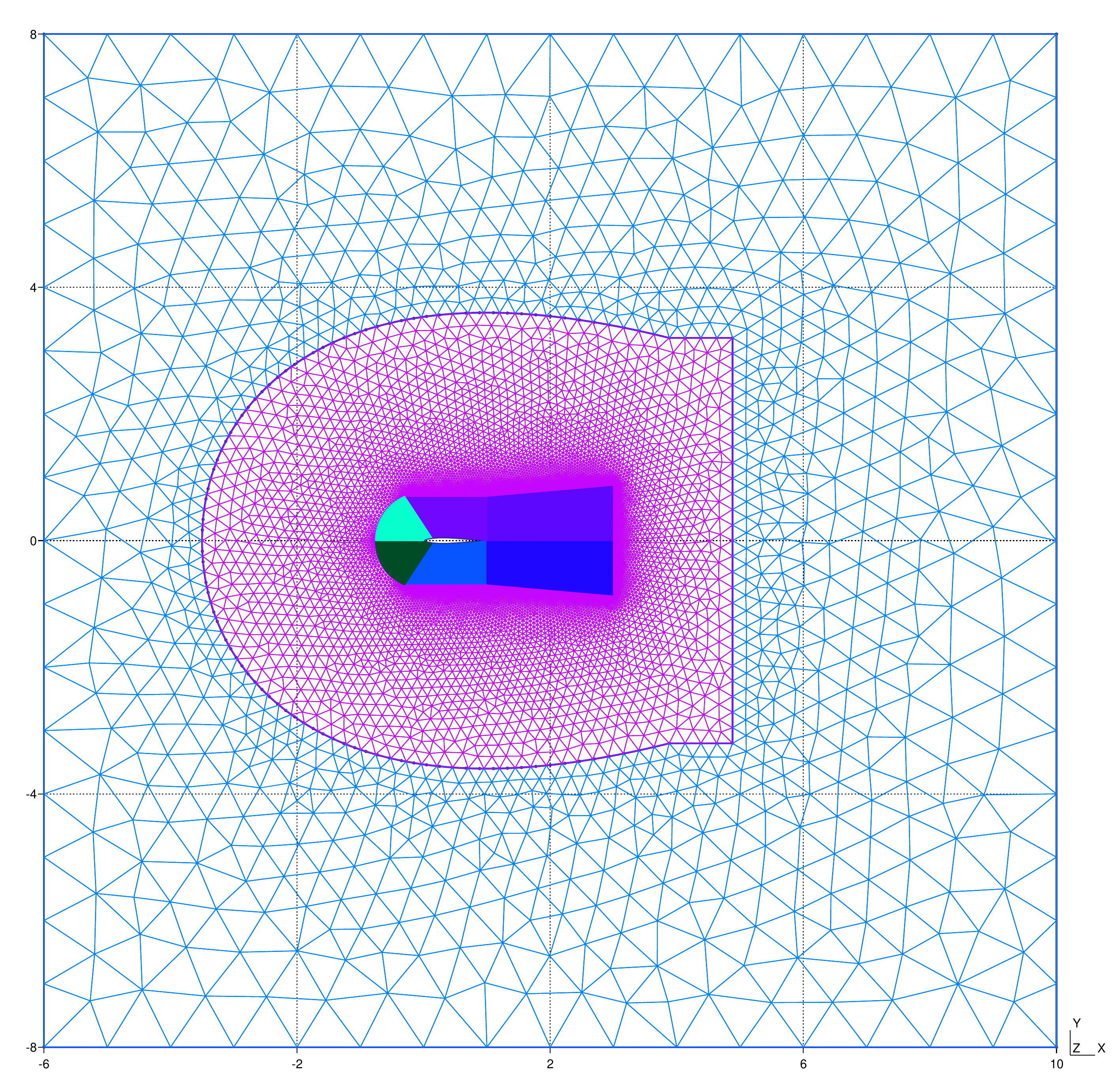}
    \caption{Computational mesh of the airfoil used in the simulations.}
    \label{fig:airfoil_mesh}
\end{figure}

The flow simulation uses a finite-volume, unstructured-mesh large-eddy simulation (LES) solver \citep{you2008discrete}. The spatially-filtered incompressible Navier-Stokes equations are solved with second-order accuracy employing cell-based, low-dissipative, and energy-conservative spatial discretization and a fully-implicit, fractional-step time-advancement method with the Crank–Nicholson scheme. The Poisson equation for pressure is solved using the bi-conjugate gradient stabilized method \citep{van1992bi}. The subgrid-scale stress is modeled using the dynamic Smagorinsky model \citep{germano1991dynamic,lilly1992proposed}.

A NACA 0012 airfoil with chord length $C$ is placed horizontally at the origin of the coordinate system as shown in Figure~\ref{fig:airfoil_mesh}. Boundary conditions for the airfoil surface are defined as solid no-slip walls. The simulation domain is periodic in the spanwise direction. Uniform inflow of $U_\infty$ is set to come in through the left and bottom surfaces of the domain. Additionally, convective outflow boundary conditions are set for both the top and right surfaces of the domain.

The computational meshes used in the simulations consist of approximately 200,000 cells. As illustrated in Figure~\ref{fig:airfoil_mesh}, the airfoil is enclosed by a C-type mesh \citep{anderson1995computational}, which features approximately 250 cells along the airfoil's circumference. The boundary layer on the airfoil is captured by a mesh with $\Delta y_1/C \approx 0.004$, where $\Delta y_1$ is the height of the first wall-normal cell. Since our experiments focus on two-dimensional flow fields, the mesh is extruded in the spanwise direction using only three uniform cells, covering a total non-dimensional spanwise length of $L_z/C = 0.002$. 

The resulting velocity fields sampled on a coarse grid, which spans from -2 to 4 in the $x$-direction and from -2 to 2 in the $y$-direction, with a uniform grid spacing of 0.01. For computational efficiency, we train and evaluate our models on a subset of this physical domain, spanning from 0.05 to 2.5 in the $x$-direction and from -1 to 1.32 in the $y$-direction. The lift force is obtained by numerically integrating the surface pressure and shear stress components normal to the freestream. The lift coefficient is then calculated using the lift force by
\begin{gather}
    C_L = \frac{F_L}{\frac{1}{2} \rho U_\infty^2 L_zC},
\end{gather}
where $F_L$ is the lift force and $\rho$ is the fluid density. 

Ten unique trajectories with different initial conditions are simulated for Reynolds numbers of 1000 and 5000. Each trajectory is generated by continuously increasing the inflow angle of attack (AoA) at a constant rate of 0.33 degrees per time unit ($ tU_{\infty}/C$). The wake transition and stall tipping points are manually identified by examining the time history of the airfoil lift coefficients.

\section{Additional results on the cloud cover system}
\label{appdx:additional_results_clouds}

\subsection{Numerical results in learning non-stationary dynamics}

\begin{table*}
\begin{center}
\begin{tabular}{l|ccccc}
\multicolumn{1}{c}{\bf Model}
&\multicolumn{1}{c}{\textbf{1-step}}
&\multicolumn{1}{c}{\textbf{2-step}}
&\multicolumn{1}{c}{\textbf{4-step}} 
&\multicolumn{1}{c}{\textbf{8-step}}
&\multicolumn{1}{c}{\textbf{16-step}}
\\
\hline 
\hline 
\rule{0pt}{1em} \RNO & $\mathbf{0.0234}$ & $\mathbf{0.0326}$ & $\mathbf{0.0449}$ & $\mathbf{0.0614}$ & $\mathbf{0.0852}$ \\
\rule{0pt}{1em} \MNO & $0.0246$ & $0.0342$ & $0.0472$ & $0.0647$ & $0.0860$ \\
\rule{0pt}{1em} \RNN & $0.0285$ & $0.0360$ & $0.0481$ & $0.0645$ & $0.0868$ \\
\hline
\hline 
\end{tabular}
\end{center}
\caption{Relative $L^2$ errors on the cloud cover system for different $n$-step prediction settings using the experimental setup described in Appendix~\ref{appdx:cloud_cover_experiment_details}. \RNO outperforms \MNO and \RNN on every forecasting interval.} 
\label{table:clouds_numerical_results}
\end{table*}  

Numerical results from our experiments are shown in Table~\ref{table:clouds_numerical_results}. We observe that \RNO outperforms \MNO and \RNN in relative $L^2$ error for every $n$-step prediction presented. We find that as the size of the forecasting interval $\Delta T$ (not to be confused with the length of the forecasting window, which is $n \cdot \Delta T$) increases, the performance of \RNN worsens substantially. We attribute this to the resolution-dependence of \RNN; as the dimensionality of the input (in this case $5 \cdot \Delta T$, since the cloud cover system is 5-dimensional) increases, the size of the \RNN model must be increased accordingly in order to adequately capture the dynamics of the system. For \RNO and \MNO, the size of the models need not be increased substantially since the input is interpreted as a function over a larger domain, and the inductive biases of these models allows them to outperform fixed-resolution models such as \RNN{}s.

\subsection{Tipping point forecasting using full ODE constraint}
\label{appdx:clouds_full_ode_tipping_point_forecast}
\begin{figure}[t]
    \centering
        \begin{subfigure}{0.5\textwidth}
            \centering
            \includegraphics[width=\textwidth]{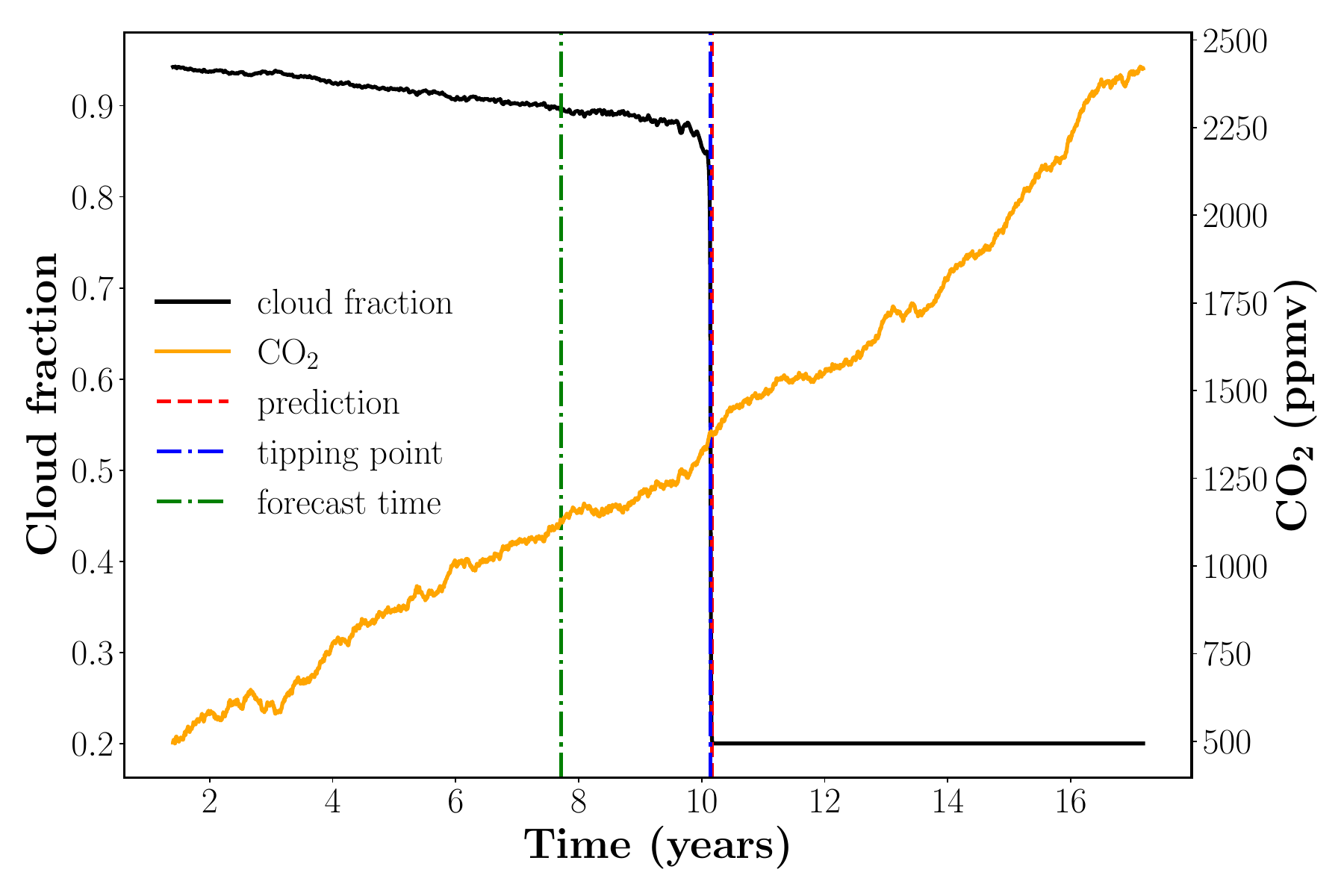}
            \caption{Tipping point forecast w.r.t. cloud fraction}
            \label{fig:rno_clouds_case1_full_ode}
            \end{subfigure}%
            \begin{subfigure}{0.5\textwidth}
            \centering
            \includegraphics[width=\textwidth]{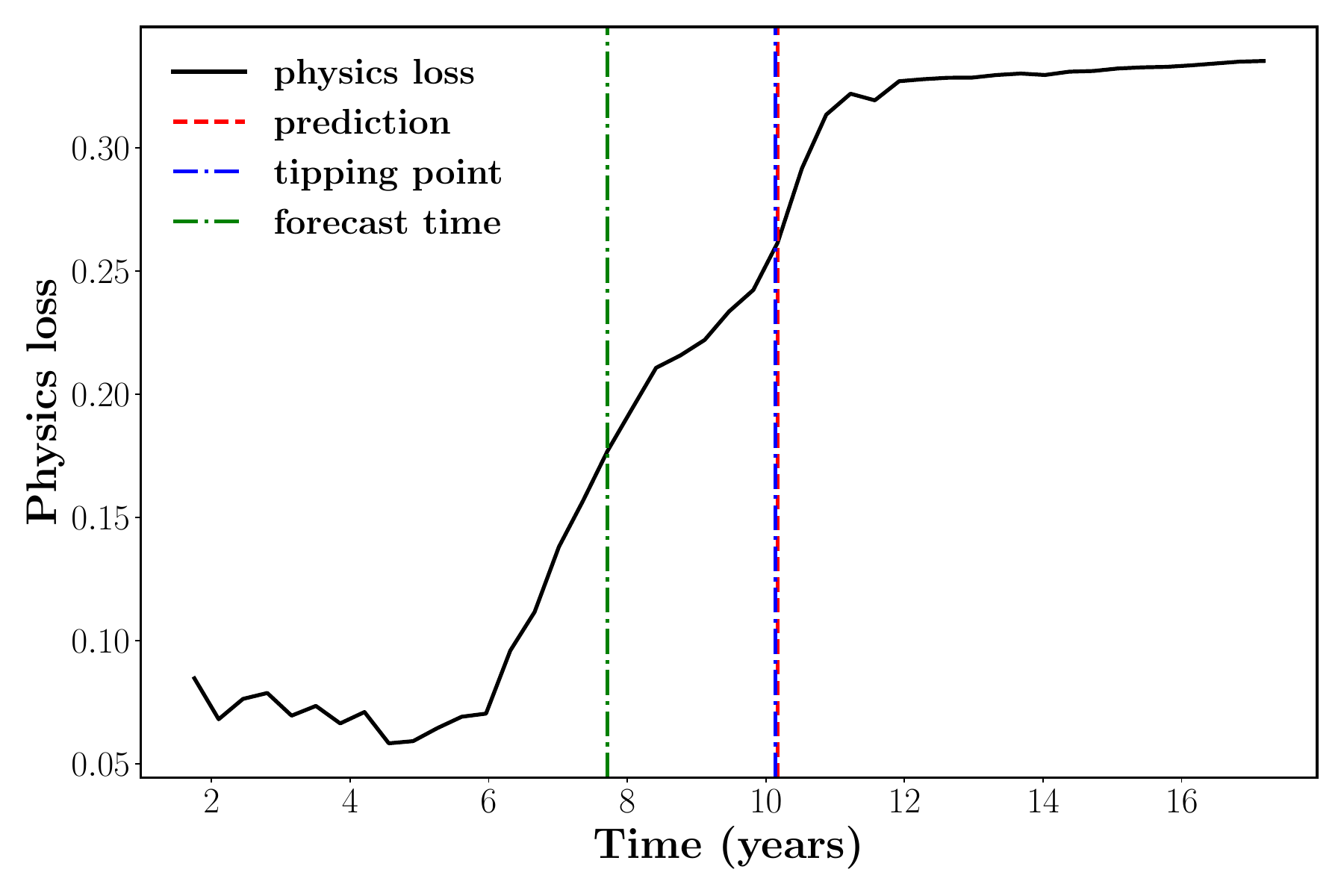}
            \caption{Tipping point forecast w.r.t. partial physics loss}
            \label{fig:rno_clouds_case2_full_ode}
        \end{subfigure}%
    \caption{Using the full ODE as the physical constraint, \RNO is capable of identifying the true tipping point of the simplified cloud cover system with an error of $0.03$ years, predicting $T = 2.45$ years ahead, at $\alpha = 0.08$.}
    \label{fig:tipping_clouds_full_ode}
\end{figure}

In this section, we demonstrate that our proposed method is capable of identifying the tipping point with low error when using the full ODE equation as the physics constraint. Figure~\ref{fig:tipping_clouds_full_ode} shows that our method is capable of forecasting the tipping point far in advance, similarly to Figure~\ref{fig:tipping_clouds}.

\begin{figure}[t]
    \centering
        \begin{subfigure}{0.5\textwidth}
            \centering
            \includegraphics[width=\textwidth]{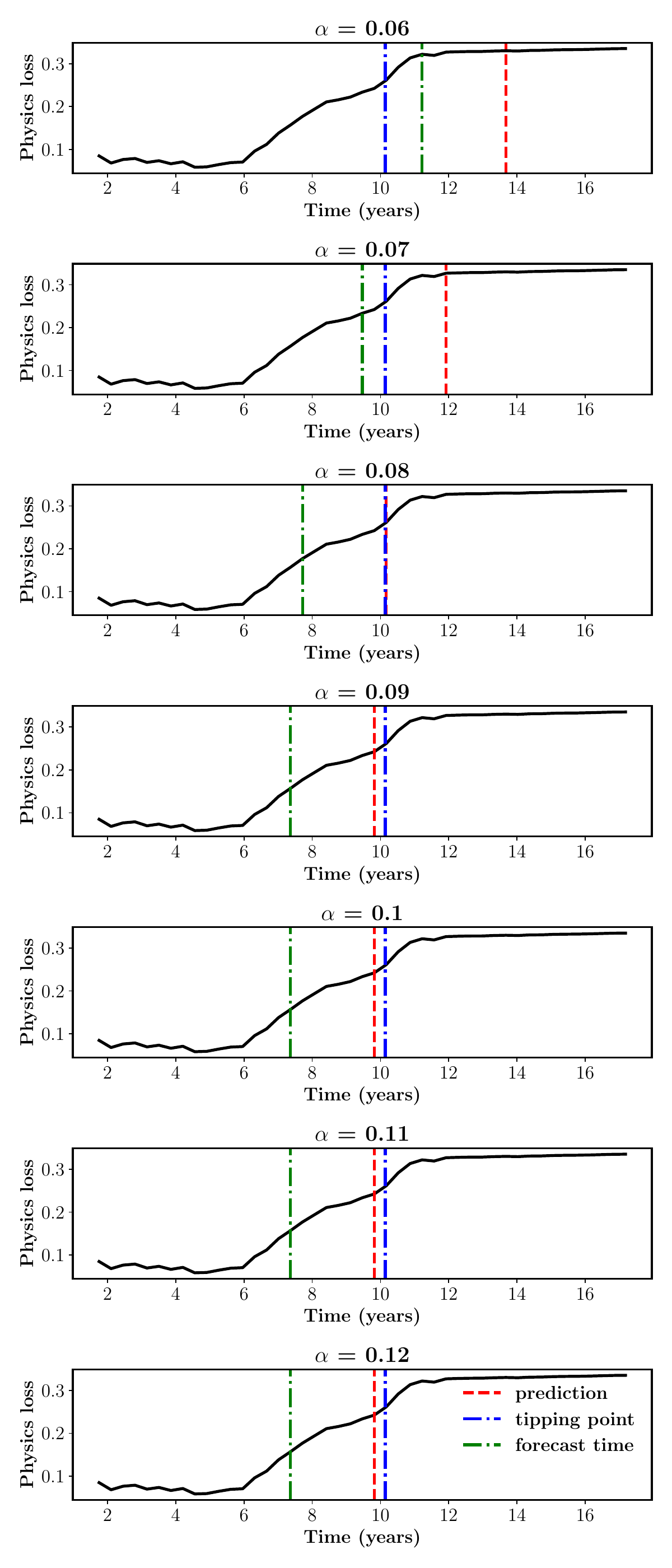}
        \end{subfigure}
        \caption{Tipping point forecasting performance for \RNO for various false-positive rates $\alpha$ in the simplified cloud cover setting. \RNO achieves very low forecasting error for the majority of $\alpha$. Further, \RNO achieves approximately zero error for $\alpha = 0.08$.}
    \label{fig:clouds_varying_alpha_rno}
\end{figure}

\begin{figure}[t]
    \centering
        \begin{subfigure}{0.5\textwidth}
            \centering
            \includegraphics[width=\textwidth]{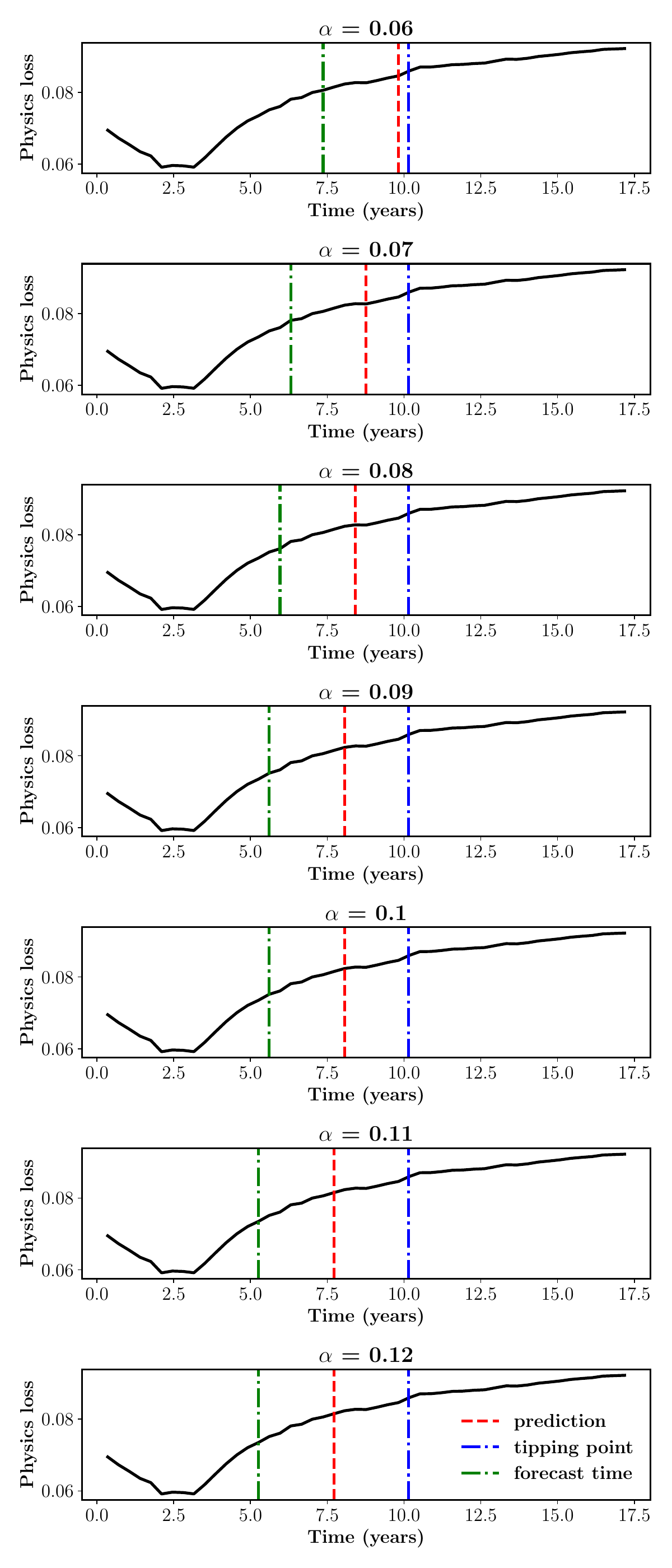}
            \caption{\MNO}
            \label{fig:mno_clouds_vary_alpha}
        \end{subfigure}%
        \begin{subfigure}{0.5\textwidth}
            \centering
            \includegraphics[width=\textwidth]{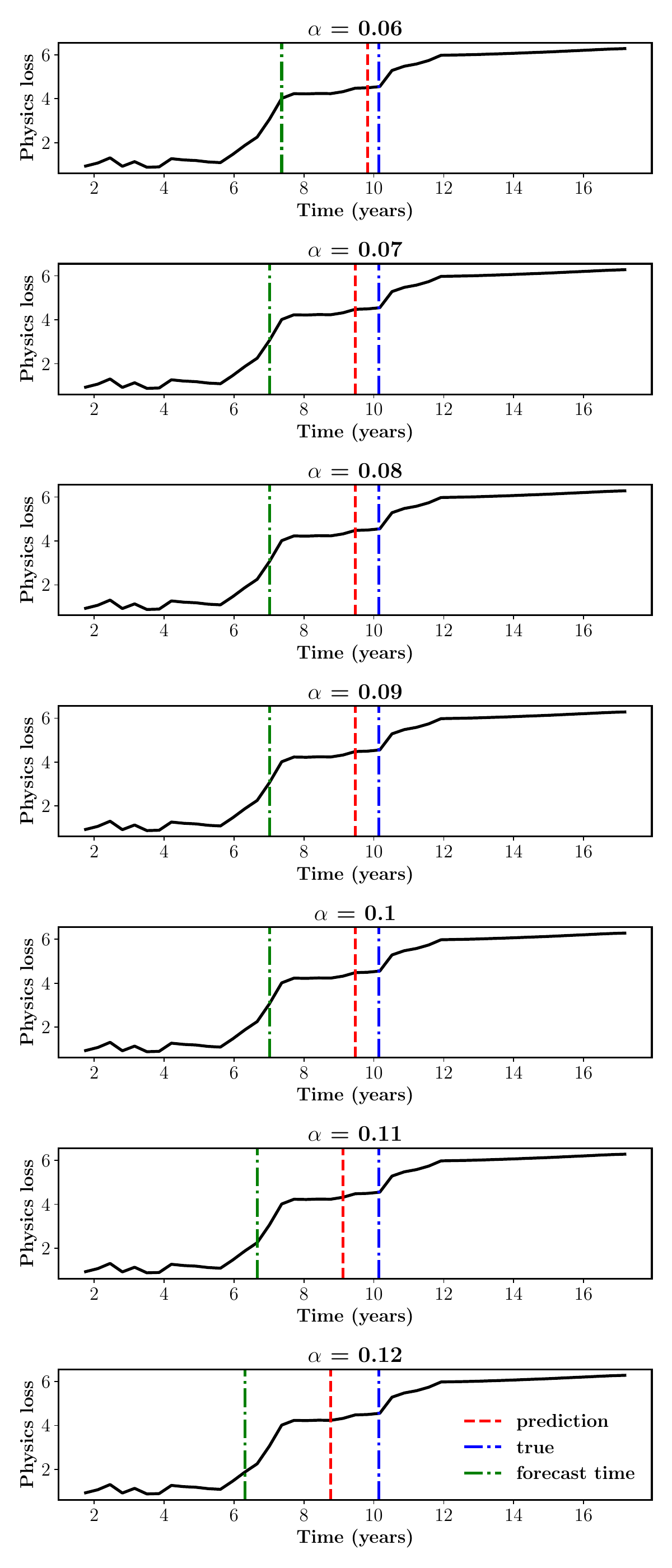}
            \caption{\RNN}
            \label{fig:gru_clouds_vary_alpha}
        \end{subfigure}%
    \caption{Comparison of tipping point forecasting performance between \MNO and \RNN for various false-positive rates $\alpha$ in the simplified cloud cover setting. \RNO (in Figure~\ref{fig:clouds_varying_alpha_rno}) tends to outperform \MNO and \RNN in tipping point forecasting error for the majority of the $\alpha$.}
    \label{fig:clouds_varying_alpha}
\end{figure}

Similarly to the nonstationary KS setting, we also compare the performance of \RNO in tipping point forecasting with \MNO and \RNN for varying values of false-positive rate $\alpha$ for the simplified cloud cover system (Figures~\ref{fig:clouds_varying_alpha_rno} and \ref{fig:clouds_varying_alpha}). We observe that \RNO overall outperforms \RNN and \MNO in tipping point forecasting for a variety of false-positive rates $\alpha$. Furthermore, \RNO is able to achieve nearly no increase in error as $\alpha$ increases, for reasonable $\alpha$.

\section{Additional results on the airfoil tipping points}
\label{appdx:additional_results_airfoil}
As mentioned in Section~\ref{subsec:airfoil_results}, our airfoil experiments involves the three settings: (1) a Reynolds number (Re) of 1000, where only the static stall occurs, (2) the wake transition in a flow with Reynolds number 5000, and (3) stall in a flow with Reynolds number 5000. Each model learns the evolution operator mapping the initial state forward in time by $\Delta T = 0.1 ~tU_\infty/C$. 

Although the main goal of the airfoil experiments is to demonstrate the effectiveness of our proposed tipping points forecasting method, we present the $L^2$ relative test error for these experiments for completeness. The results for \RNN, \MNO, and \RNO on the pre-tipping Re 1000 stall, Re 5000 wake, and Re 5000 stall settings can be found in Tables~\ref{table:re1000_results}, \ref{table:re5000_wake_results}, and \ref{table:re5000_stall_results}, respectively. We present results for long-time rollouts for each model, going up to 25 steps for Re 1000 stall and Re 5000 wake and up to 8 steps for the harder Re 5000 stall setting.

\begin{table*}[h]
\begin{center}
\begin{tabular}{l|ccccc}
\multicolumn{1}{c}{\bf Model} & \multicolumn{1}{c}{\textbf{5-step}} & \multicolumn{1}{c}{\textbf{10-step}} & \multicolumn{1}{c}{\textbf{15-step}} & \multicolumn{1}{c}{\textbf{20-step}} & \multicolumn{1}{c}{\textbf{25-step}} \\
\hline
\hline
\rule{0pt}{1em} \bf RNN & $0.0121$ & $0.0222$ & $0.0326$ & $0.0432$ & $\mathbf{0.0540}$ \\
\rule{0pt}{1em} \bf MNO & $0.0120$ & $0.0220$ & $0.0317$ & $0.0502$ & $0.0603$ \\
\rule{0pt}{1em} \bf RNO & $\mathbf{0.0101}$ & $\mathbf{0.0215}$ & $\mathbf{0.0313}$ & $\mathbf{0.0411}$ & $0.0553$ \\
\hline
\hline
\end{tabular}
\end{center}
\caption{Relative $L^2$ errors for the Re 1000 stall transition.}
\label{table:re1000_results}
\end{table*}

\begin{table*}[t]
\begin{center}
\begin{tabular}{l|ccccc}
\multicolumn{1}{c}{\bf Model} & \multicolumn{1}{c}{\textbf{5-step}} & \multicolumn{1}{c}{\textbf{10-step}} & \multicolumn{1}{c}{\textbf{15-step}} & \multicolumn{1}{c}{\textbf{20-step}} & \multicolumn{1}{c}{\textbf{25-step}} \\
\hline
\hline
\rule{0pt}{1em} \bf RNN & $\mathbf{0.0166}$ & $\mathbf{0.0269}$ & $\mathbf{0.0382}$ & $0.0501$ & $0.0621$ \\
\rule{0pt}{1em} \bf MNO & $0.0173$ & $0.0291$ & $0.0385$ & $\mathbf{0.0495}$ & $\mathbf{0.0611}$ \\
\rule{0pt}{1em} \bf RNO & $0.0180$ & $0.0294$ & $0.0388$ & $0.0509$ & $0.0631$ \\
\hline
\hline
\end{tabular}
\end{center}
\caption{Relative $L^2$ errors for the Re 5000 wake transition.}
\label{table:re5000_wake_results}
\end{table*}

\begin{table*}[t]
\begin{center}
\begin{tabular}{l|cccc}
\multicolumn{1}{c}{\bf Model} & \multicolumn{1}{c}{\textbf{2-step}} & \multicolumn{1}{c}{\textbf{4-step}} & \multicolumn{1}{c}{\textbf{6-step}} & \multicolumn{1}{c}{\textbf{8-step}} \\
\hline
\hline
\rule{0pt}{1em} \bf RNN & $0.1314$ & $0.1341$ & $0.1369$ & $0.1371$ \\
\rule{0pt}{1em} \bf MNO & $0.0105$ & $\mathbf{0.0118}$ & $\mathbf{0.0130}$ & $\mathbf{0.0146}$ \\
\rule{0pt}{1em} \bf RNO & $\mathbf{0.0070}$ & $0.0165$ & $0.0250$ & $0.0413$ \\
\hline
\hline
\end{tabular}
\end{center}
\caption{Relative $L^2$ errors for the Re 5000 stall transition.}
\label{table:re5000_stall_results}
\end{table*}

For the Re 1000 stall and Re 5000 wake settings, we observe that the performance of each model is very similar, even up to autoregressively rolling out for 25 steps. This is likely attributable to the simpler dynamics of pre-tipping Re 1000 stall and Re 5000 wake, where the dynamics are smoother. In contrast, the pre-tipping $L^2$ errors in the Re 5000 stall setting are more interesting because the models must successfully forecast beyond the wake transition. We observe that \RNN is unable to learn these dynamics and that the neural operator methods vastly outperform the finite-dimensional \RNN. Interestingly, \MNO outperforms \RNO for the portion of the rollout beyond the training regime, whereas \RNO outperforms \MNO within the training regime of 3 autoregressive steps. However, \RNO in general outperforms both \MNO and \RNO in forecasting the tipping points.

\begin{figure}[t]
    \centering
        \begin{subfigure}{0.33\textwidth}
            \centering
            \includegraphics[width=\textwidth]{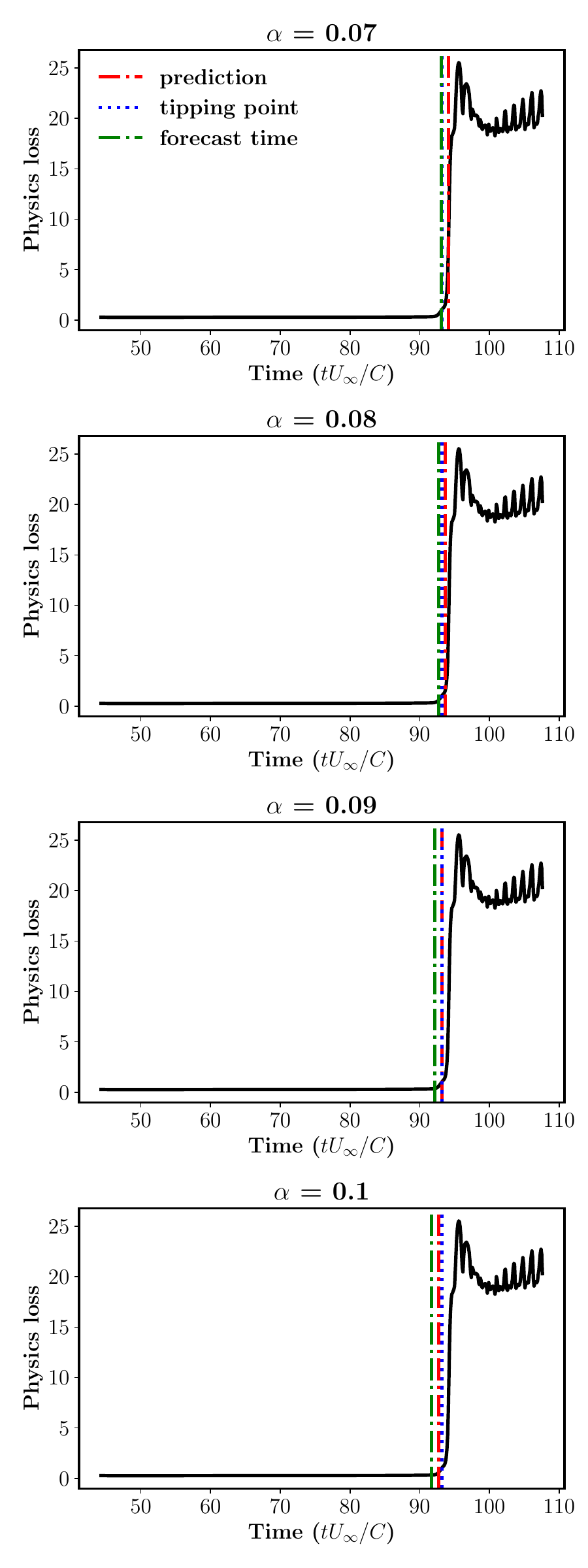}
            \caption{\RNO}
            \label{fig:rno_re1000_wake}
        \end{subfigure}%
        \begin{subfigure}{0.33\textwidth}
            \centering
            \includegraphics[width=\textwidth]{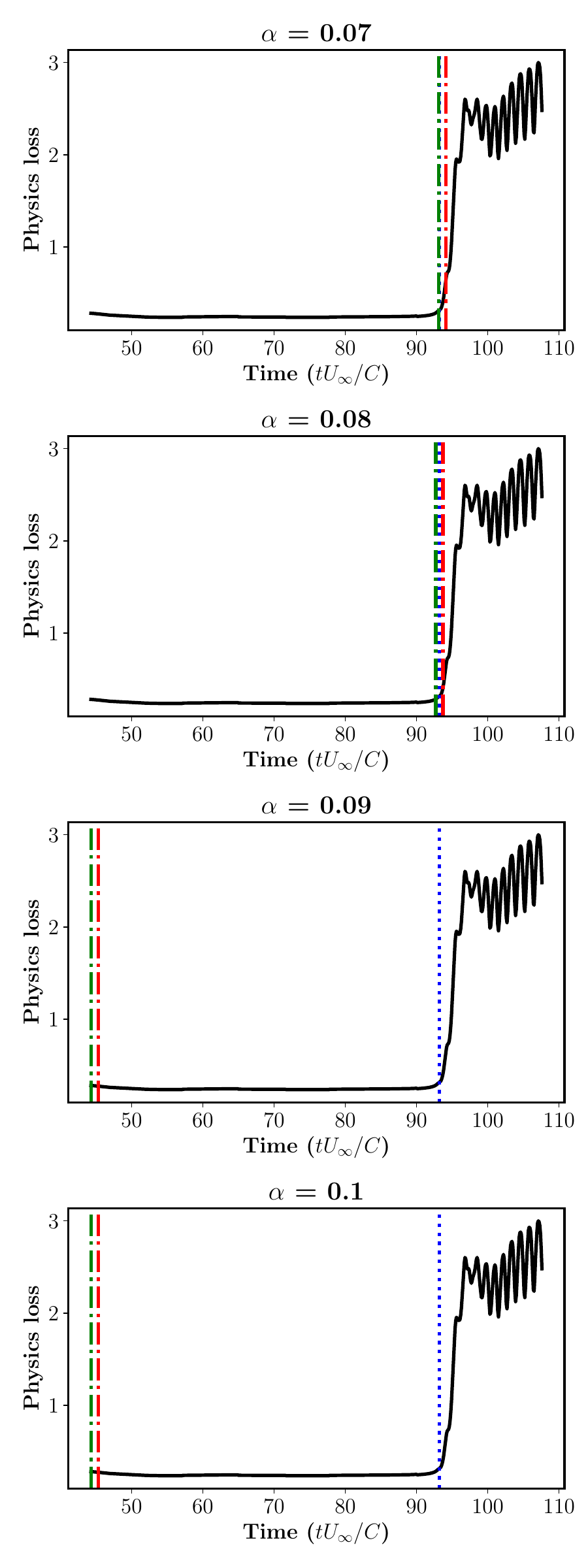}
            \caption{\MNO}
            \label{fig:mno_re1000_wake}
        \end{subfigure}%
        \begin{subfigure}{0.33\textwidth}
            \centering
            \includegraphics[width=\textwidth]{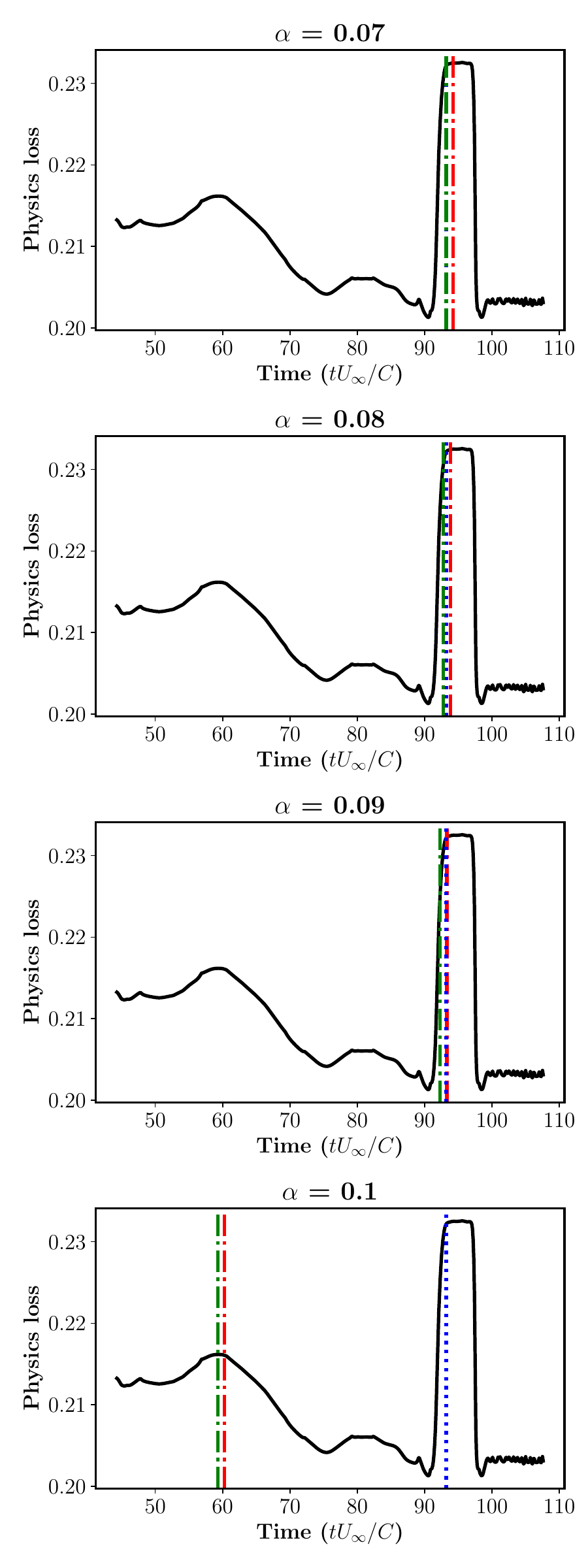}
            \caption{\RNN}
            \label{fig:rnn_re1000_wake}
        \end{subfigure}%
    \caption{Comparison of Re 1000 stall tipping point forecasting performance between \RNO, \MNO, and \RNN for various false-positive rates $\alpha$. For all presented $\alpha$, \RNO outperforms the other two models in tipping point forecasting error.}
    \label{fig:re1000_wake_varying_alpha}
\end{figure}

The airfoil tipping points in this paper are forecasted at a temporal horizon of $T = 1.0~ tU_\infty/C$ in advance. The divergence computation in the physics loss $\ell_P(u) = \|\nabla \cdot u \|_2$ is implemented using finite differences. Figure~\ref{fig:re1000_wake_varying_alpha} shows a comparison of \RNO, \MNO, and \RNN on the forecasting the Re 1000 stall transition. From the figure, we see that \RNO produces the most stable physics loss pre-tipping, and it is thus the most accurate at forecasting the tipping point for the various choices of $\alpha$.

\begin{figure}[t]
    \centering
        \begin{subfigure}{0.33\textwidth}
            \centering
            \includegraphics[width=\textwidth]{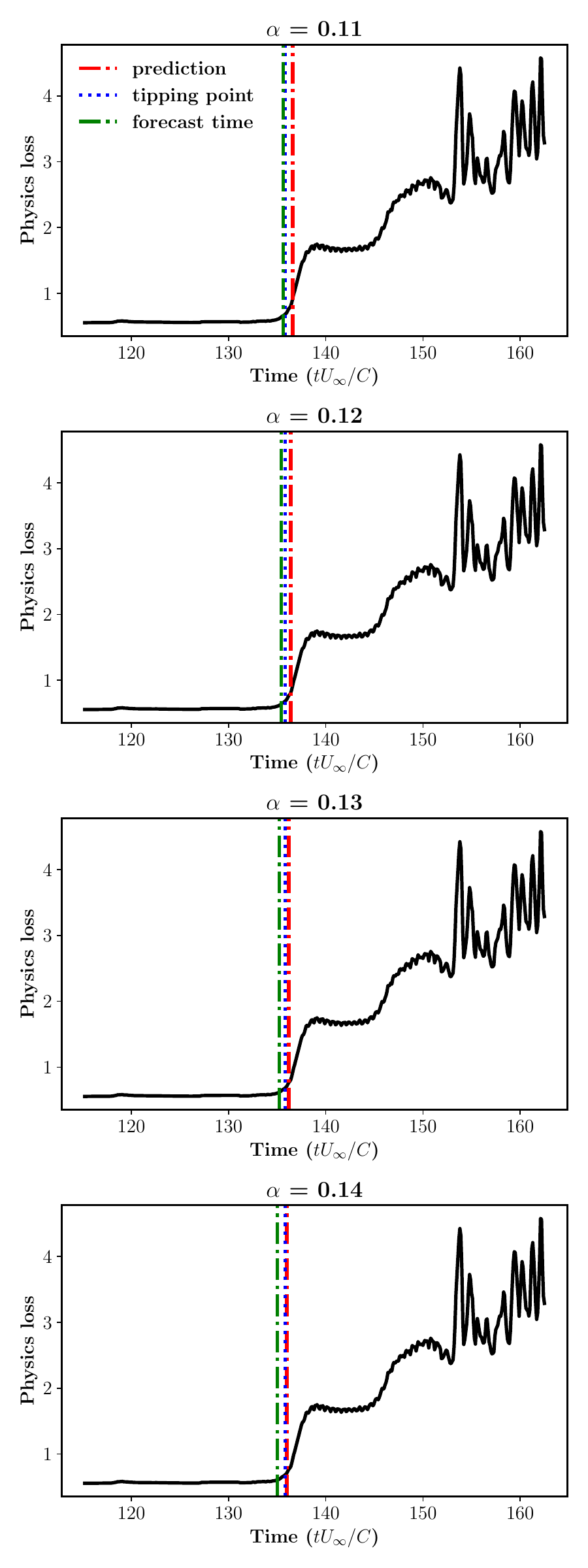}
            \caption{\RNO}
            \label{fig:rno_re5000_wake}
        \end{subfigure}%
        \begin{subfigure}{0.33\textwidth}
            \centering
            \includegraphics[width=\textwidth]{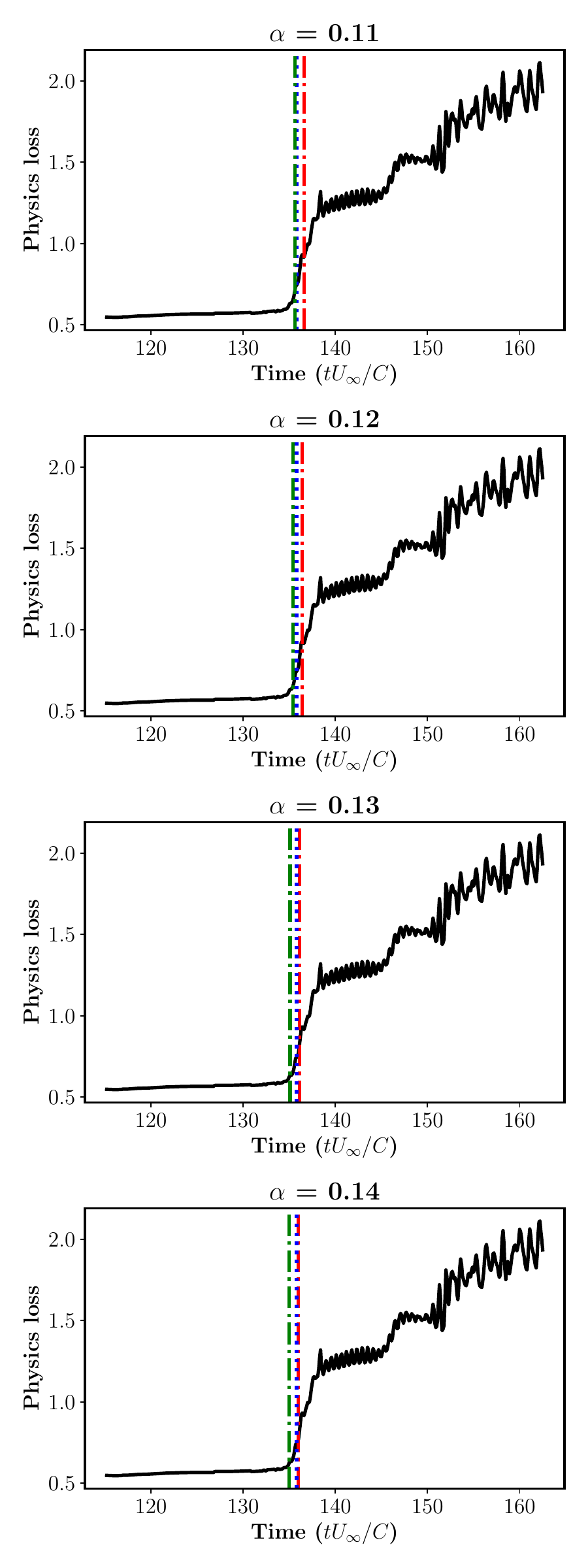}
            \caption{\MNO}
            \label{fig:mno_re5000_wake}
        \end{subfigure}%
        \begin{subfigure}{0.33\textwidth}
            \centering
            \includegraphics[width=\textwidth]{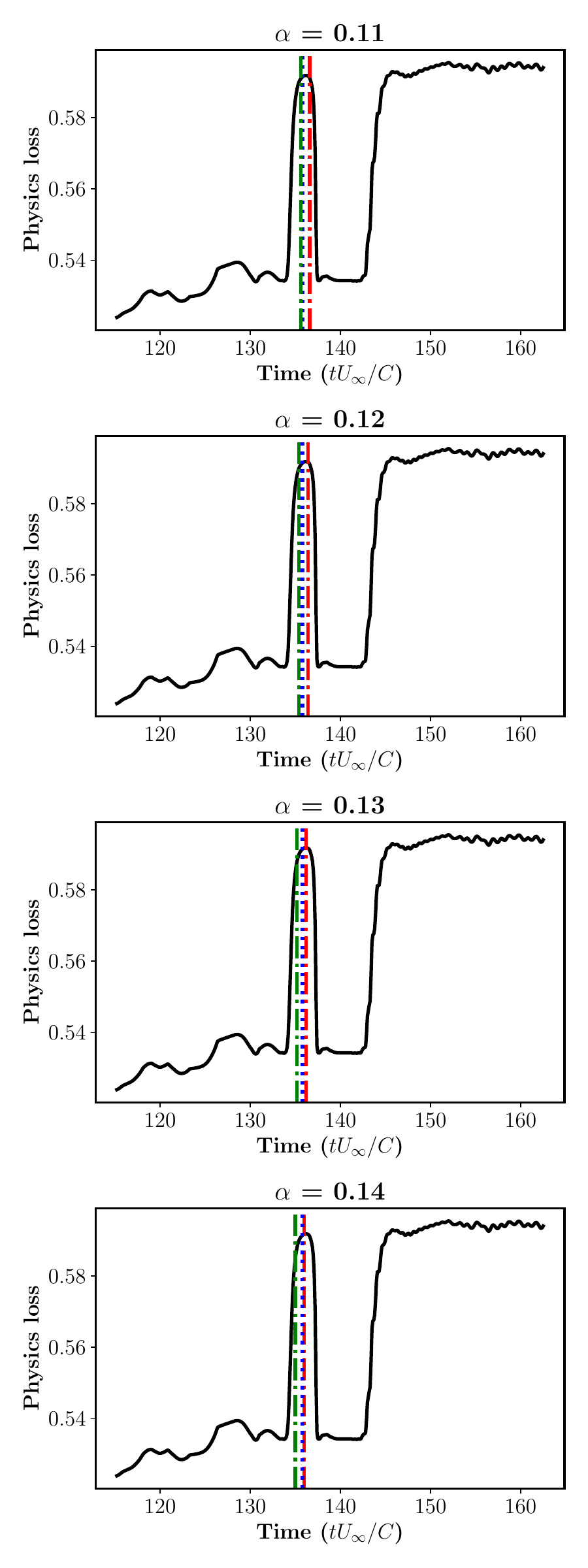}
            \caption{\RNN}
            \label{fig:rnn_re5000_wake}
        \end{subfigure}%
    \caption{Comparison of Re 5000 wake tipping point forecasting performance between \RNO, \MNO, and \RNN for various false-positive rates $\alpha$.}
    \label{fig:re5000_wake_varying_alpha}
\end{figure}

\begin{figure}[t]
    \centering
        \begin{subfigure}{0.33\textwidth}
            \centering
            \includegraphics[width=\textwidth]{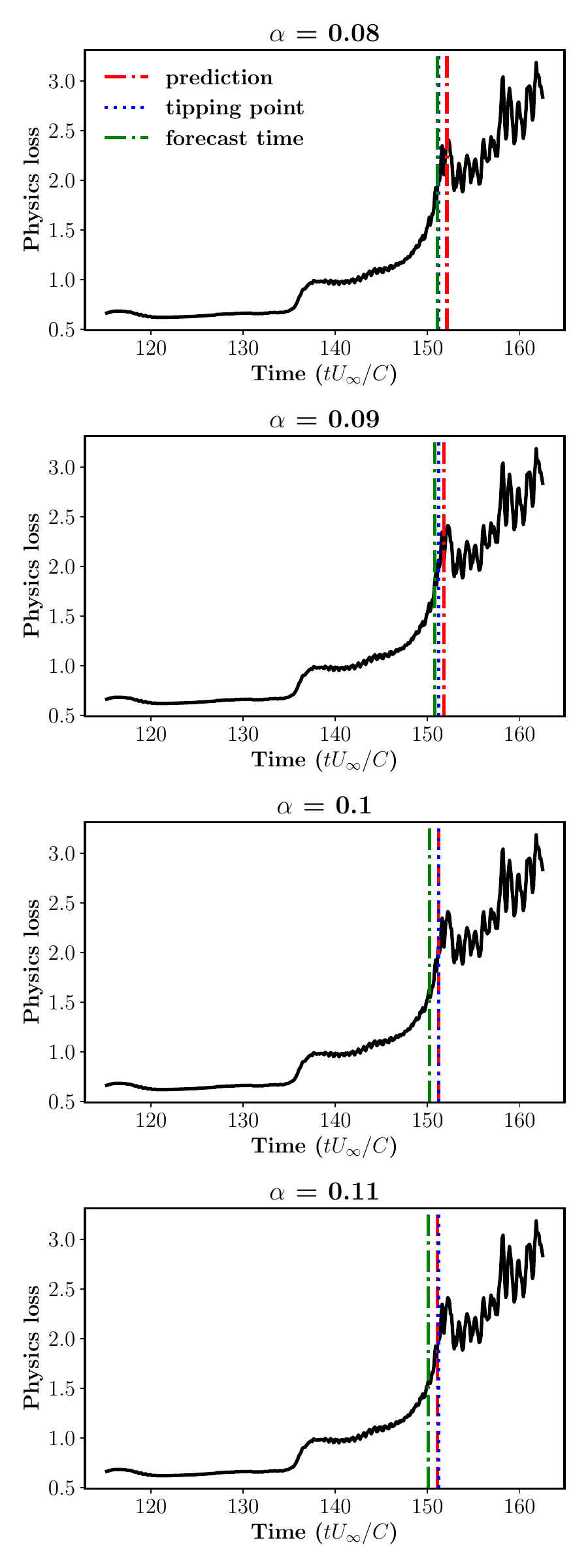}
            \caption{\RNO}
            \label{fig:rno_re5000_stall}
        \end{subfigure}%
        \begin{subfigure}{0.33\textwidth}
            \centering
            \includegraphics[width=\textwidth]{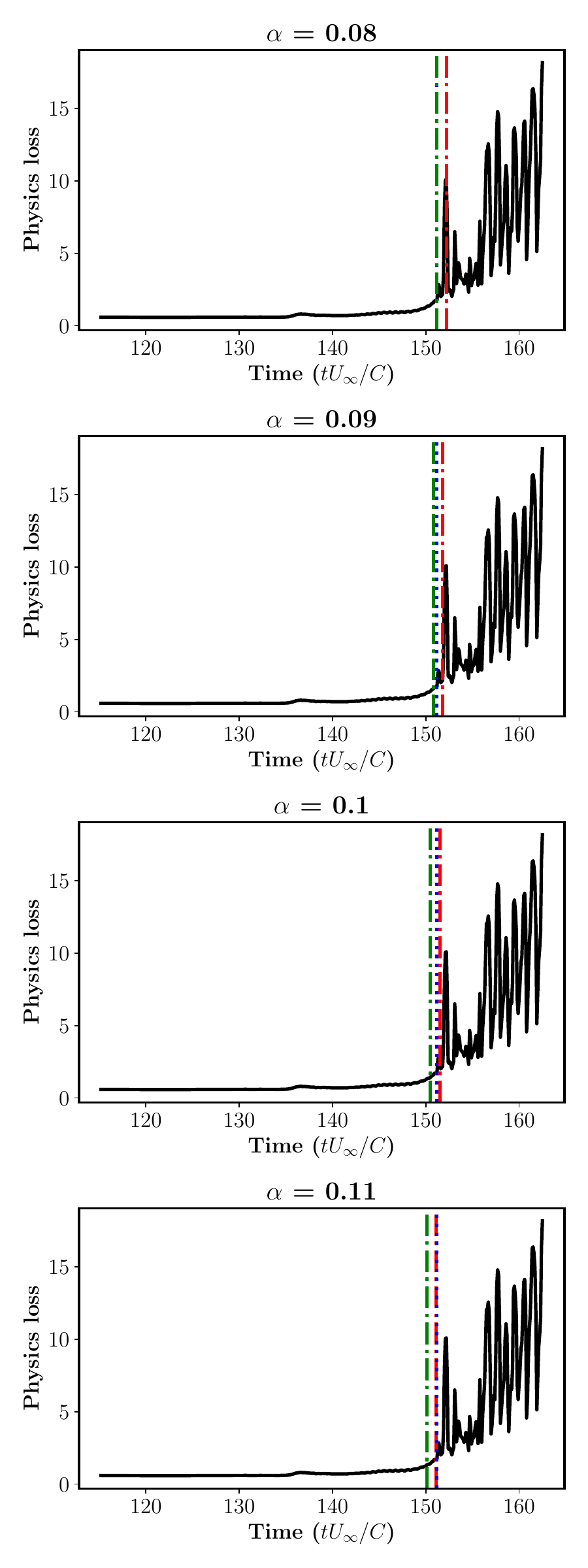}
            \caption{\MNO}
            \label{fig:mno_re5000_stall}
        \end{subfigure}%
        \begin{subfigure}{0.33\textwidth}
            \centering
            \includegraphics[width=\textwidth]{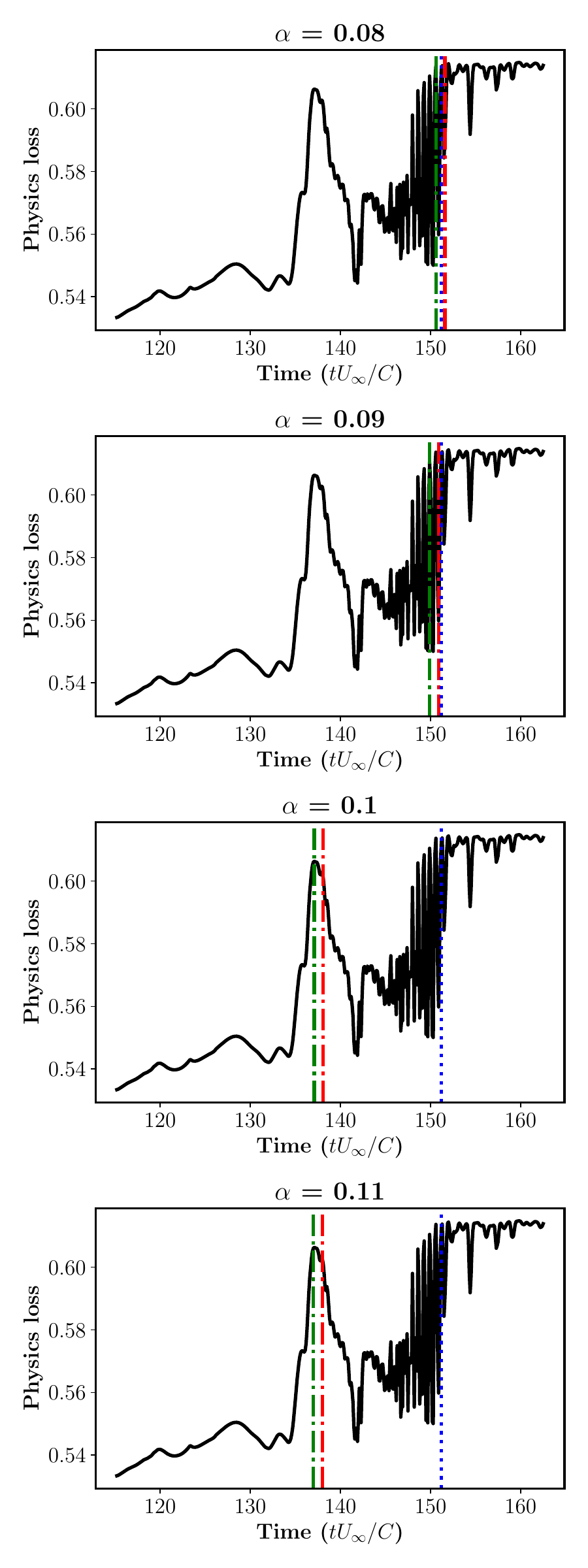}
            \caption{\RNN}
            \label{fig:rnn_re5000_stall}
        \end{subfigure}%
    \caption{Comparison of Re 5000 stall tipping point forecasting performance between \RNO, \MNO, and \RNN for various false-positive rates $\alpha$.}
    \label{fig:re5000_stall_varying_alpha}
\end{figure}

We observe a similar trend in Figure~\ref{fig:re5000_wake_varying_alpha} and Figure~\ref{fig:re5000_stall_varying_alpha}, where \RNO performs at least as well as \MNO and \RNN in forecasting each tipping point. In this Re 5000 setting, we can more easily observe the benefits of neural operators: compared to \RNO and \MNO, the \RNN physics loss is very noisy and oscillatory, particularly around the stall point. In contrast, both \RNO and \MNO exhibit a slight increase in physics loss around the wake tipping point and a larger increase around the stall tipping point. This behavior is seen in both cases when we train the models on pre-wake and pre-stall dynamics.

\begin{figure}[t]
    \centering
        \begin{subfigure}{0.33\textwidth}
            \centering
            \includegraphics[width=\textwidth]{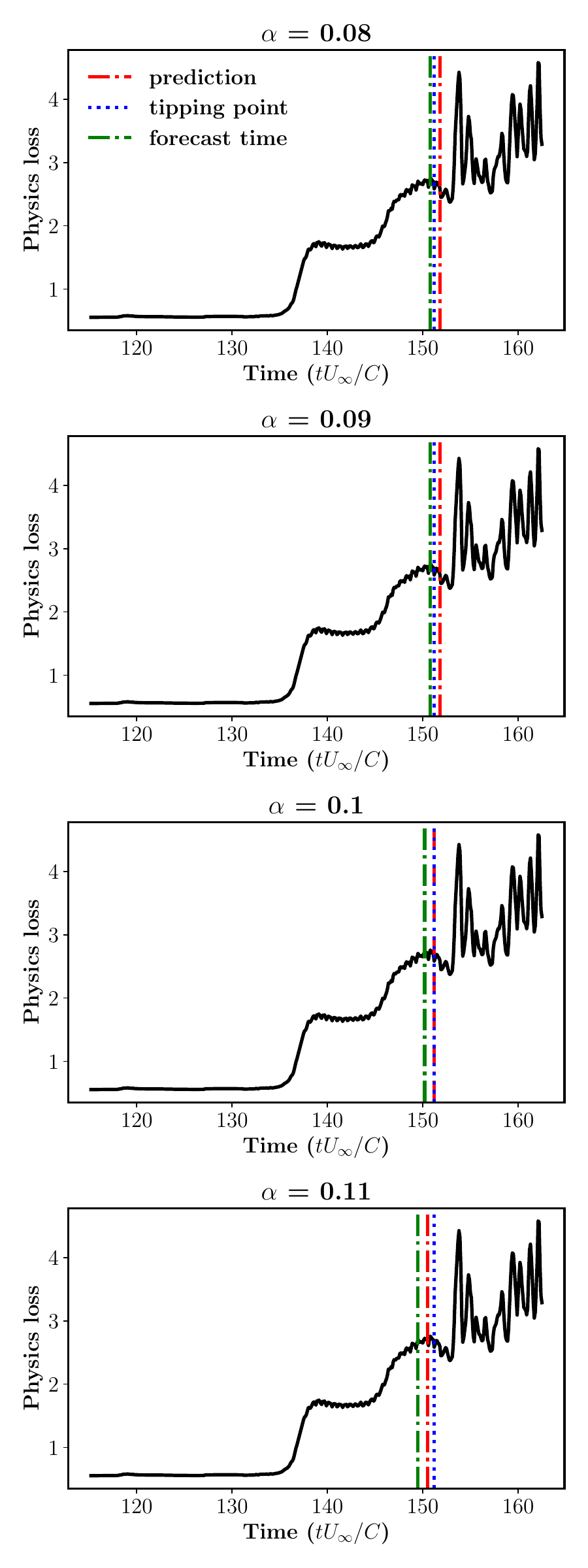}
            \caption{\RNO}
            \label{fig:rno_re5000_stall_trained_on_pre_wake}
        \end{subfigure}%
        \begin{subfigure}{0.33\textwidth}
            \centering
            \includegraphics[width=\textwidth]{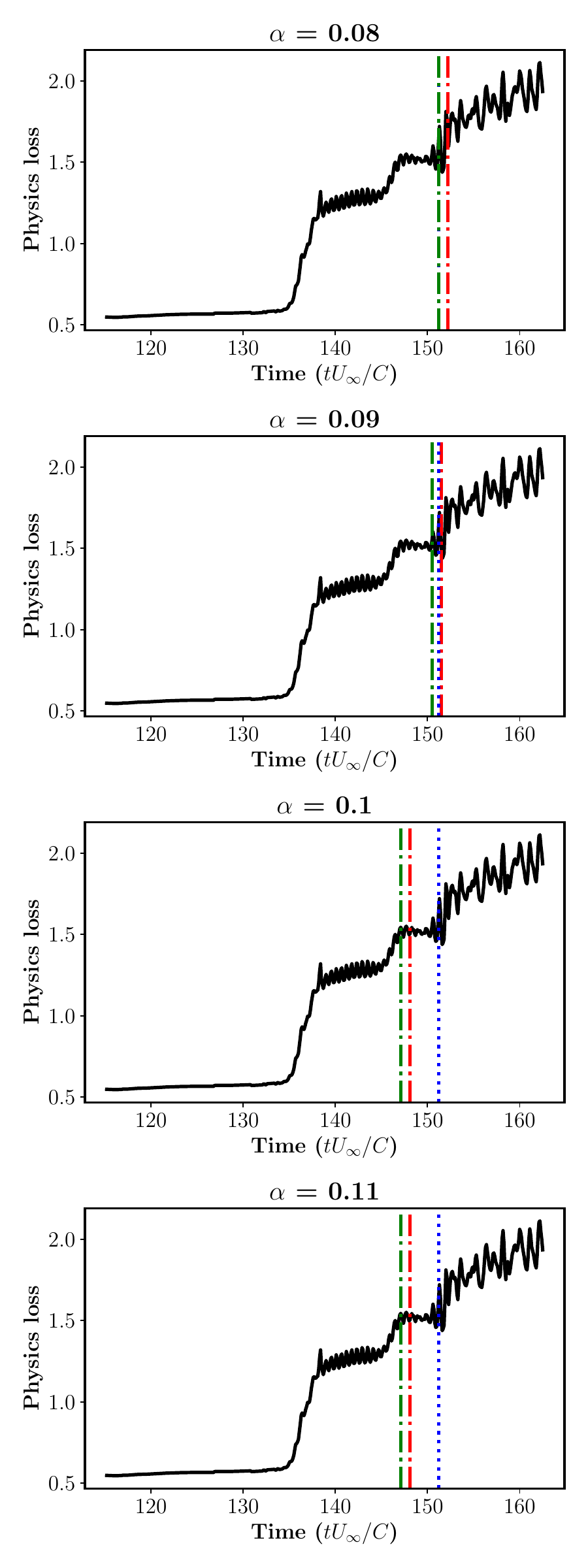}
            \caption{\MNO}
            \label{fig:mno_re5000_stall_trained_on_pre_wake}
        \end{subfigure}%
        \begin{subfigure}{0.33\textwidth}
            \centering
            \includegraphics[width=\textwidth]{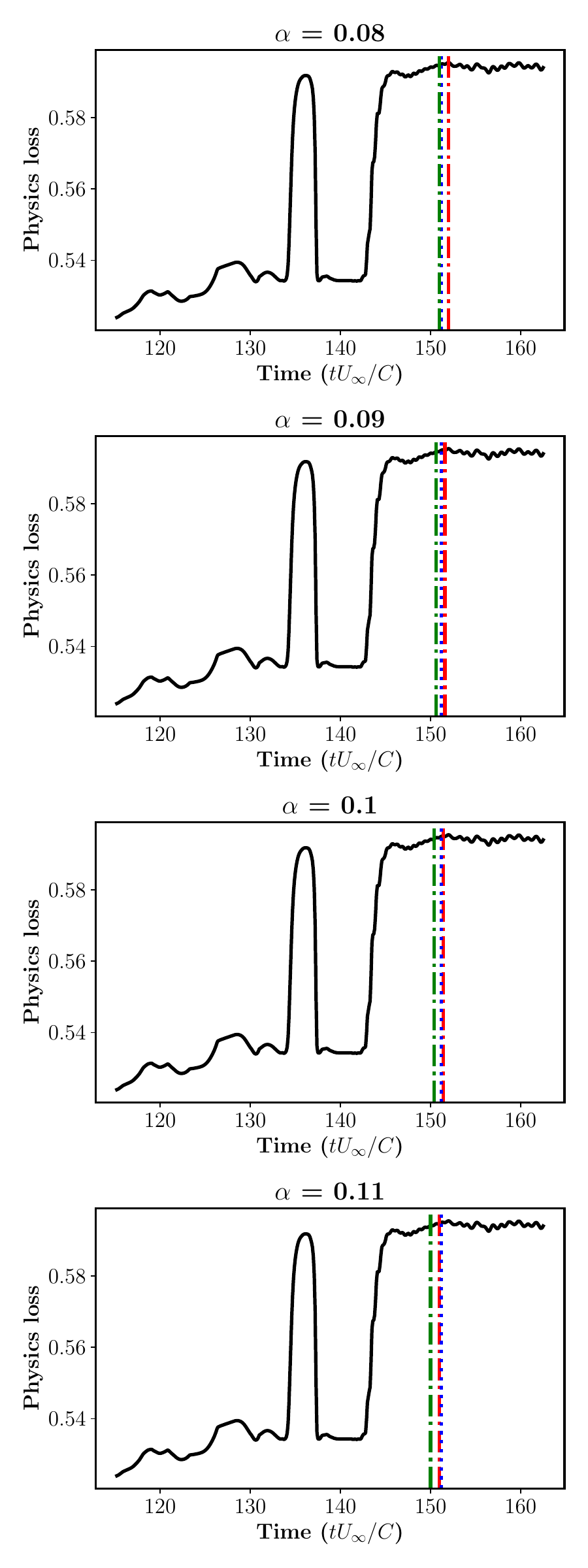}
            \caption{\RNN}
            \label{fig:rnn_re5000_stall_trained_on_pre_wake}
        \end{subfigure}%
    \caption{Comparison of Re 5000 stall tipping point forecasting performance between \RNO, \MNO, and \RNN trained on pre-wake dynamics for various false-positive rates $\alpha$.}
    \label{fig:re5000_stall_trained_on_pre_wake_varying_alpha}
\end{figure}

\begin{figure}[t]
    \centering
        \begin{subfigure}{0.33\textwidth}
            \centering
            \includegraphics[width=\textwidth]{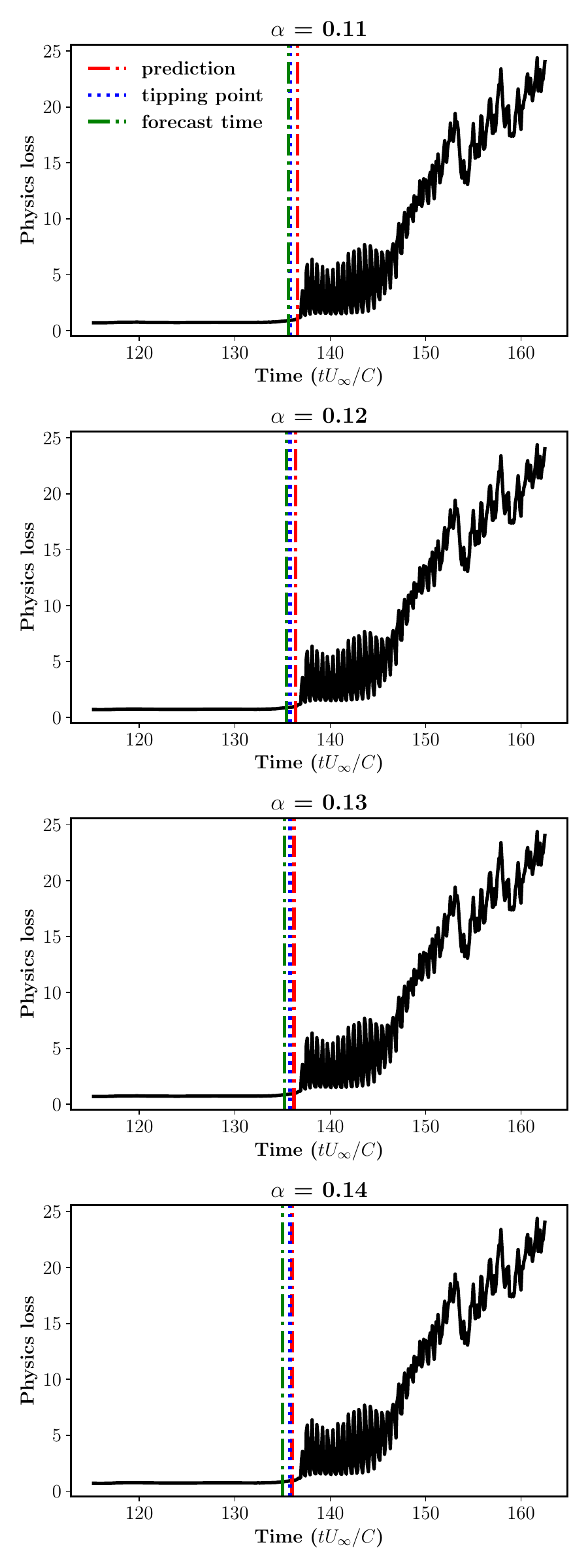}
            \caption{\RNO}
            \label{fig:rno_re5000_wake_re1000_transfer}
        \end{subfigure}%
        \begin{subfigure}{0.33\textwidth}
            \centering
            \includegraphics[width=\textwidth]{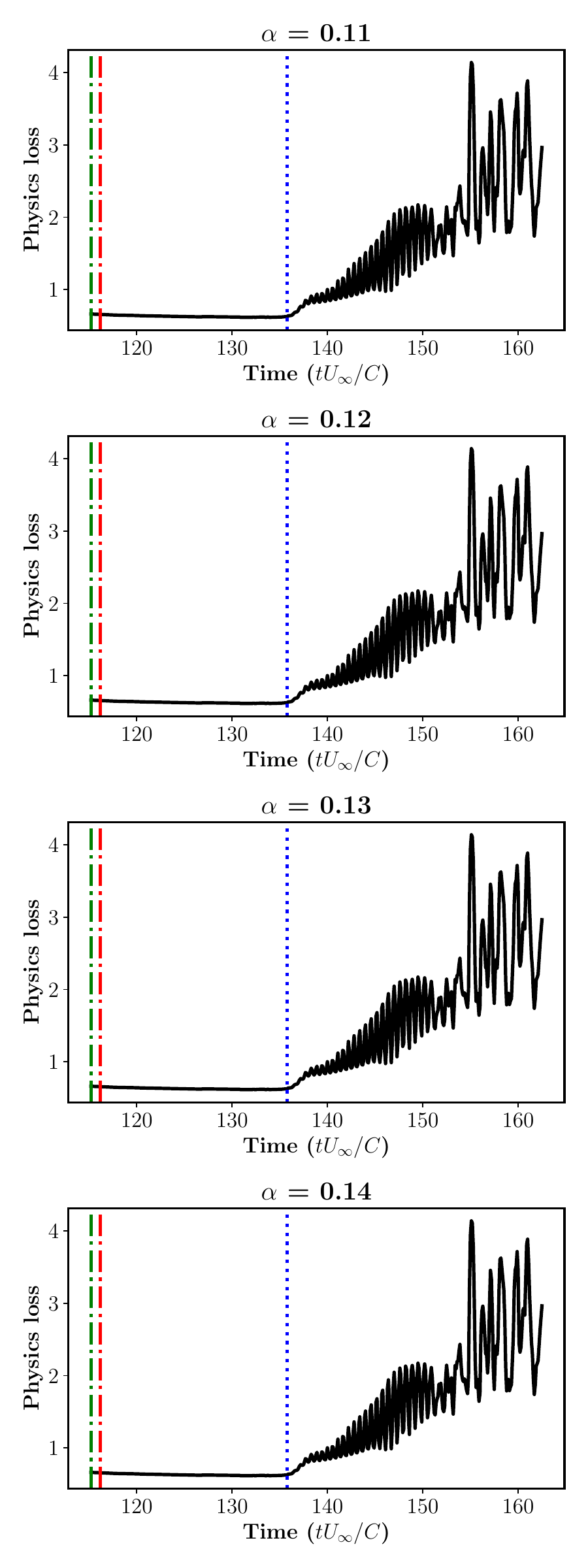}
            \caption{\MNO}
            \label{fig:mno_re5000_wake_re1000_transfer}
        \end{subfigure}%
        \begin{subfigure}{0.33\textwidth}
            \centering
            \includegraphics[width=\textwidth]{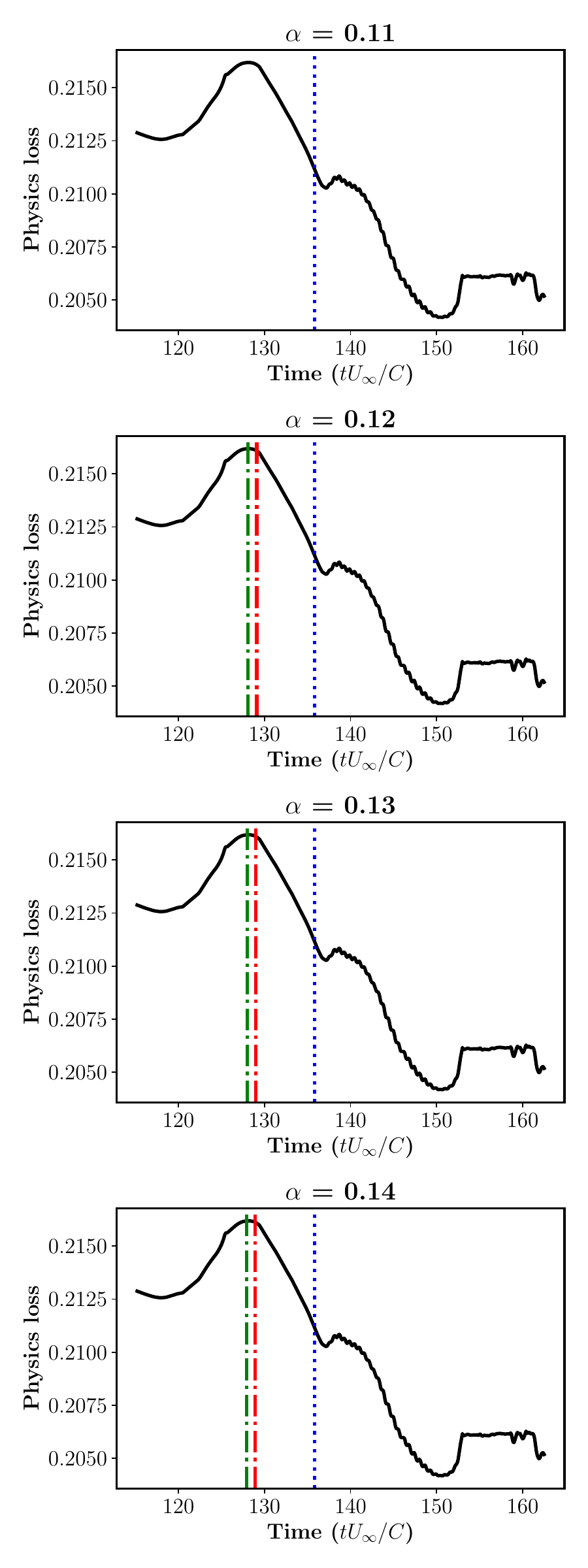}
            \caption{\RNN}
            \label{fig:rnn_re5000_wake_re1000_transfer}
        \end{subfigure}%
    \caption{Comparison of Re 5000 wake tipping point forecasting performance between \RNO, \MNO, and \RNN trained on Re 1000 pre-stall dynamics for various false-positive rates $\alpha$. For $\alpha = 0.11$, the \RNN predictions do not forecast any tipping point within the time horizon.}
    \label{fig:re5000_wake_re1000_transfer_varying_alpha}
\end{figure}

\begin{figure}[t]
    \centering
        \begin{subfigure}{0.33\textwidth}
            \centering
            \includegraphics[width=\textwidth]{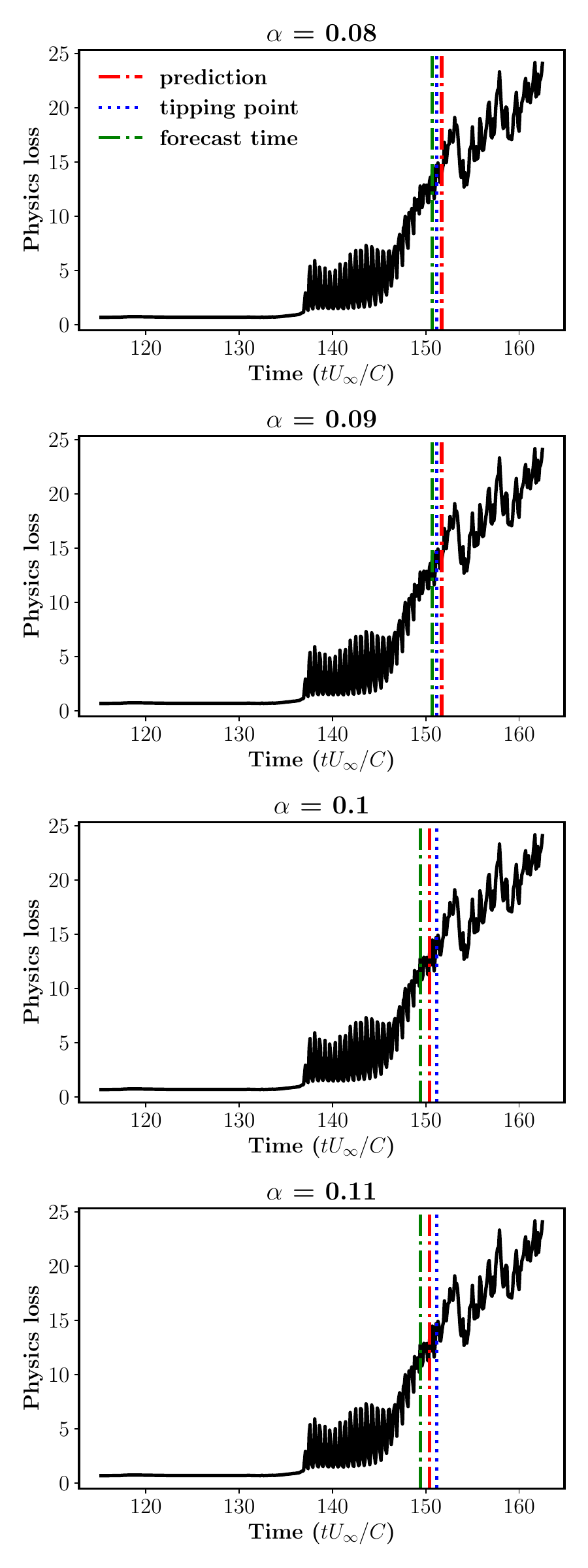}
            \caption{\RNO}
            \label{fig:rno_re5000_stall_re1000_transfer}
        \end{subfigure}%
        \begin{subfigure}{0.33\textwidth}
            \centering
            \includegraphics[width=\textwidth]{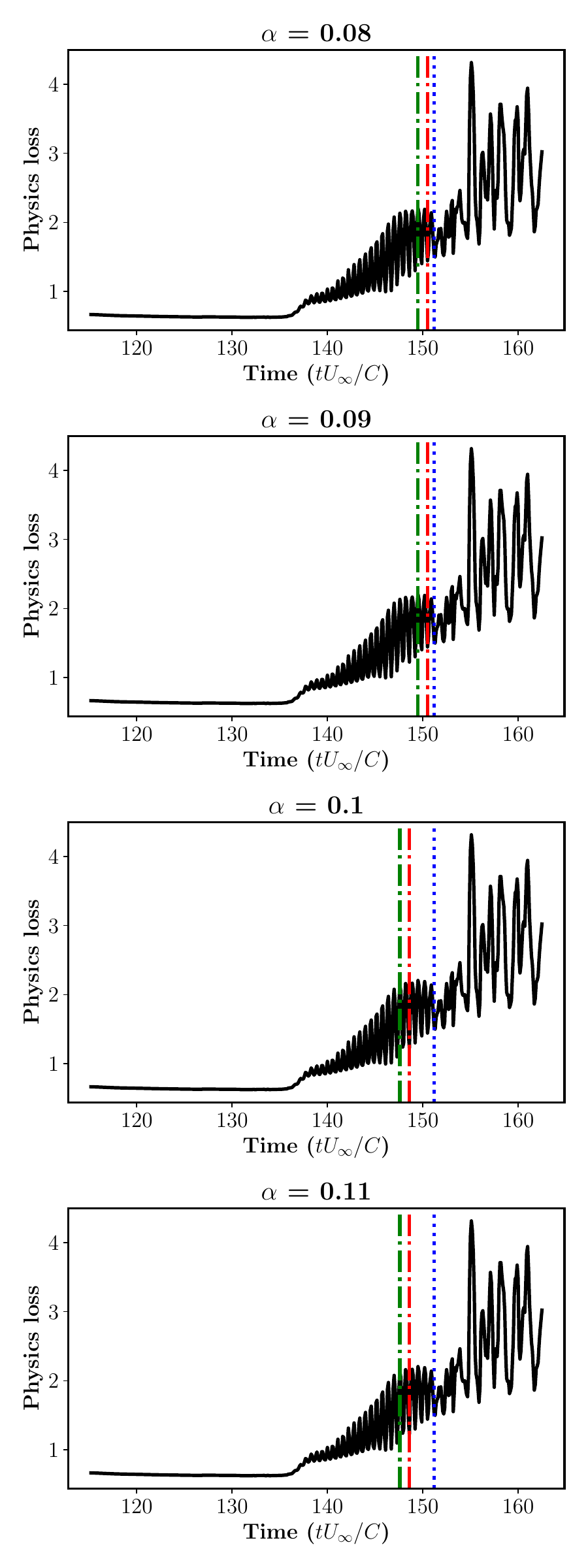}
            \caption{\MNO}
            \label{fig:mno_re5000_stall_re1000_transfer}
        \end{subfigure}%
        \begin{subfigure}{0.33\textwidth}
            \centering
            \includegraphics[width=\textwidth]{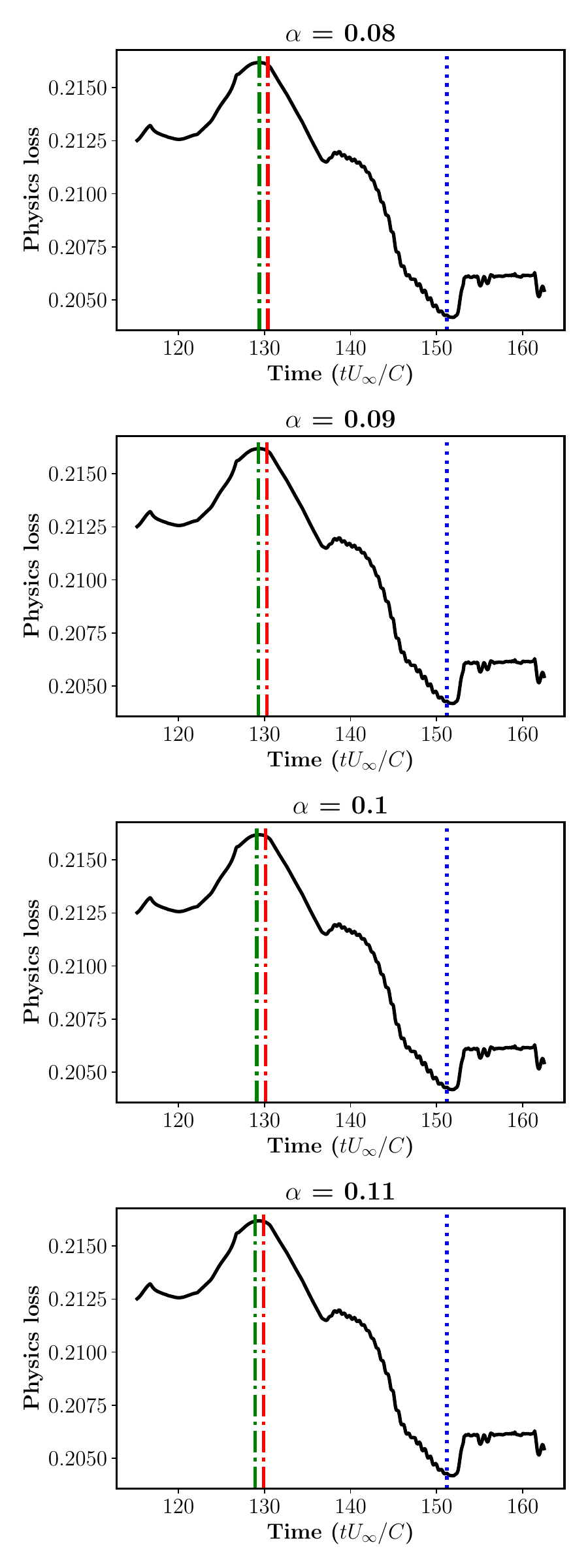}
            \caption{\RNN}
            \label{fig:rnn_re5000_stall_re1000_transfer}
        \end{subfigure}%
    \caption{Comparison of Re 5000 stall tipping point forecasting performance between \RNO, \MNO, and \RNN trained on Re 1000 pre-stall dynamics for various false-positive rates $\alpha$.}
    \label{fig:re5000_stall_re1000_transfer_varying_alpha}
\end{figure}

Finally, as shown in Section~\ref{subsec:airfoil_results}, we explore the generalization ability of our methods beyond the regime of the training dynamics. Figure~\ref{fig:re5000_stall_trained_on_pre_wake_varying_alpha} shows that \RNO, \MNO, and \RNN trained on pre-wake data can be used to accurately forecast the stall tipping point at inference time. Figure~\ref{fig:re5000_wake_re1000_transfer_varying_alpha} shows the results of \RNO trained on Re 1000 pre-stall dynamics can be used directly to forecast the Re 5000 wake tipping point, and Figure~\ref{fig:re5000_stall_re1000_transfer_varying_alpha} shows the analogous result for the Re 5000 stall tipping point. In the latter two cases, we find that \MNO and \RNN lag behind \RNO's generalization ability. All of the these trends hold true across a variety of $\alpha$.

\section{Robustness to rate of parameter change and non-tipping scenarios}
\label{appdx:rate_induced_no_tip}

In this paper, we demonstrate the applicability of our method for forecasting tipping points across a variety of different systems, such as non-stationary Lorenz, non-stationary Kuramoto-Sivashinsky, a cloud cover dynamics model, and stall and wake transitions in a flow over an airfoil. For each of these experiments, we fix a rate of change of a critical system parameter to induce non-stationarity. In this section, we show that our method is robust to variations in this rate of change. 

\begin{figure}[t]
    \centering
        \begin{subfigure}{\textwidth}
            \centering
            \includegraphics[width=\textwidth]{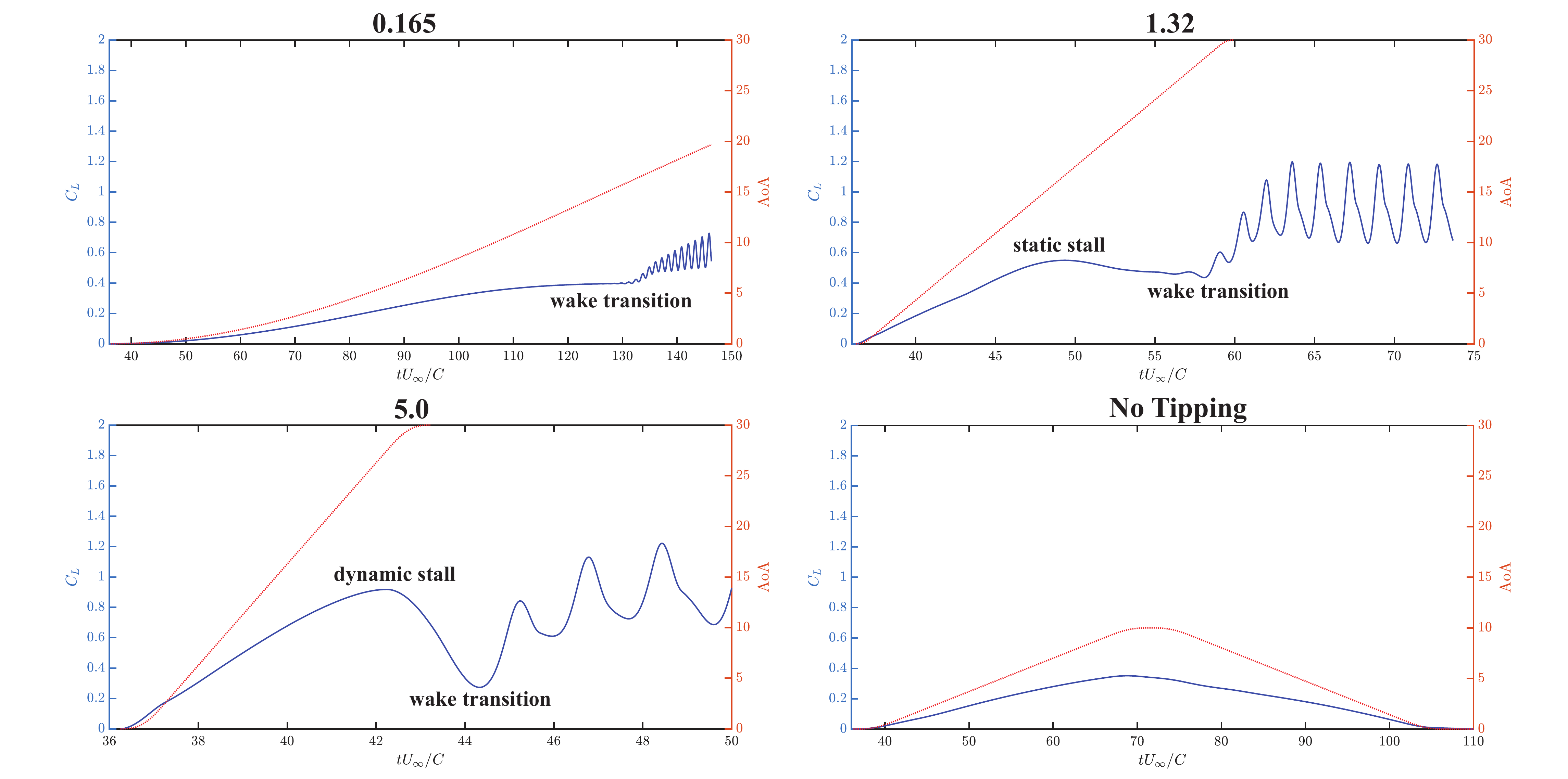}
        \end{subfigure}
        \caption{Lift coefficient and angle-of-attack trajectories for different pitch rates. The lift coefficient, $C_L$ (blue solid line), and angle of attack, AoA (red dotted line), are shown as functions of nondimensional time for cases starting from the same initial condition but with different rates of change of AoA. The rate for each case is indicated in the corresponding subplot title in units of $\mathrm{AoA}/t^*$, where $t^* = t U_\infty / C$. The tipping point is labeled in each subplot.}
    \label{fig:lift_AoA_traj}
\end{figure}

\begin{figure}[t]
    \centering
        \begin{subfigure}{0.33\textwidth}
            \centering
            \includegraphics[width=\textwidth]{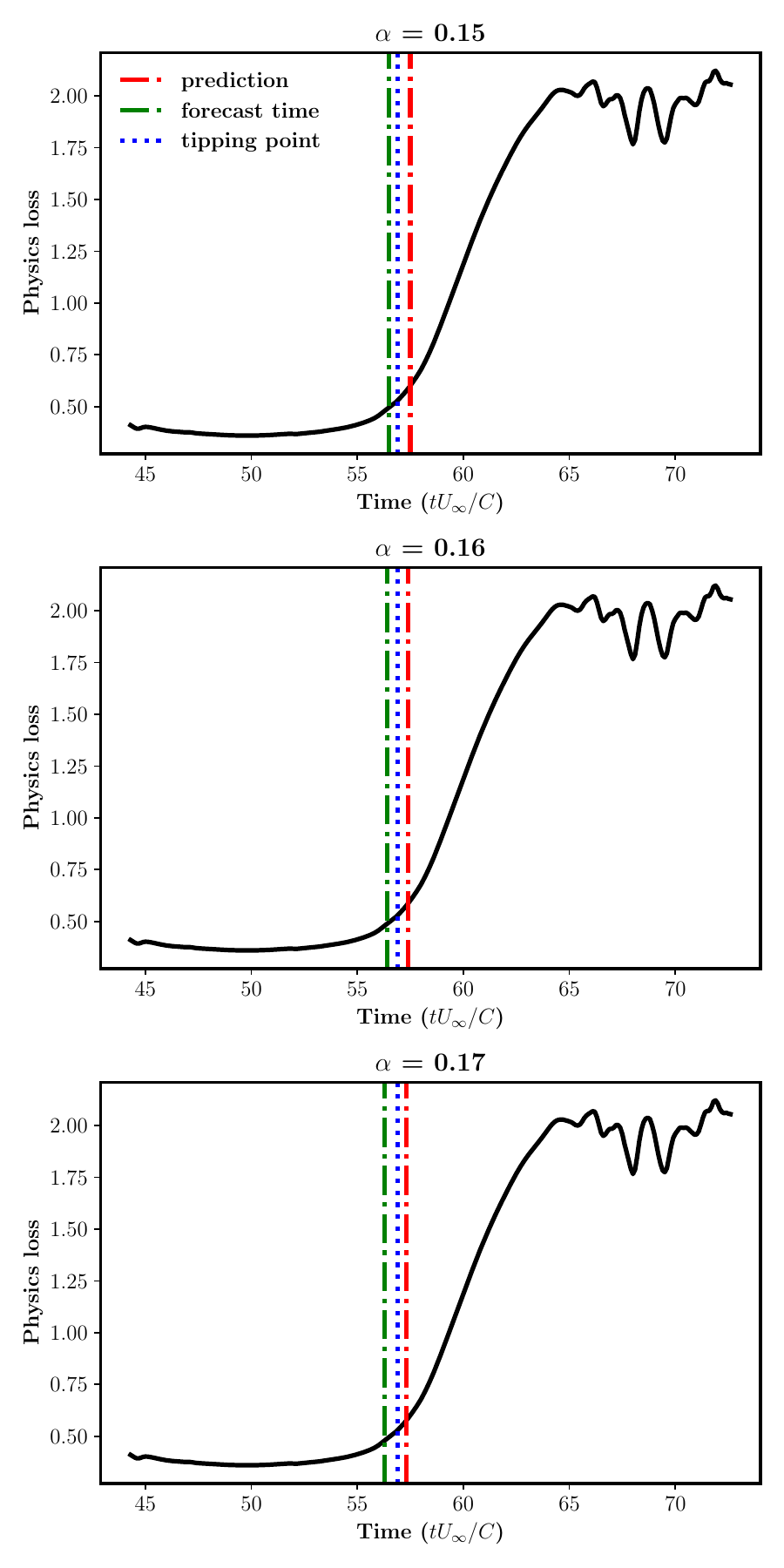}
            \caption{\RNO}
            \label{fig:rno_run1.32}
        \end{subfigure}%
        \begin{subfigure}{0.33\textwidth}
            \centering
            \includegraphics[width=\textwidth]{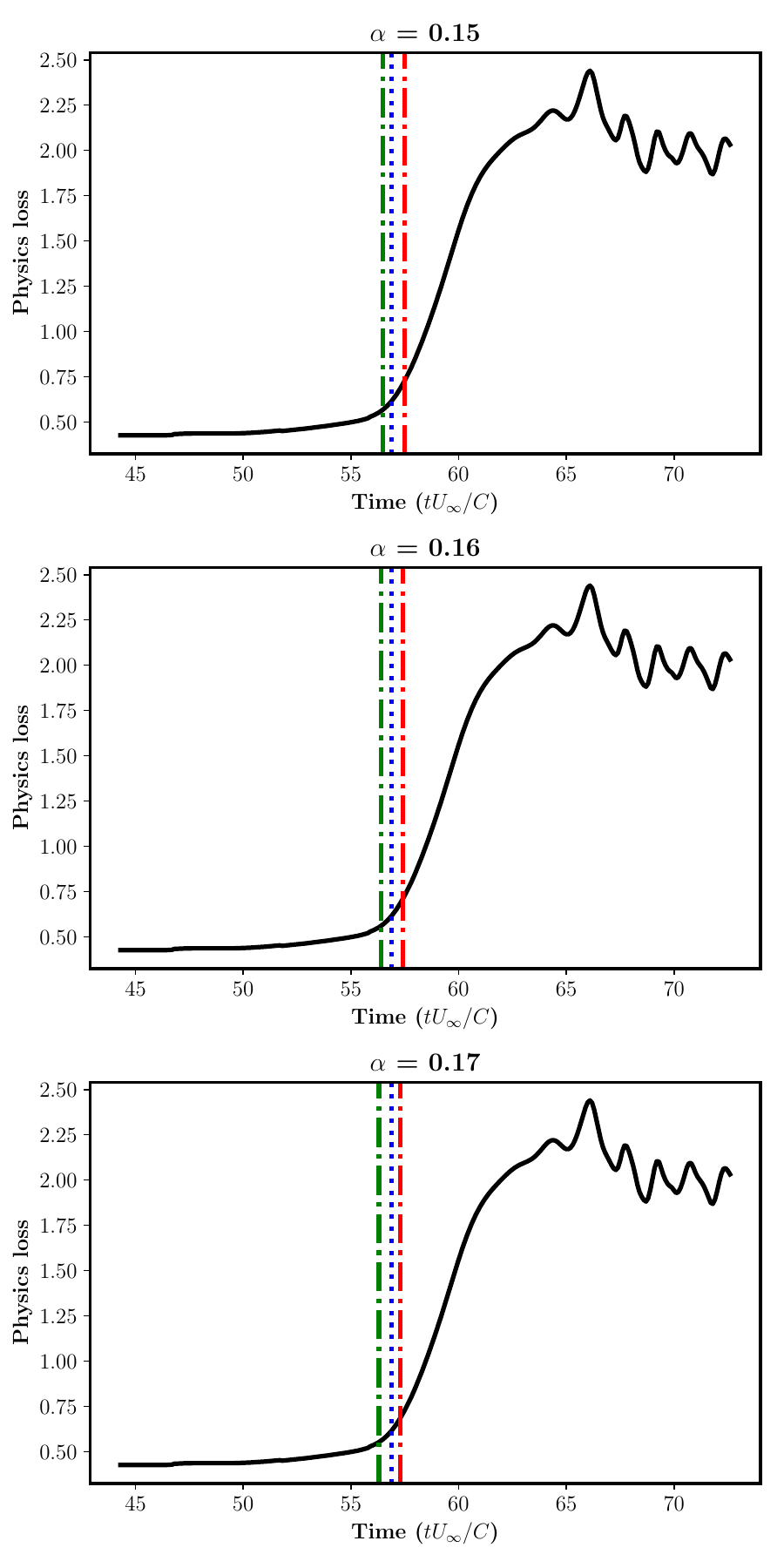}
            \caption{\MNO}
            \label{fig:mno_run1.32}
        \end{subfigure}%
        \begin{subfigure}{0.33\textwidth}
            \centering
            \includegraphics[width=\textwidth]{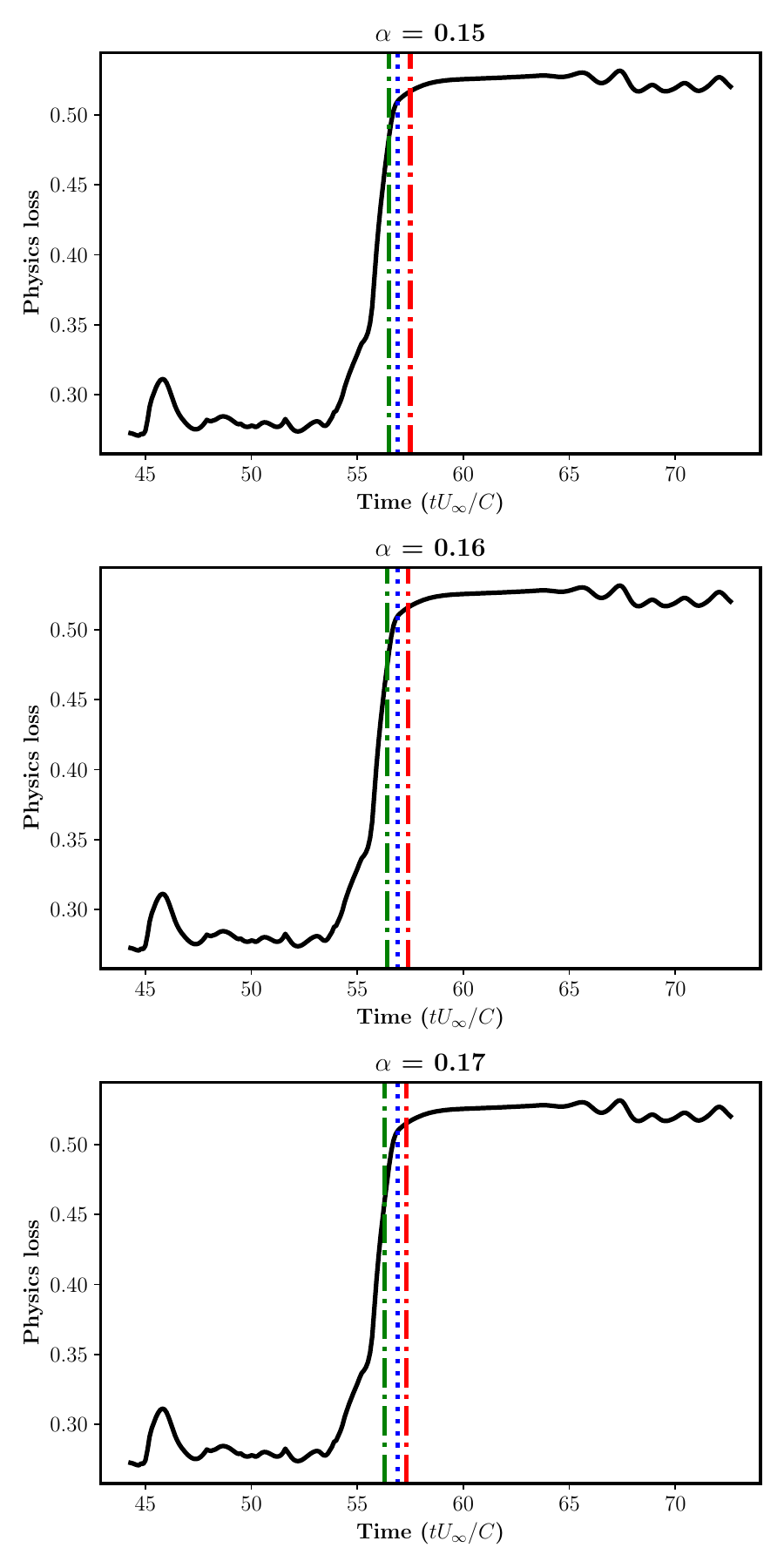}
            \caption{\RNN}
            \label{fig:rnn_run1.32}
        \end{subfigure}%
    \caption{Comparison of tipping point forecasting performance between \RNO, \MNO, and \RNN trained on 1.32 AoA pre-stall dynamics for various false-positive rates $\alpha$.}
    \label{fig:1.32AoA}
\end{figure}

\begin{figure}[t]
    \centering
        \begin{subfigure}{0.33\textwidth}
            \centering
            \includegraphics[width=\textwidth]{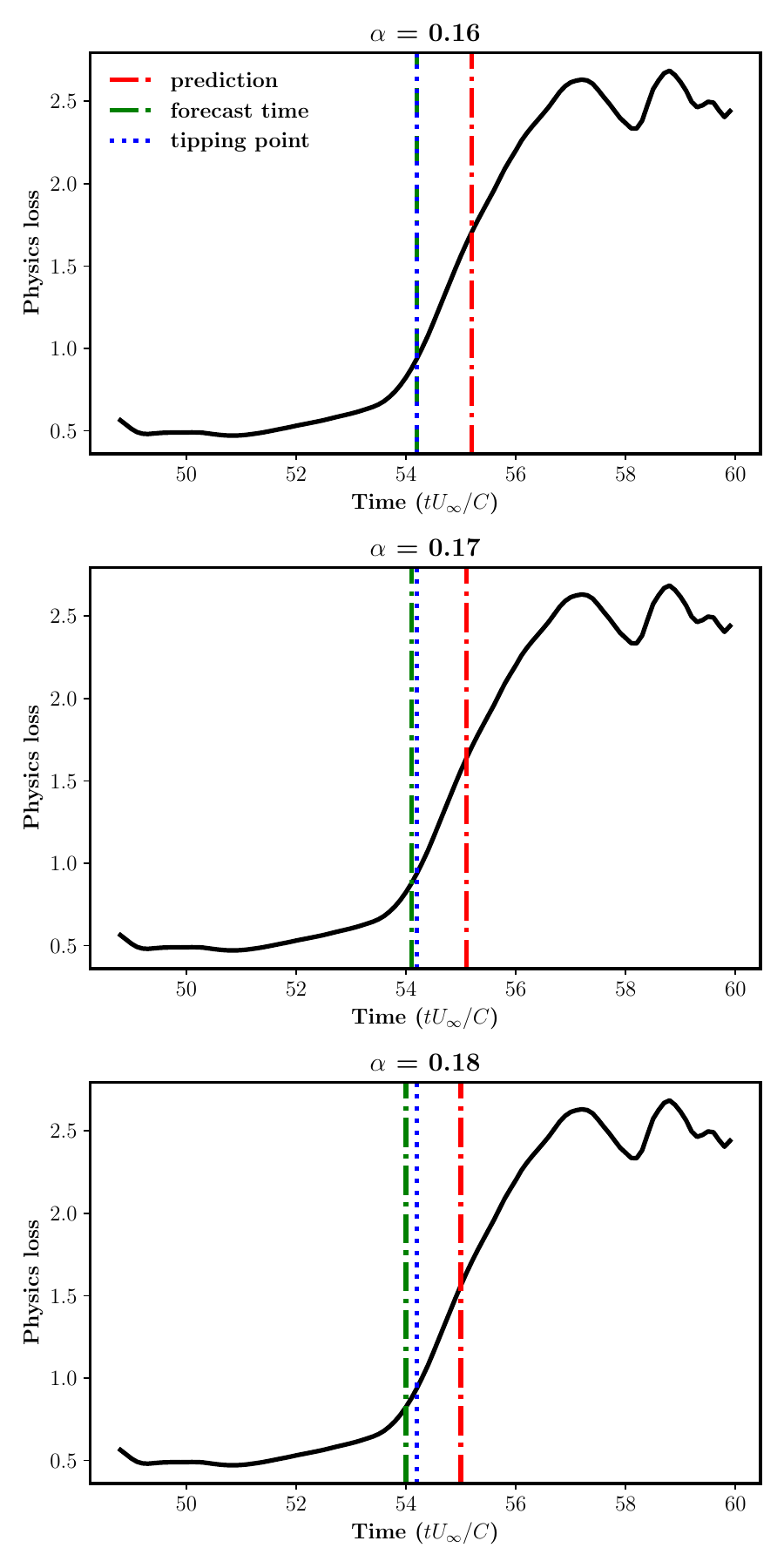}
            \caption{\RNO}
            \label{fig:rno_run5.0}
        \end{subfigure}%
        \begin{subfigure}{0.33\textwidth}
            \centering
            \includegraphics[width=\textwidth]{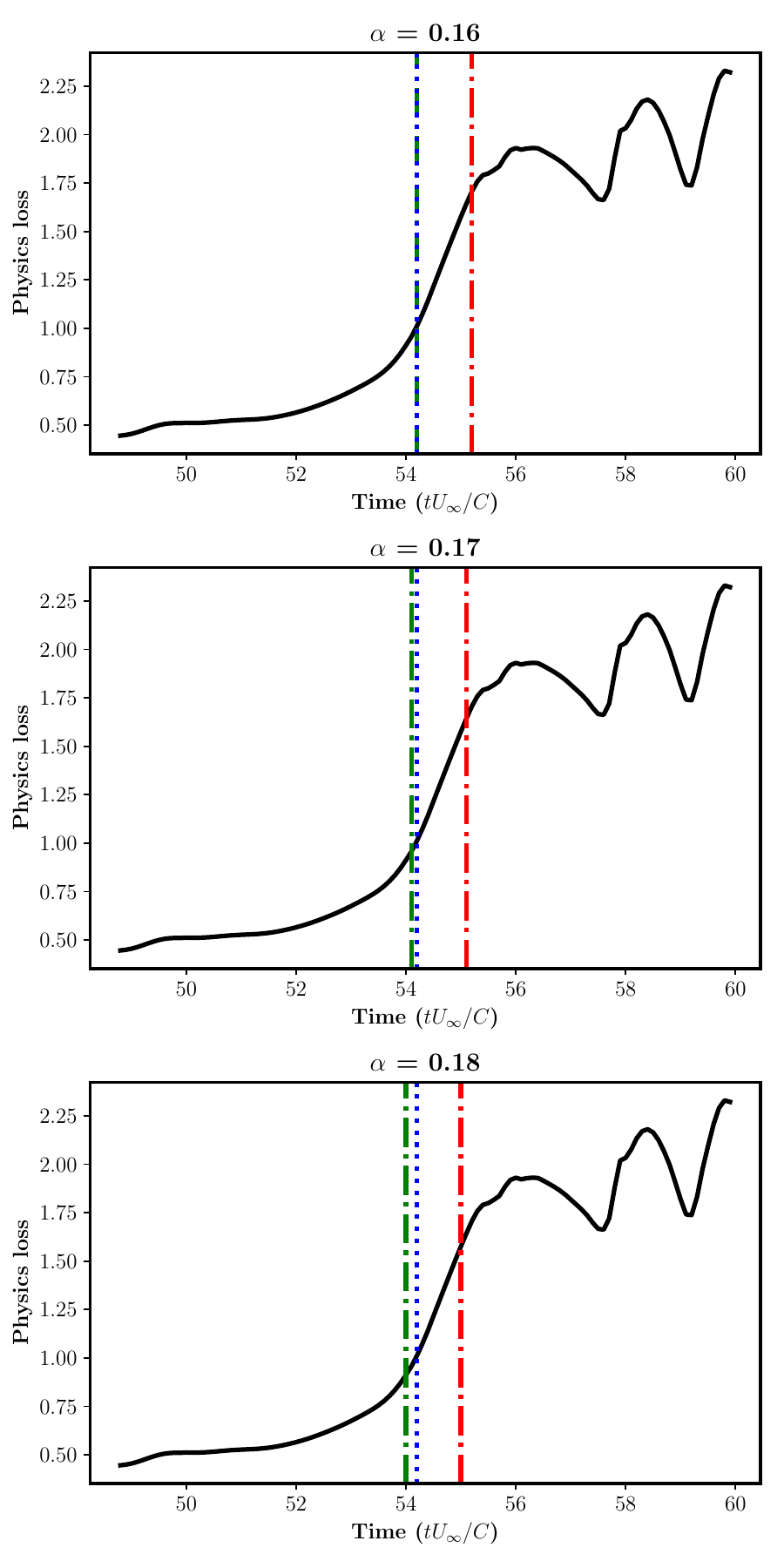}
            \caption{\MNO}
            \label{fig:mno_run5.0}
        \end{subfigure}%
        \begin{subfigure}{0.33\textwidth}
            \centering
            \includegraphics[width=\textwidth]{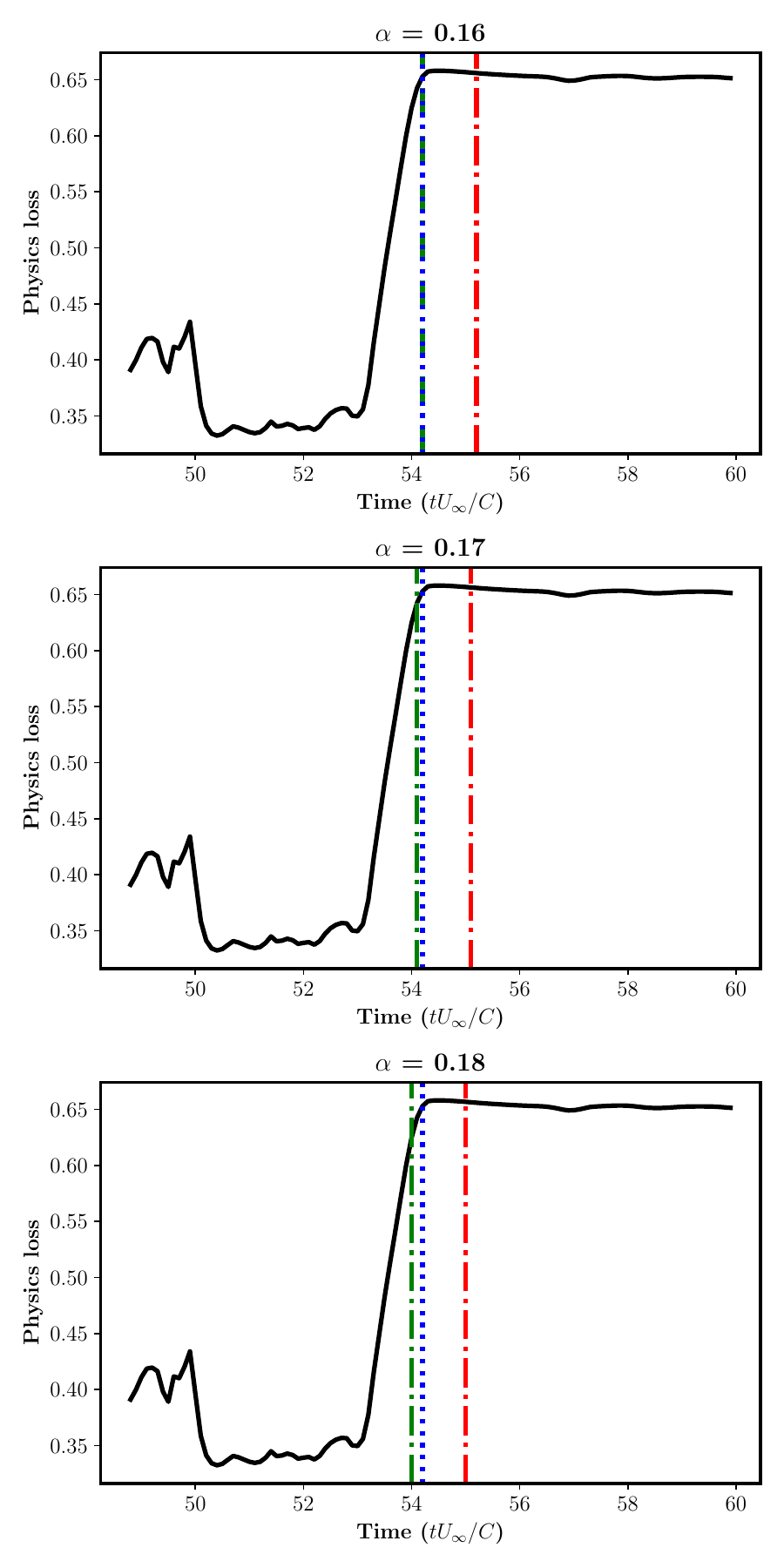}
            \caption{\RNN}
            \label{fig:rnn_run5.0}
        \end{subfigure}%
    \caption{Comparison of tipping point forecasting performance between \RNO, \MNO, and \RNN trained on 5.0 AoA pre-stall dynamics for various false-positive rates $\alpha$.}
    \label{fig:5.0AoA}
\end{figure}

We select the more difficult airfoil setting for these experiments. We show that \RNN, \MNO, and \RNO are all capable of forecasting stall transitions in the airfoil setting under varying rates of change. In particular, we train and evaluate models under the same setup as explained in Appendix~\ref{appdx:airfoil_dataset} except with two different angles of attack: 1.32 and 5.0 degrees per time unit (which we shorten to 1.32 AoA and 5.0 AoA, respectively). The dynamics are qualitatively distinct from the 0.33-degrees case: in the former, we predict the static stall transition and in the latter we predict a dynamic stall. See Figure~\ref{fig:lift_AoA_traj} for more details. Comparisons between \RNO, \MNO, and \RNN can be found in Figure~\ref{fig:1.32AoA} and Figure~\ref{fig:5.0AoA}. In the 1.32 AoA case, we see that all three models are capable of forecasting the tipping point with high accuracy, whereas in the faster dynamic stall case (5.0 AoA setting), all three models tend to predict after the true tipping point for the $\alpha$ values studied. We observe that for larger values of $\alpha$, the model can accurately forecast the tipping point, but for consistency with previous results and to maintain fairness in evaluation, we keep the three smallest values of $\alpha$ (rounded to two decimal places) permitted by the size of our calibration set.

\begin{figure}[t]
    \centering
        \begin{subfigure}{0.33\textwidth}
            \centering
            \includegraphics[width=\textwidth]{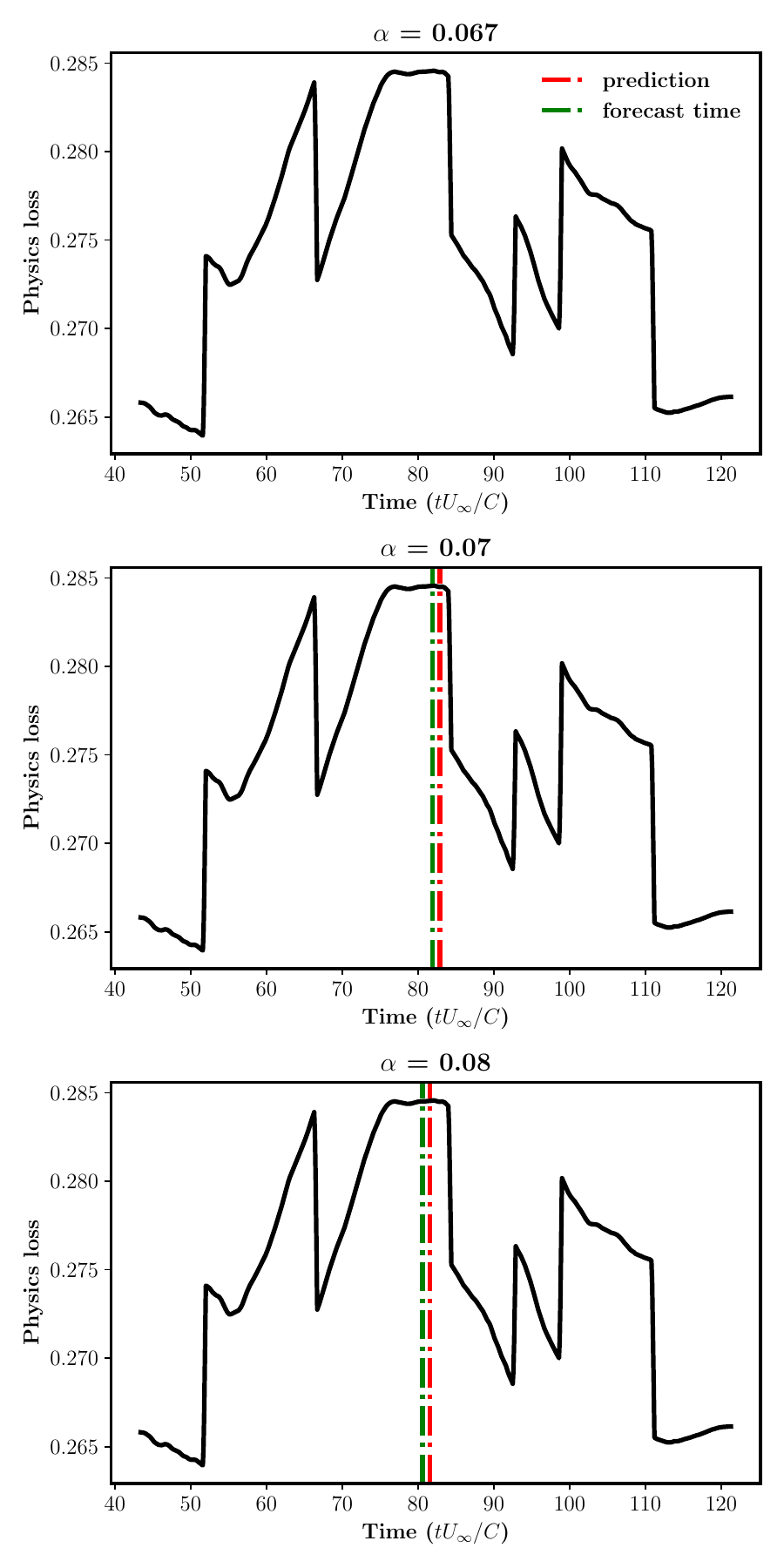}
            \caption{\RNO}
            \label{fig:rno_runNoTip}
        \end{subfigure}%
        \begin{subfigure}{0.33\textwidth}
            \centering
            \includegraphics[width=\textwidth]{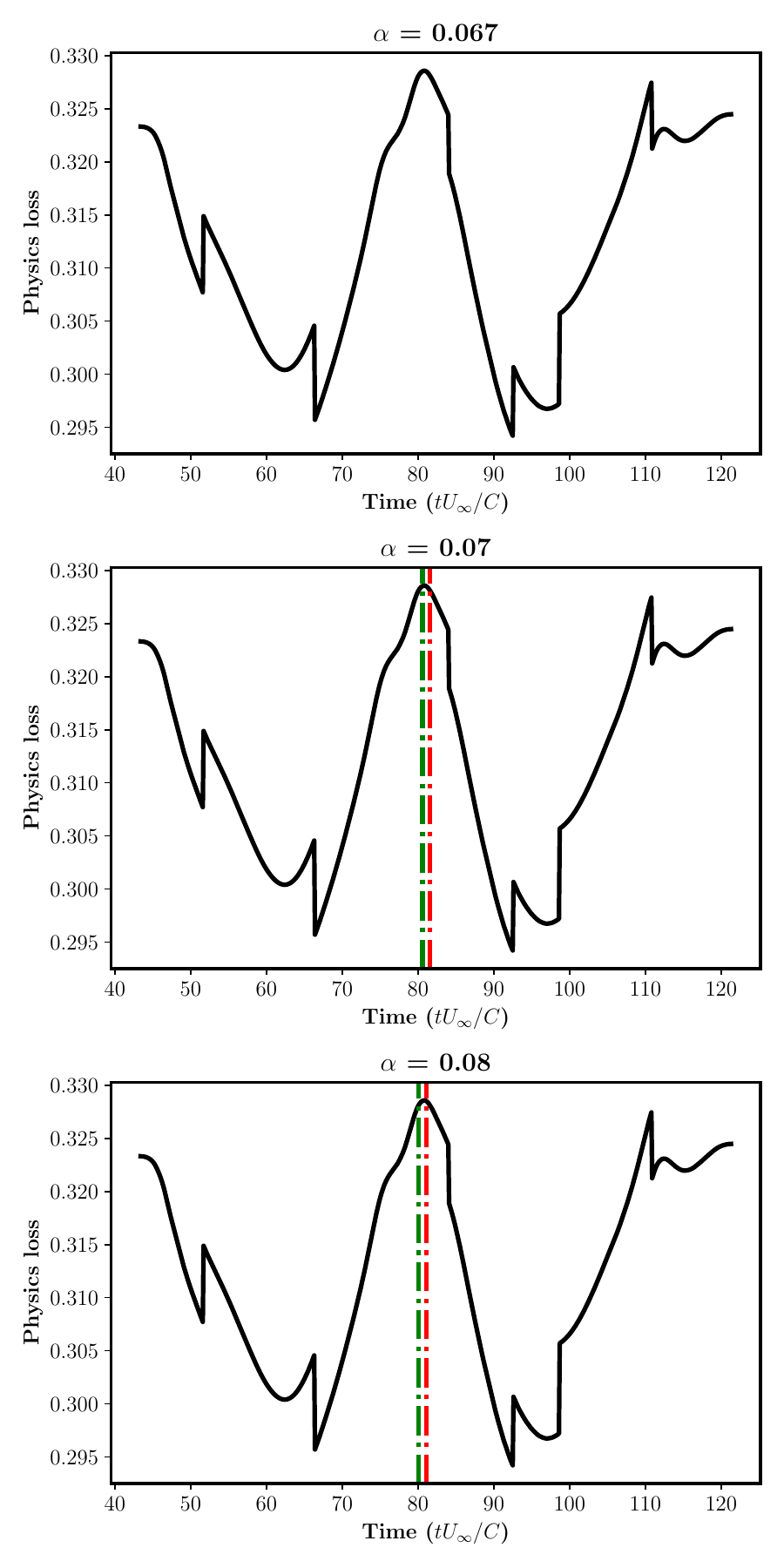}
            \caption{\MNO}
            \label{fig:mno_runNoTIp}
        \end{subfigure}%
        \begin{subfigure}{0.33\textwidth}
            \centering
            \includegraphics[width=\textwidth]{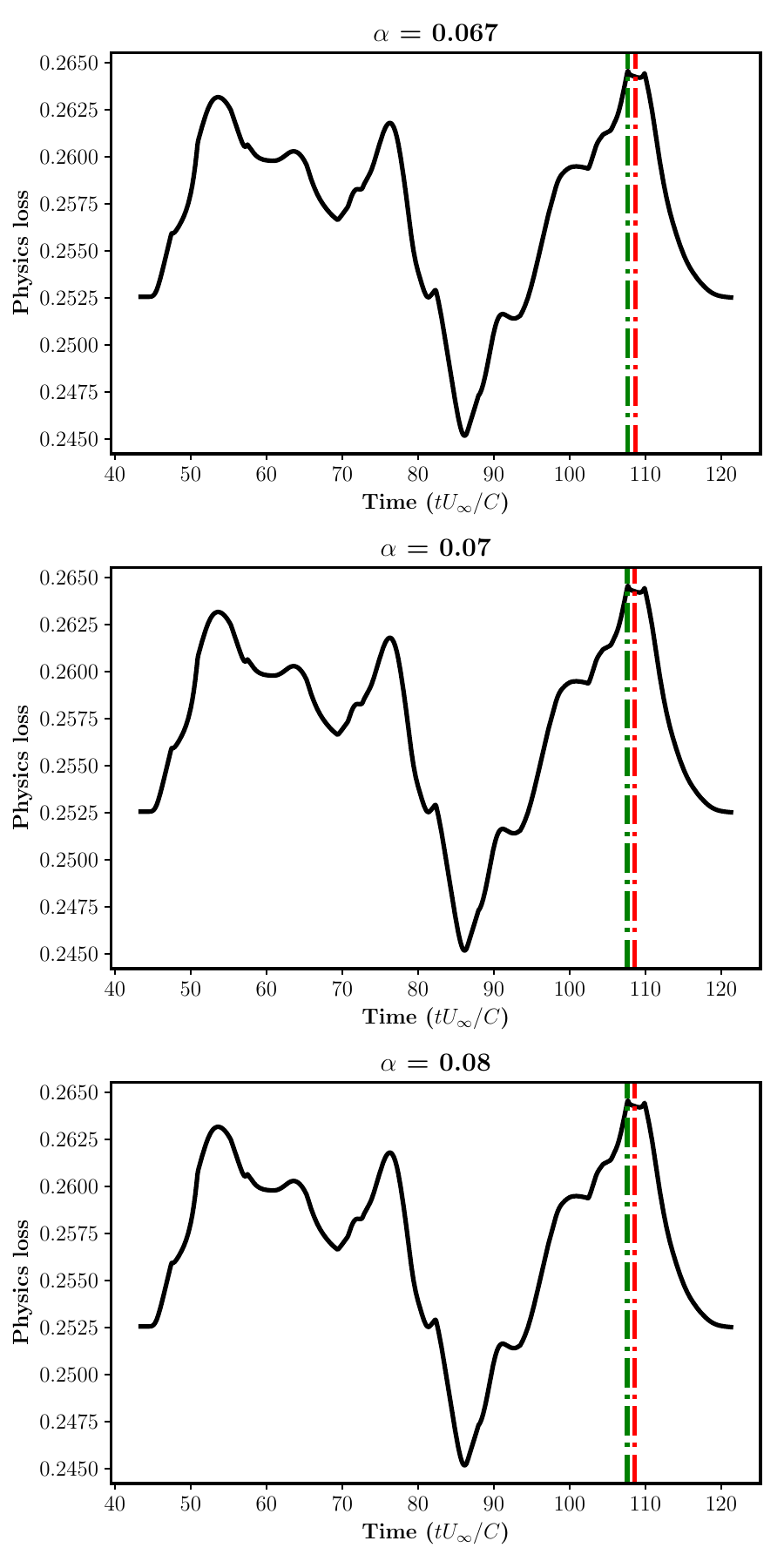}
            \caption{\RNN}
            \label{fig:rnn_runNoTip}
        \end{subfigure}%
    \caption{Comparison of tipping point forecasting performance between \RNO, \MNO, and \RNN trained on non-tipping dynamics for various false-positive rates $\alpha$.}
    \label{fig:results_NoTip}
\end{figure}

We also train and evaluate our models on a non-stationary setting with no tipping point. In this setting, we keep the same 0.33 pitch rate as the airfoil experiments in the main paper, up to an angle of attack of 10 degrees, where we reverse the pitch rate until the angle of attack returns to zero. The results are shown in Figure~\ref{fig:results_NoTip}. We compare across three false positive rates, where the first is the minimal possible (according to the calibration set size) $\alpha$. For all three models, we see that the variance of the physics loss is substantially smaller than for the tipping cases. Further, we see that \RNO and \MNO can successfully avoid predicting a tipping point at the smallest $\alpha$.

We note that by construction, it is difficult to avoid predicting a tipping point for any significance level $\alpha$ larger than the minimum allowable value when the evaluation trajectory is sufficiently long. In this case, the smallest admissible $\alpha$ (determined by the calibration set size) corresponds roughly to declaring a tipping point only when the observed physics error exceeds the maximum calibration error. As the calibration set grows, this threshold becomes increasingly effective at suppressing false positives.

Although $\alpha$ has a clear statistical interpretation as the significance level governing the conformal prediction procedure, its practical operating point inevitably depends on the quality of the learned dynamics model (e.g., \RNO, \MNO, \RNN, etc.) and the application-specific tradeoff between early detection and false positives. When representative trajectories with known tipping events are available, practitioners may therefore select $\alpha$ empirically to achieve the desired operating characteristics prior to deployment.

\section{Details on numerical experiments}
\label{appdx:numerical_details}

In our \RNO experiments, we divide inference into a warm-up phase and a prediction phase. This choice is motivated by the empirical observation that predictions $\wh u_1, \wh u_2, \dots$ for small steps $\tau$ tend to be poor because the hidden function $h_0(x,t)$ does not contain sufficient information to properly learn complex nonstationary dynamics (which inherently require history). The effect of the warm-up phase is to allow the \RNO to construct a useful hidden representation $h_\tau$ before its predictions $\wh u_\tau$ are used to construct long-time predicted trajectories. We use this procedure both during training and test times.

Furthermore, \MNO has achieved state-of-the-art accuracy in forecasting the evolution of stationary dynamical systems by using a neural operator (Section ~\ref{sec:prelim}) to learn the Markovian solution operator of such systems. As proposed in \citep{li2022learning}, \MNO takes as input the solution at a given time, $u(x,t)$, and outputs $u(x,t + h)$. While the Markovian solution operator holds for some stationary systems~\citep{li2022learning}, non-stationary systems must be handled with more care. As such, we instead compare \RNO against a variant of \MNO that maps $u(x,[t, t + \Delta T]) \mapsto u(x,[t + \Delta T, t + 2\Delta T])$. This variant of \MNO equips the model with some temporal memory.

We perform hyperparameter tuning (e.g., for model size, learning rates, etc.) for all architectures used in our experiments. We report our final hyperparameter values for each experiment below.

\subsection{\RNN baseline}
\label{appdx:rnn_details}
We use a gated recurrent unit (\GRU) \citep{cho2014learning} architecture for the \RNN baseline for the non-stationary Lorenz-63, KS equation, and simplified cloud cover model experiments. We use shallow neural networks to map inputs to a hidden representation that is propagated between time intervals with the \GRU, and the final \GRU hidden-state is mapped to the output space with another shallow neural network. We observe optimal performance for the \RNN when we adopt the same inference process as with \RNO, with a warm-up phase and a prediction phase. We observe that the number of time intervals for the warm-up phase needed to achieve acceptable results tends to be higher for \RNN than for \RNO across our experiments. For instance, we warm up with $10$ intervals for \RNN and $5$ for \RNO, in both the Lorenz-63 and KS settings. In the cloud cover experiments, we warm up the \RNN and \RNO both with $5$ time intervals.

In Table~\ref{table:rnn_lorenz}, we note the effectiveness of multi-step fine-tuning in learning the evolution of longer trajectories in non-stationary systems (compare \RNN-8 and \RNN-1). However, if multi-step fine-tuning is taken too far (i.e., $M$ is large), we empirically observe convergence to suboptimal local minima (e.g., \RNN-12).

\subsection{Non-stationary Lorenz-63 system}
\label{sec:lorenz_details}

In our experiments, we generate 15 trajectories using a fourth-order Runge-Kutta method with an integration step of $0.001$. Ten of these trajectories are used for training and the remaining five are used for calibration and testing. Each trajectory is on the range $t \in [-600, 200]$. The initial condition for each trajectory is initialized randomly, and the solution is integrated for $50$ integrator seconds and then discarded to allow the system to reach the periodic dynamics. For training and testing, the data is temporally subsampled into a temporal discretization of $0.05$ integrator seconds.

Both the \RNO and \MNO used for experiments in Section~\ref{sec:exp} were trained with a width of $48$ and $24$ Fourier modes. See \citet{li2020fourier} for a description of the hyperparmaters of Fourier layers. We use $L = 3$ layers in the \RNO and $4$ layers for \MNO. We implement the baseline \RNN as described in Section~\ref{appdx:rnn_details}, with $3$ layers and a $248$-dimensional hidden state. We set the length of each input/output time interval to $\Delta T = 0.32$ seconds. We also fix the number of warm-up samples $N_\tau = 5$ for \RNO experiments. In our experiments, we set $M = 12$ and $\lambda_n = 1$ for all $n$, for the corresponding terms in our data loss (Eq.~\ref{eq:loss}). All models were optimized using Adam \citep{kingma2014adam} with an initial learning rate of $10^{-4}$ and a batch size of $64$. We use a step learning rate scheduler that halves the learning rate every $4000$ weight updates. We train \RNO and \MNO for $5$ epochs, and we train \RNN for $15$ epochs. We pre-train \MNO for 50 epochs on one-step prediction.

\subsection{Non-stationary KS equation}
\label{appdx:ks_experimental_details}
\begin{figure}[t]
    \centering
        \begin{subfigure}{0.75\textwidth}
            \centering
            \includegraphics[width=\textwidth]{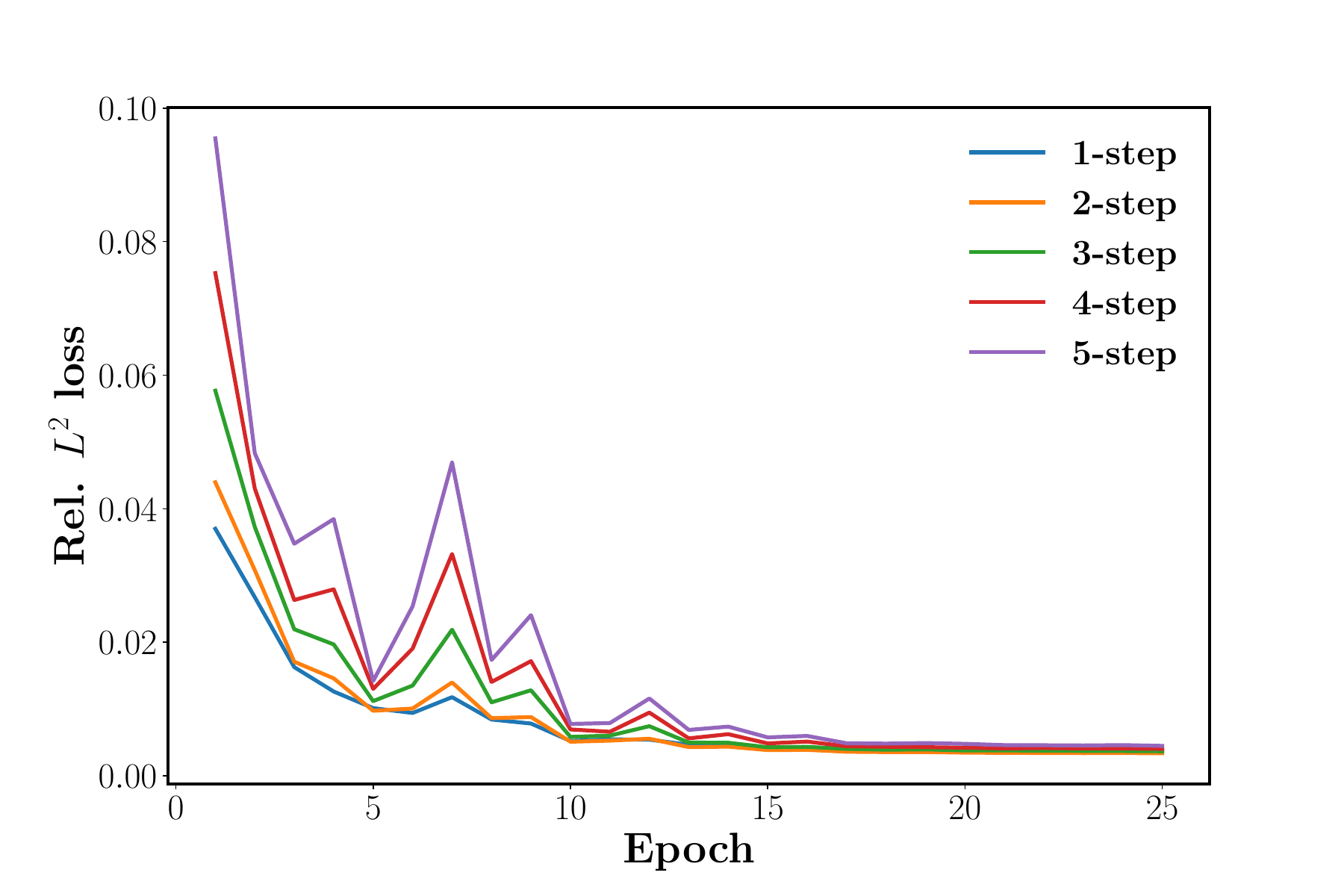}
        \end{subfigure}
        \caption{Convergence plot of test relative $L^2$ error for \RNO trained on the pre-tipping regime of the KS equation. We observe that \RNO ultimately converges to near-zero error for all steps.}
    \label{fig:ks_convergence}
\end{figure}

In our experiments, we generate 200 trajectories of the evolution of the non-stationary KS equation using a time-stepping scheme in Fourier space with a solver time-step of $0.001$. 160 of these trajectories were used for training, 30 of them were used for calibration, and 10 trajectories were used for testing. Each trajectory is defined on the range $t \in [0, 100]$. The initial condition is initialized randomly, and the solution is integrated for 100 integrator seconds and then discarded to allow the system to reach periodic dynamics. For training and testing, the data is temporally subsampled into a temporal discretization of $0.1$ integrator seconds. Our models map the previous time interval of length $\Delta T$ to the next time interval of length $\Delta T$. In our model, we use $\Delta T = 1.6$ seconds.

Both the \RNO and \MNO used in our experiments used $20$ Fourier modes across both the spatial and temporal domain. The \RNO used has a width of $28$, and \MNO used has a width of $32$. We use $L = 3$ layers in the \RNO and $4$ layers for the \MNO. We implement the baseline \RNN with $3$ layers and a $512$-dimensional hidden state. We fix the number of warm-up samples $N_\tau = 5$ for \RNO experiments. In our experiments, we set $M = 5$ and $\lambda_n = 1$ for all $n$, for the corresponding terms in our data loss (Eq.~\ref{eq:loss}). All models were optimized using Adam \citep{kingma2014adam} with a batch size of $32$. \RNO was trained for $25$ epochs at an initial learning rate of $10^{-3}$, while halving the learning rate every $2000$ weight updates. \MNO was pre-trained on one-step prediction for $100$ epochs at a learning rate of $0.005$, halving the learning rate every $200$ weight updates. \MNO was then fine-tuned using the loss in Eq.~\ref{eq:loss} at a learning rate of $10^{-5}$, halving the learning rate every $250$ weight updates. \RNN was pre-trained on one-step prediction for $40$ epochs at a learning rate of $0.001$, halving the learning rate every $500$ weight updates. \MNO was then fine-tuned using the loss in Eq.~\ref{eq:loss} at a learning rate of $10^{-5}$, halving the learning rate every $500$ weight updates.

Figure~\ref{fig:ks_convergence} depicts the convergence in test $L^2$ error of \RNO trained on the KS equation. We obseve that \RNO converges to a test $L^2$ error close to zero for all steps. Note that at times, training is slightly unstable for $1$-step prediction, and this error propagates to larger degrees for $2$-step, $3$-step, etc., predictions. Empirically, we do not observe that this behavior prevents \RNO from adequately learning the multi-step dynamics. In fact, observe that after minor increases in $1$-step error in Figure~\ref{fig:ks_convergence}, when the $1$-step error decreases, the multi-step error decreases accordingly.

\begin{table*}
\begin{center}
\begin{tabular}{l|cccccccc}
\multicolumn{1}{c}{\bf Model}
&\multicolumn{1}{c}{\textbf{1-step}}
&\multicolumn{1}{c}{\textbf{2-step}}
&\multicolumn{1}{c}{\textbf{4-step}} 
&\multicolumn{1}{c}{\textbf{8-step}}
&\multicolumn{1}{c}{\textbf{16-step}}
&\multicolumn{1}{c}{\textbf{32-step}} 
&\multicolumn{1}{c}{\textbf{64-step}}
&\multicolumn{1}{c}{\textbf{128-step}} 
\\
\hline 
\hline 
\rule{0pt}{1em} \RNO & $\mathbf{0.0053}$ & $\mathbf{0.0033}$ & $\mathbf{0.0027}$ & $\mathbf{0.0028}$ & $\mathbf{0.0068}$ & $\mathbf{0.0434}$ & $\mathbf{0.1549}$ & $\mathbf{0.2585}$ \\

\rule{0pt}{1em} \RNN-8 & $0.0141$ & $0.0139$ & $0.0173$ & $0.0150$ & $0.0332$ & $0.1096$ & $0.2366$ & $0.3434$ \\

\rule{0pt}{1em} \RNN-12 & $0.0532$ & $0.0594$ & $0.0598$ & $0.0618$ & $0.0813$ & $0.1433$ & $0.2231$ & $0.3052$ \\

\rule{0pt}{1em} \RNN-1 & $0.0114$ & $0.1644$ & $0.2729$ & $0.3546$ & $0.4030$ & $0.4153$ & $0.4057$ & $0.4006$ \\
\hline
\hline 
\end{tabular}
\end{center}
\caption{Relative $L^2$ errors on non-stationary Lorenz-63 for different $n$-step prediction settings and number of \RNN fine-tuning steps $M$ (denoted ``\RNN-$M$''). \RNO outperforms \RNN on every forecasting interval and for every \RNN training setup.} 
\label{table:rnn_lorenz}
\end{table*}  

\subsection{Cloud cover equations}
\label{appdx:cloud_cover_experiment_details}
We generate 150 trajectories of the evolution of the non-stationary cloud cover equations~\citep{Singer2023a, Singer2023b} described in Appendix~\ref{sec:cloud_cover_model} using a 5th order Runge-Kutta Rosenbrock method. We use a integration step of 1 day and solve the system for 20 years for each trajectory. We train our \RNO model on 45 trajectories, use 100 for calibration, and 5 for testing. In our models, we set the forecast time length to be $\Delta T = 0.36$ years. 

The \RNO used in our experiments used $64$ Fourier modes across the temporal domain and used a width of $128$. We use $L = 5$ layers and multi-step fine-tuning up to $M = 10$ steps. The \MNO model used $4$ layers, $32$ Fourier modes, and a width of $96$. We implement the baseline \RNN with $3$ layers and a $1024$-dimensional hidden state. We fix the number of warm-up samples $N_\tau = 5$ for \RNO experiments. In our experiments, we set $M = 10$ and $\lambda_n = 1$ for all $n$, for the corresponding terms in our data loss (Eq.~\ref{eq:loss}). All models were optimized using Adam \citep{kingma2014adam} with a batch size of $64$. \RNO was trained for $25$ epochs at an initial learning rate of $10^{-3}$, while halving the learning rate every $300$ weight updates. Both \MNO and \RNN were also directly trained using the loss in Eq.~\ref{eq:loss} at a learning rate of $10^{-3}$, halving the learning rate every $500$ weight updates.

\subsection{Airfoil experiments}
\label{appdx:airfoil_experimental_details}
The details regarding data generation and problem setup can be found in Section~\ref{appdx:airfoil_dataset}. The \RNO used in our experiments used $26$ Fourier modes aong each spatial dimension and used a width of $28$. We use multi-step fine-tuning up to $M = 3$ steps. For the Re 1000 experiments, we use $L = 3$ layers and $L = 4$ layers for Re 5000. The \MNO model used $4$ layers, $26$ Fourier modes along each spatial dimension, and a width of $28$. We implement the baseline \RNN with $3$ layers and a $64$-dimensional hidden state. We fix the number of warm-up samples $N_\tau = 4$ for \RNO experiments. In our experiments, we set $M = 3$ and $\lambda_n = 1$ for all $n$, for the corresponding terms in our data loss (Eq.~\ref{eq:loss}). All models were optimized using Adam \citep{kingma2014adam} with a batch size of $5$. \RNO was trained for $50$ epochs at an initial learning rate of $10^{-3}$, while halving the learning rate every $10$ epochs. Both \MNO and \RNN were also trained for the same number of epochs and schedule. However, \MNO used an initial learning rate of $5 \cdot 10^{-4}$, and \RNN used a learning rate of $10^{-4}$.

\section{Comparisons to prior works}

\subsection{Comparison to prior machine learning methods}
\label{appdx:comparison}

In recent years, there has been a several works using machine learning for tipping point forecasting and prediction~\citep{patel2022using,bury2021deep,patel2021using,kong2021machine,lim2020predicting, li2023tipping, deb2022machine, sleeman2023generative, huang2024deep, panahi2024machine, fabiani2024task, panahi2026unsupervised}. For instance, \citet{bury2021deep} uses a convolutional long short-term memory (LSTM) model to predict specific types of bifurcations. \citet{deb2022machine} proposes a similar methodology of classifying critical transitions, smooth transitions, and no transitions. \citet{huang2024deep} proposes to use a 1-d CNN to predict the probabilities of rate-induced tipping events.
However, such methods are not directly applicable to the large-scale spatiotemporal systems (i.e., on function spaces) that motivate our work and require access to post-tipping data. In another vein of research, reservoir computing (RC)-based methods have been used to learn the dynamics of 3d-Lorenz equation~\citep{patel2022using,patel2021using,kong2021machine,lim2020predicting, li2023tipping, panahi2024machine}. However, RC approaches do not operate on function spaces and are thus not suitable for many large-scale spatiotemporal scientific computing problems. 

More specifically, \citet{patel2022using} makes the observation that when a tipping point happens, a well-trained machine learning model on the pre-tipping regime makes a significant error, using this signal as an indicator for tipping points. However, this method requires the use of post-tipping information to compare their model forecasts against. In our work, we extend this observation into our proposed tipping point forecasting method that does not require post-tipping data. Instead, we compare our forecast against the physics constraints or differential equations driving the underlying dynamics.

Among other reservoir computing works, \citet{kong2021machine} and \citet{panahi2024machine} train data-driven reservoir models on pre-tipping dynamics, conditioning the model on certain values of a bifurcation-inducing external parameter. While such methods appear successful for simpler toy systems, for real-world systems (e.g., climate) there may be a variety of external parameters that may affect the dynamics of a non-stationary system, which may be difficult to estimate and identify. \citet{panahi2026unsupervised} addresses this by using a variational autoencoder to estimate latent driving factors from data directly, without requiring prior parameter knowledge. However, their method is based on RC and does not natively extend to function spaces or to large spatiotemporal systems at arbitrary resolutions. \citet{lim2020predicting} also use reservoir computing but require access to the full ground-truth system, which may be unknown in real-world systems. In contrast, our method can operate on partial or approximate physics knowledge of the underlying system. \citet{li2023tipping} also presents a method for tipping point prediction using reservoir computing, but this method suffers from the same scalability concerns of reservoir computing and also requires tipping points to be in the training set, both of which are addressed by our method.

In another realm of methods, \citet{sleeman2023generative} introduces an adversarial framework for tipping point prediction. However, this framework requires the querying of an oracle, which is extremely computationally expensive in large-scale spatiotemporal systems of interest. Another interesting direction is proposed in \citet{fabiani2024task}, which proposes a multi-step procedure of dimensionality reduction, learning a mesoscopic dynamical model, and identifying tipping points via numerical bifurcation analysis of this model. While this procedure proposes a unique direction compared to many prior works, it has yet to be shown whether its components can be scaled to the complex real-world systems that motivate our work.

\subsection{Comparisons with traditional early warning signals}
\label{appdx:ews_comparison}

\begin{figure}[t]
    \centering
        \begin{subfigure}{0.5\textwidth}
            \centering
            \includegraphics[width=\textwidth]{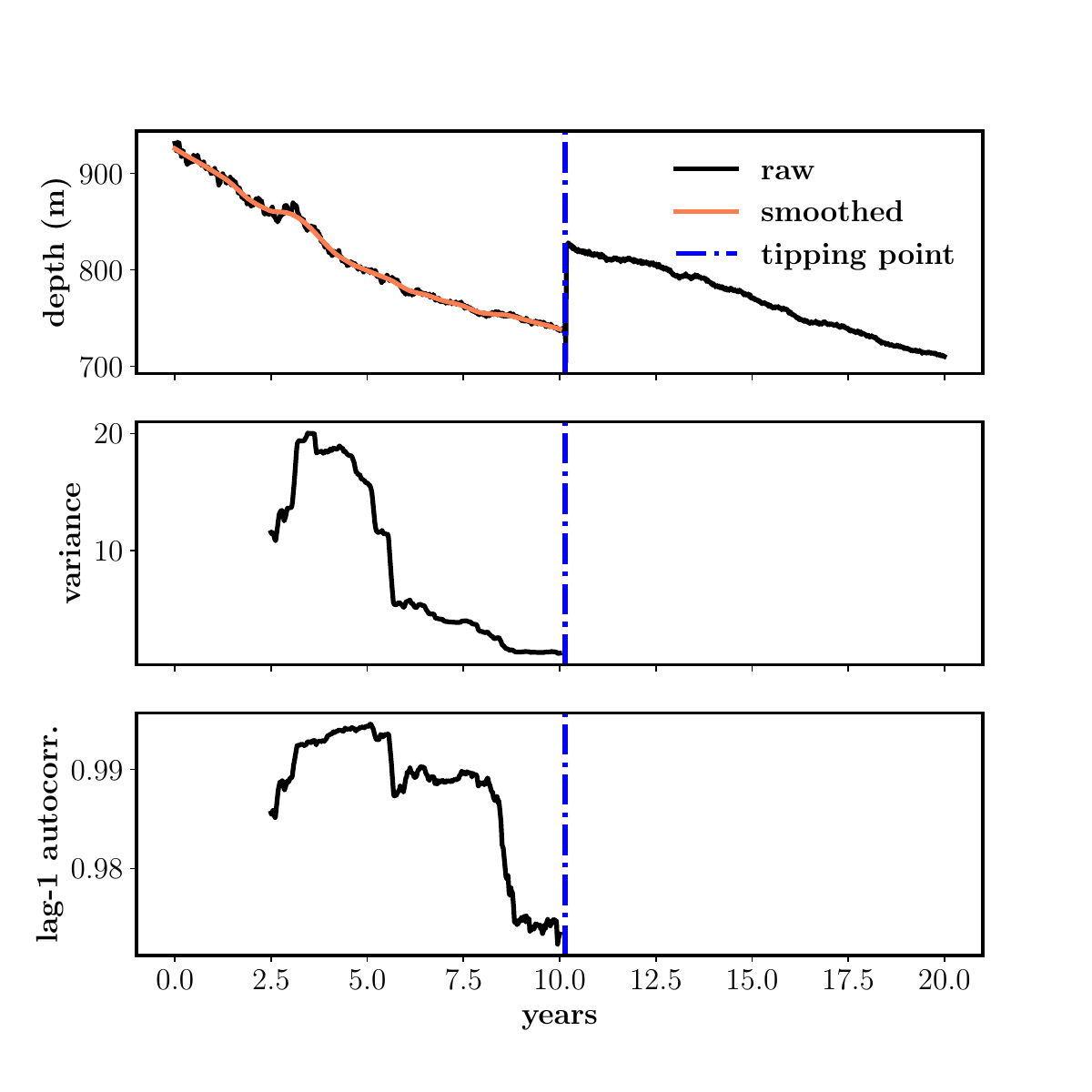}
            \caption{Boundary layer depth}
            \label{fig:ews_coord1}
        \end{subfigure}%
        \begin{subfigure}{0.5\textwidth}
            \centering
            \includegraphics[width=\textwidth]{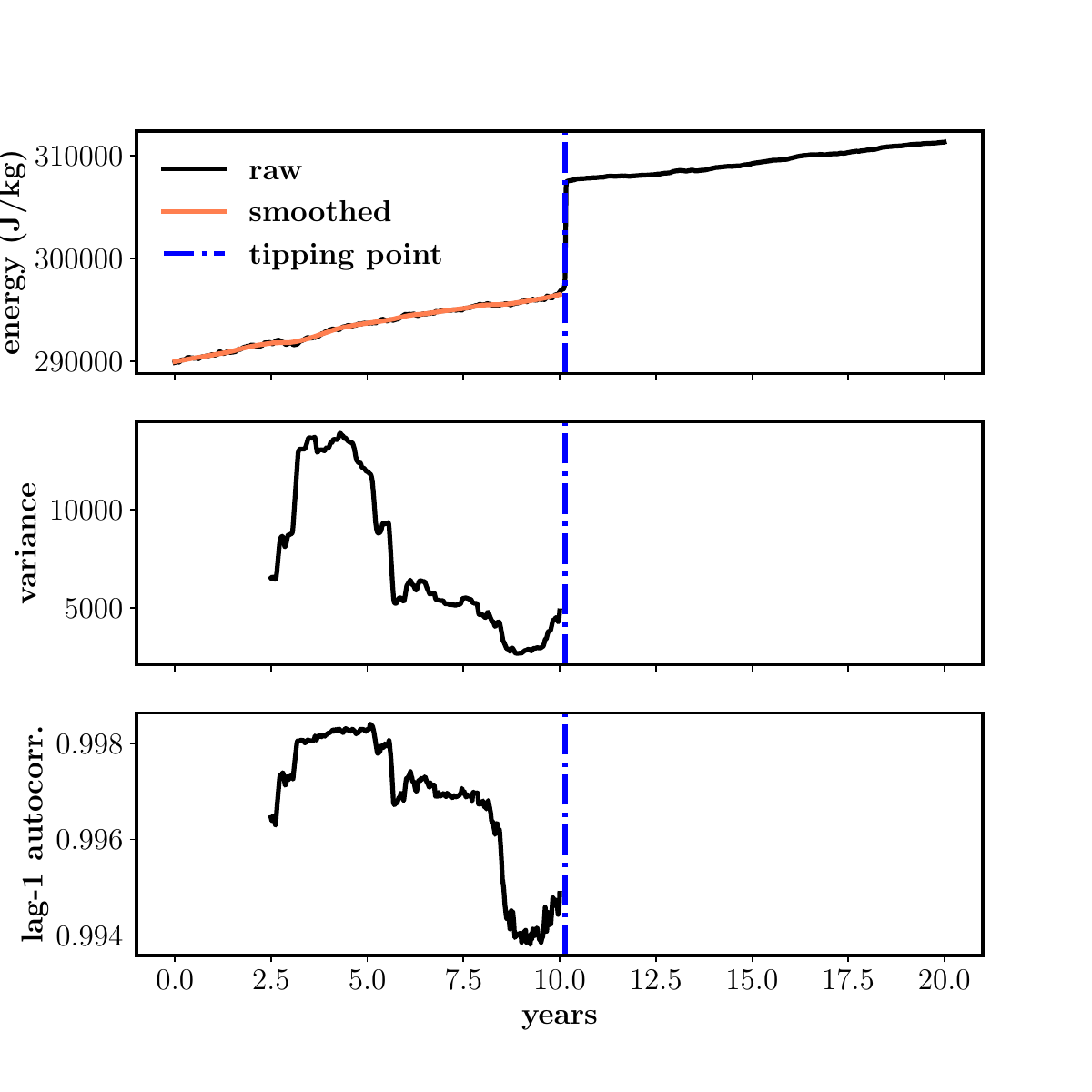}
            \caption{Liquid water static energy}
            \label{fig:ews_coord2}
        \end{subfigure}
        \begin{subfigure}{0.5\textwidth}
            \centering
            \includegraphics[width=\textwidth]{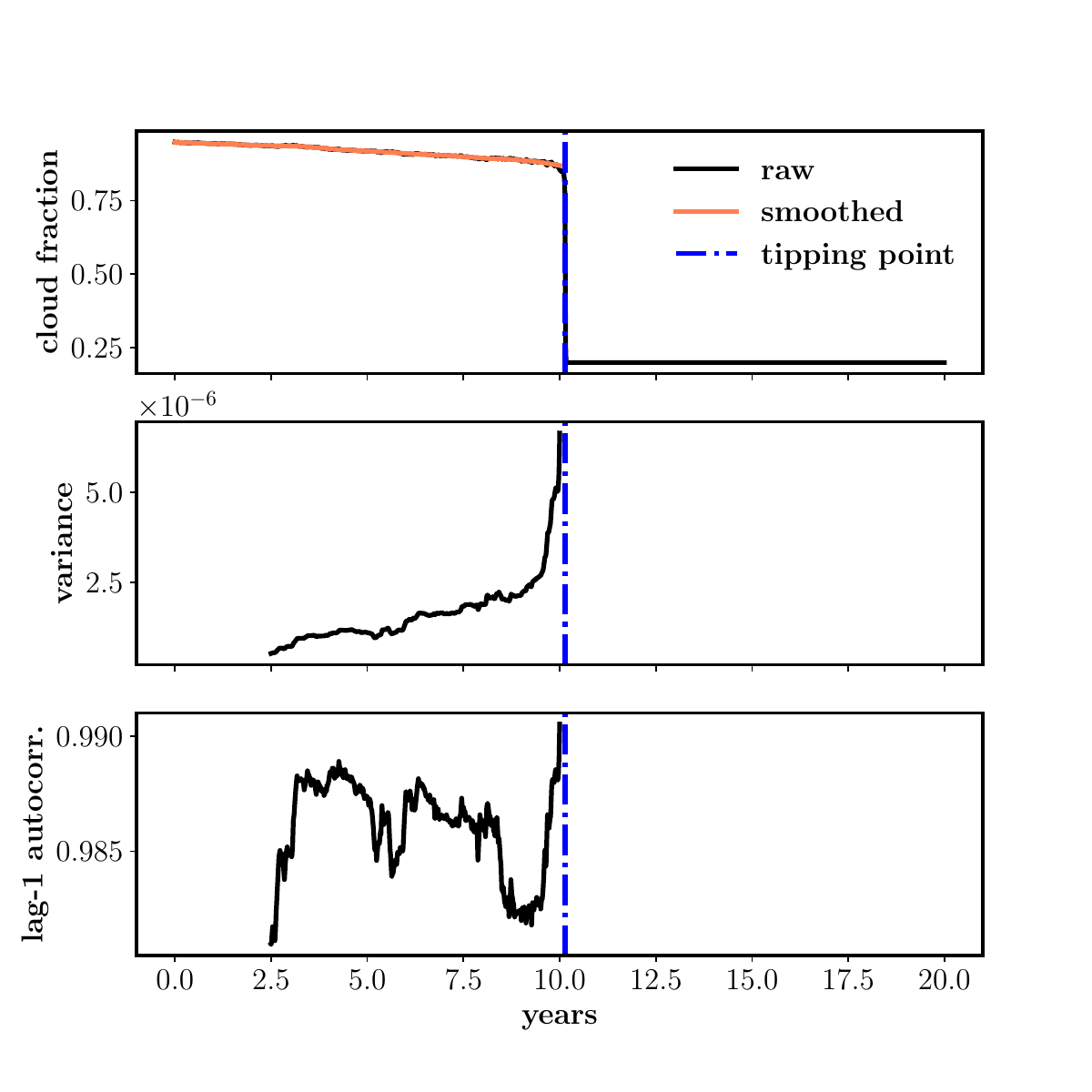}
            \caption{Cloud fraction}
            \label{fig:ews_coord5}
        \end{subfigure}
    \caption{Variance and lag-1 autocorrelation applied as early warning signals to forecast impending tipping points in the surrogate cloud cover system. Kendall's $\tau$ values for (a) variance: $-0.83$, autocorrelation: $-0.48$, (b) variance: $-0.69$, autocorrelation: $-0.59$, and (c) variance: $0.89$, autocorrelation: $-0.33$. Despite approaching a tipping point, autocorrelation did not tend to increase. Only the variance of cloud fraction increased.}
    \label{fig:ews_clouds}
\end{figure}

To analyze the ability of traditional early warning signals (EWS) to predict tipping points beyond simple well-studied systems, we apply EWS to the cloud cover system~\citep{Singer2023a, Singer2023b} in our experiments. In particular, prior works have found that for some systems, increases in variance and autocorrelation are associated with critical slowing down, a phenomenon of slow recovery from perturbations for some systems approaching bifurcation-induced tipping points \citep{scheffer2009early, dakos2012methods}. As such, we compute the variance and lag-1 autocorrelation using the methods in \citet{dakos2012methods} for each of the five variables in the cloud cover system (see Appendix~\ref{sec:cloud_cover_model} for background). The variance and autocorrelation for three of these variables is shown in Figure~\ref{fig:ews_clouds}.

Despite autocorrelation and variance being established indicators of tipping phenomena in some systems, we observe that in this real-world system, autocorrelation never has a positive Kendall $\tau$, and variance only has a positive Kendall $\tau$ in the cloud fraction setting. These results suggest that traditional EWS are not reliable as general indicators of tipping phenomena.

We follow the methodology described in \citet{dakos2012methods}, and we compute the variance and autocorrelation using the \texttt{ewstools} package~\citep{bury2023ewstools}. We smooth each univariate time-series using LOWESS smoothing with a span of $0.1$ times the length of the trajectory. We compute lag-1 autocorrelation and variance in rolling windows of length $0.25$ times the length of the trajectory.

\section{Technical discussion and analysis of methods}
\label{appdx:rno_technical_analysis}

\subsection{Memory and inference times}
The details for the best-performing \RNO and \MNO models in each of the experimental settings are provided in Appendix~\ref{appdx:numerical_details}. In this section we discuss and compare the memory usage and the inference times of \RNO and \MNO. Timing benchmarks were performed on one NVIDIA P100 GPU with 16~GB of memory.

In general comparisons of \RNO and \MNO with the same number of layers, Fourier modes, and width, \RNO tends to have a significantly larger memory footprint, since all of the gating operations are implemented with Fourier integral operators. However, in situations where memory may be scarce, weight-sharing or factorization methods can be used to vastly reduce memory footprint \citep{tran2021factorized}. Mixed-precision neural operators~\citep{white2023speeding} can also be used to dramatically decrease memory usage and inference times. Apart from the additional Fourier layers for the gating operations, \RNO inference has a warm-up period of $N_\tau$ steps to achieve a reasonable hidden state representation. This warm-up period scales with $O(N_\tau)$, since the computation is computed serially. Despite \RNO{}'s longer inference times compared to \MNO, we find that these inference times are still much faster than the numerical solvers for a variety of applications, particularly for larger-scale spatiotemporal problems.

For the non-stationary Lorenz-63 setting, our \RNO model and \MNO model both used $24$ Fourier modes and had a width of $48$. The respective memory usages are 8.1 MB and 1.9 MB, respectively. The respective inference times to generate an entire trajectory is 30.8 seconds and 11.1 seconds, respectively (averaged over 10 instances). While \RNO is slower than \MNO, it is still significantly faster than the numerical solver, which takes around 90 seconds to generate each trajectory.

For the non-stationary KS equation, we compare \RNO and \MNO models with $20$ Fourier modes across both time and space and a width of $28$. The respective memory usages are about 90 MB and 7 MB, and the respective inference times to generate a full trajectory are about 1.3 seconds and 0.3 seconds (both averaged over 100 instances). These are both significantly faster than the numerical solver, which takes about 19 seconds to simulate one trajectory.

For the simplified cloud cover experiments, the best-performing \RNO model used $64$ Fourier modes and had a width of $128$, whereas the best-performing \MNO model used $32$ Fourier modes and had a width of $96$. The corresponding memory usages are 254 MB and 10 MB, respectively. However, note that while these models are the best-performing for their architecture, they do not have the same hyperparameters. \MNO with $64$ modes and width $128$ has a memory footprint of 34 MB. The inference time to generate an entire trajectory for \RNO is 1.068 seconds, and for \MNO the inference time is 0.223 seconds, averaged over 100 instances. Note that these are both orders of magnitude faster than the numerical solver, which takes about 100 seconds to generate an entire trajectory.

\subsection{On the choice of hyperparameters for \RNO}
The primary hyperparameters of our proposed \RNO model is $L$, the number of \RNO layers, the number of Fourier modes across each dimension of the input, the width (i.e., co-dimension) of each \RNO layer, $N_\tau$, the number of warm-up intervals, and $M$, the number of auto-regressive steps used during training. 

In general, \RNO{}'s $L$ and width is analogous to the depth and width of standard neural networks, or the number of layers and dimensionality of the hidden state in \RNN{}'s. Increasing $L$ allows for more non-linear and expressive mappings to be learned, and the number of parameters in the model increases linearly with $L$. In general, under our implementation we observe that increasing $L$ beyond $5$ or $10$ layers does not confer additional benefits for the systems studied in this paper. We also observe that increasing the width of \RNO rarely improves performance significantly, only marginally, if at all.

The number of Fourier modes necessary depends heavily on the underlying dynamics of the data. For instance, for highly turbulent and chaotic fluid flows, a large of number of modes is useful to effectively parameterize and capture high-frequency information. On the other hand, for laminar fluid flows, a large number of Fourier modes may not be necessary.

Adjusting the optimal number of auto-regressive steps $M$ used during training is a trade-off between ease of training and performance (in $L^2$ error) on longer time horizons. For large values of $M$, optimization may be difficult and complicated training policies (e.g., progressive steps of pre-training for some $\tilde M < M$, then fine-tuning at $M$, etc.) may be necessary to improve performance. However, we find that using some $M > 1$ is crucial to achieving good long-term performance. This trade-off can be observed in Table~\ref{table:rnn_lorenz}. In general, the choice of $M$ may also depend on the system and the goals for the learned surrogate model. For instance, for highly chaotic systems, large $M$ are less likely to provide benefit over smaller $M$, since chaotic systems tend to quickly decorrelate from the past, making precise long-term predictions very challenging~\citep{li2022learning}.

Finally, the choice of the number of warm-up intervals $N_\tau$ also depends on the degree of non-stationarity of the system. For stationary Markovian systems, no memory warm-up is needed, in principle, so $N_\tau = 0$. For highly non-stationary systems (especially those with many latent variables), increasing $N_\tau$ can lead to an improvement in model performance. Also, for chaotic systems that decorrelate from the past quickly, large values of $N_\tau$ are likely to not bring better performance. In general, we find that $N_\tau$ need not be very large; values between 3 and 10 appear to work well for most systems. We also observe that \RNO performance appears to be quite robust to $N_\tau$, and tuning this hyperparameter often does not bring significant changes in performance.

\subsection{Limitations and discussion of proposed methods}
As with all data-driven methods, the performance of \RNO depends on its training data. On one hand, data-driven approaches are flexible, unconstrained, and have been shown to learn complex dynamics purely from data~\citep{li2022learning, pathak2022fourcastnet}. On the other hand, purely data-driven approaches often lack theoretical guarantees; in the realm of scientific computing such useful guarantees may be adhering to physically-meaningful conservation laws, for instance.

Despite lacking many theoretical guarantees, our proposed conformal prediction method is capable of quantifying the distribution of model error with respect to the physics loss. That is, our method does not assume that the data-driven \RNO (or other data-driven surrogate model) follows physical constraints perfectly in the pre-tipping regime. Using our rigorous proposed uncertainty quantification method, our tipping point prediction method is robust to model errors.

Since \RNO is purely data-driven, it can, in principle, be applied to learning the dynamics of arbitrary dynamical systems. For stationary dynamical systems, it is possible for \RNO to learn to ignore historical information and thus to simplify to \MNO. Unfortunately, if data is sparse or of low-resolution, it may be difficult for \RNO to generalize and truly learn the underlying evolution operator of the system. Providing a heuristic rule-of-thumb for the minimum resolution of the training data that \RNO needs to adequately learn the dynamics is difficult; this largely depends on the dynamics of the system in question. In general, providing training data at which the underlying physics is not resolved introduces an ill-posed learning problem. On the question of minimum amounts of training data, we find empirically that simple dynamics can be learned with $O(1000)$ training pairs, whereas complex dynamics may need $O(10000)$ training pairs to adequately learn. This is a very coarse estimate derived from our experimental settings and may not generalize to arbitrary systems.


\end{document}